\documentclass{IEEEtran}

\usepackage{amsmath,amsfonts}
\usepackage{array}
\usepackage[caption=false,font=normalsize,labelfont=sf,textfont=sf]{subfig}
\usepackage{textcomp}
\usepackage{url}
\usepackage{verbatim}
\usepackage{graphicx}
\usepackage{cite}
\usepackage{float}
\usepackage{tikz,pgf}
\usetikzlibrary{patterns}
\usepackage{enumitem}
\usepackage{mathtools}
\usepackage{cite}
\usepackage{amssymb}
\usepackage{subfig}
\usepackage[ruled,linesnumbered]{algorithm2e}
\usepackage{algpseudocode}
\usetikzlibrary{calc}
\usepackage{makecell}
\usepackage{etoolbox}

\makeatletter
\patchcmd{\@algocf@start}
  {-1.5em}
  {0pt}
  {}{}
\makeatother

\usepackage{bbm}
\usepackage{optidef}

\usepackage{xcolor} 
\usepackage{tikz}
\usetikzlibrary{fit}

\colorlet{pink}{red!40}
\colorlet{blue}{cyan!60}
\colorlet{lblue}{cyan!30}
\colorlet{dblue}{cyan!90}

\definecolor{amaranth}{rgb}{0.9, 0.17, 0.31}
\colorlet{green}{green!20}
\colorlet{yellow}{yellow!60}
\definecolor{darkgreen}{rgb}{0,0.5,0}

\SetKwComment{Comment}{\color{green!80!black!190}\color{red}{ $\triangleright$ DP}}{}
\SetKwComment{Com}{\color{blue!80!black!20}\color{blue}{ $\triangleright$}}{}

\SetKwComment{CommMerge}{\color{blue!80!black!20}\color{dblue}{ $\triangleright$ inter-group epoch}}{}

\SetKwComment{CommNoMerge}{\color{blue!80!black!20}\color{dblue}{ $\triangleright$ intra-group epoch}}{}

\usepackage{threeparttable}
\usepackage{booktabs}

\newcommand\myeq{\mathrel{\stackrel{\makebox[0pt]{\mbox{\normalfont\tiny def}}}{=}}}

\newtheorem{theorem}{Theorem}
\newtheorem{definition}{Definition}

\newtheorem{lem}{Lemma}

\newtheorem{example}{Example}

\newtheorem{assumption}{Assumption}

\newcommand{\round}{t} 
\newcommand{\iteration}{l}
\newcommand{\maxiter}{L}
\newcommand{\mper}{\mathbf{\phi}} 

\newcommand{\model}{\mathbf{\theta}} 
\newcommand{\dset}{\mathcal{D}}
\newcommand{\dsetelement}{(\mathbf{x},y)}

\newcommand{\func}{\bar{\mathcal{L}}} 
\newcommand{\emfunc}{\tilde{\mathcal{L}}} 
\newcommand{\lossfunc}{\mathcal{L}} 

\newcommand{\adj}{\mathbf{A}}
\newcommand{\randgauss}{\mathbf{G}}
\newcommand{\randfunc}{\mathbf{F}}
\newcommand{\datapool}{\calD}

\newcommand{\distr}{P}
\newcommand{\wdistr}{p}
\newcommand{\datadist}{\text{dist}}
\newcommand{\groupdist}{\rho}
\newcommand{\smooth}{\beta}
\newcommand{\lipcont}{\zeta}

\newcommand{\lsicont}{e}

\newcommand{\lrloc}{\eta}

\newcommand{\cc}[1]{\mathcal{#1}}
\newcommand{\calX}{\cc{X}}

\newcommand{\calD}{\cc{D}}

\newcommand{\calK}{\cc{K}}

\newcommand{\calM}{\cc{M}}

\newcommand{\calB}{\cc{B}}
\newcommand{\calW}{\cc{W}}
\newcommand{\calN}{\cc{N}}

\newcommand{\calR}{\cc{R}}

\newcommand{\calH}{\cc{H}}

\newcommand{\calI}{\cc{I}}

\begin{document}

\title{Controlled privacy leakage propagation\\ throughout overlapping grouped learning}

\author{
Shahrzad~Kiani,~\IEEEmembership{Graduate~Student~Member,~IEEE,} Franziska~Boenisch,  and~Stark~C.~Draper,~\IEEEmembership{Senior~Member,~IEEE}

\thanks{
Manuscript received 29 October 2023; revised 19 Feb 2024; accepted
11 June 2024; published in the IEEE Journal on Selected Areas in Information Theory
 (JSAIT), vol. 5, pp. 442-463, 2024, doi: 10.1109/JSAIT.2024.3416089. This work was supported in part by a Discovery Research Grant from the Natural Sciences and Engineering Research Council of Canada (NSERC), by an NSERC Alexander Graham Bell Canada Graduate Scholarship-Doctoral (CGS D3), and by a DiDi graduate award.} 
\thanks{This paper was presented in part at the 2024 IEEE International Symposium on Information Theory (ISIT).}
\thanks{S. Kiani and S. C. Draper are with the Department of Electrical and Computer  Engineering, University of Toronto, Toronto, ON, Canada (Emails: shahrzad.kianidehkordi@mail.utoronto.ca, stark.draper@utoronto.ca). }
\thanks{Franziska Boenisch is with the CISPA Helmholtz Center for Information Security, Saarbrücken, Germany (Email: boenisch@cispa.de). Part of the work was done while F. Boenisch was at the Vector Institute, Toronto, ON, Canada.}}

\maketitle

\begin{abstract}
Federated Learning (FL) is the standard protocol for collaborative learning. 
In FL, multiple workers jointly train a shared model. 
They exchange model updates calculated on their data, while keeping the raw data itself local.  Since workers naturally form groups based on common interests and privacy policies, we are motivated to extend standard FL to reflect a setting with multiple, potentially overlapping groups.  
In this setup where workers can belong and contribute to more than one group at a time, complexities arise in understanding privacy leakage and in adhering to privacy policies. To address the challenges, we propose differential private overlapping grouped learning (DP-OGL), a novel method to implement privacy guarantees within overlapping groups. 
Under the honest-but-curious threat model, we derive novel privacy guarantees between arbitrary pairs of workers. 
These privacy guarantees describe and quantify two key effects of privacy leakage in DP-OGL: \textit{propagation delay}, i.e., the fact that information from one group will leak to other groups only with temporal offset through the common workers and \textit{information degradation}, i.e., the fact that noise addition over model updates limits information leakage between workers.  
Our experiments show that applying DP-OGL enhances utility while maintaining strong privacy compared to standard FL setups.

\end{abstract}


\section{Introduction}
With machine learning (ML) constantly growing in popularity, many local user devices (known as edge workers) are increasingly involved in collaborative learning applications~\cite{ozfatura202212, xu2023unleashing}. In sync with such growth, federated learning (FL)~\cite{mcmahan2017communication,kairouz2021advances} has emerged as a new distributed learning paradigm. FL enables collaborative training of ML models based on workers’ data without the need to transfer the raw data to the cloud or share it with a central node (a.k.a. \textit{master}). 
Instead, the workers compute local model updates based on their data and share these updates with the master who aggregates them and applies them to the global model for iterative training.

In its standard version, FL trains a single global model that aims to yield good average performance across all workers~\cite{mcmahan2017communication}. However, in practice, FL often faces {\em data heterogeneity} when data is not independent and identically distributed  (i.i.d.) among workers~\cite{zhu2021federated}. This can bias the learned model toward specific data distributions among workers. 

Furthermore, the design of standard FL implements the implicit assumption
that the workers share the same level of trust toward all other workers. Yet, in practice, for various social, economic, or logistical reasons, workers may associate or trust certain workers more than others and hence cluster into (often overlapping) groups for the purpose of collaborative model development. For example, privacy constraints could prohibit healthcare or financial institutions in certain
regions from collaborating directly with distant institutes. However, collaboration within proximity or upon agreed terms might be possible. In such a context, overlapping groups might arise when institutes share mutual interests across regions. 

The baseline FL approach that confines collaboration to a single group misses the possible benefits of multi-group collaboration, which we now describe. 
\begin{itemize}[leftmargin=*]
\item \textbf{Personalization benefit:} Multiple groups allow for personalization of workers' models~\cite{deng2020adaptive,DBLP:journals/corr/abs-2002-10619,gasanov2021flix}. Workers can align their group memberships according to their data distributions and interests. Furthermore, allowing groups to overlap allows participants to maintain access to diverse group-spanning data, as they do in the baseline FL. This helps personalized models access sufficient data for effective generalization.  

\item \textbf{Communications benefit:} Another rationale for overlapping groups arises from wireless edge networks~\cite{han2021fedmes,qu2022convergence}. In a single-master FL scenario when bandwidth limits are operative, a multi-group approach can enable fairer bandwidth allocation. Moreover, in large-scale real-world FL setups, not all workers can directly connect to a single master; particularly when some are far from the server. Therefore, as mentioned in~\cite{han2021fedmes,qu2022convergence}, forming groups based on proximity and stable communication connections allows the multi-group setup to align better with limited wireless resources.

\item \textbf{Privacy benefit:} In this paper, our motivation arises from accommodating heterogeneous workers with diverse privacy requirements~\cite{yu2022per, boenisch2022individualized,boenisch2023have}. Privacy requirements not only vary among workers who hold personal data but also depend on the recipient of that personal data. For example, a patient's privacy requirements may be different when sharing patient data solely with a healthcare provider versus sharing with other institutions for research purposes. Shifting from a single group to multiple groups of smaller sizes facilitates the sharing of models within smaller, more trusted groups. 
\end{itemize}

With all the benefits of multi-group collaboration, overlap between groups yields novel privacy risks. Prior works~\cite{zhu2019deep, geiping2020inverting, boenisch2023curious} have shown that shared models have the potential to leak sensitive information about datasets. 
As a consequence, when groups overlap and workers participate in multiple groups, private information from group members that trust each other more
can leak to other less trusted groups through the overlapping workers. This yields additional privacy exposure that needs to be carefully calibrated and accounted for.

\subsection{Background}
Many personalization methods have been introduced in FL to address {\em data heterogeneity}. Some personalization methods combine local and global models, either personalizing the global model locally~\cite{jiang2019improving, fallah2020personalized} or creating a balanced mix of local and global models~\cite{deng2020adaptive,DBLP:journals/corr/abs-2002-10619,gasanov2021flix}. Some others consider pairwise collaboration between worker pairs~\cite{smith2017federated, vanhaesebrouck2017decentralized, zantedeschi2020fully}. However, the concept of a {\em group} as a multi-way collaboration structure is absent. These papers therefore do not capture the effects of overlapping groups or the associated issue of privacy leakage propagation. Group collaboration is implicitly considered in certain federated multi-task learning methods (FMTL)~\cite{ghosh2020efficient, marfoq2021federated}. Yet, they rely on simplified assumptions that limit the applicability of them to specific types of group collaboration. 

For example, clustered FL~\cite{DBLP:journals/corr/abs-2002-10619,ghosh2020efficient} is constrained to disjoint groups, neglecting potential overlaps. Moreover, in clustered FL, group memberships can dynamically change during training. Such dynamic changes in group memberships increase the risk of privacy leakage propagation. But, such risks remain unexplored. Another FMTL method by~\cite{marfoq2021federated} assumes workers' data distributions are a mixture of unknown
underlying distributions, each associated with a group. Workers collaborate to learn underlying distributions and personalize their weights for mixing. However, as the grouping structures evolve dynamically during training, these works face privacy leakage propagation risks. This is not addressed in~\cite{marfoq2021federated}.

Overlapping grouped learning is studied in~\cite{han2021fedmes,qu2022convergence}, similar to our work albeit with different motivations. These papers group workers based on proximity and stable communication connections in order to reduce communications costs. In~\cite{han2021fedmes}, a basic algorithm is proposed. In~\cite{qu2022convergence}, the authors extend~\cite{han2021fedmes} to more generalized overlapping group settings. In Sec.~\ref{sec:benchmarkAlg}, we detail the distinction between our approach to that of~\cite{qu2022convergence}.

Differential privacy (DP)~\cite{dwork2006differential} is the gold standard for reasoning about privacy guarantees, including in FL. In FL,~\cite{geyer2017differentially, wei2020federated, zhang2022understanding} have studied the potential exposure of sensitive information when workers share their local models and employ DP to enhance privacy resilience. However, they provide only a uniform worst-case privacy guarantee for all workers as they assume that workers belong to a single group. They overlook the varied privacy preferences of workers who may belong to multiple (overlapping) groups. While prior works have also explored non-uniform privacy guarantees~\cite{alaggan2016heterogeneous, jorgensen2015conservative, ebadi2015differential}, and brought this concept to ML~\cite{yu2022per, boenisch2022individualized}, they address a problem distinct from ours. They focus on a centralized (non-FL) setup and provide distinct privacy guarantees for workers. However, they guarantee a single privacy level for every worker. They do not consider group structure. In their papers, a worker cannot guarantee different levels of privacy protection based on which nodes intend to infer the worker's personal information. Some other works~\cite{feldman2021individual, jordon2019differentially, yu2023individual} perform individual privacy accounting. However, they do it again in a centralized setup and provide a single-level privacy protection for each individual. 
 
Despite all advancements in personalizing DP, to our knowledge, there is a research gap characterizing DP-enabled learning with varying privacy guarantees across overlapping groups and, importantly, the issue of privacy leakage propagation.

\subsection{Contributions}

This paper integrates overlapping group structures into FL while analyzing the concerns related to privacy leakage propagation that arise from the integration. 

\begin{figure}[!t]
    \centering \vspace{-2.5ex}
    \input{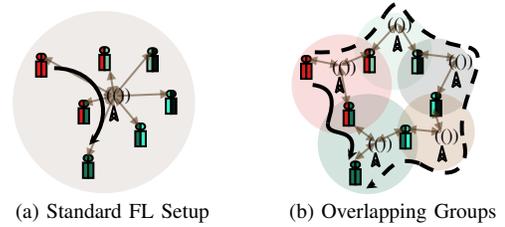}
    \caption{Seven workers collaborate in (a) one, (b) five groups. Groups are shown as colored circles, workers as humanoid shapes, and masters as antenna symbols. Workers of the same color share interests and privacy constraints.} \label{fig:structure} \vspace{-3ex}
\end{figure}

We design the algorithm ``differential private overlapping grouped learning'' (DP-OGL). DP-OGL runs periodic inter-group updates, where workers mix information about their groups. In between consecutive inter-group updates, DP-OGL runs multiple rounds of intra-group updates that confine information within each group. Privacy leakage is via \textit{propagation paths} that connect groups to groups via common workers. Figure~\ref{fig:structure} (a) depicts a traditional FL structure with seven workers wherein privacy leaks between any two workers through the master. Figure~\ref{fig:structure}(b) depicts two paths (dashed and solid black lines) between the red and blue workers who collaborate within an overlapping group structure consisting of five groups. DP-OGL uses a DP mechanism to preserve worker-level privacy from both in-group and out-of-group ``honest-but-curious'' (HbC) worker nodes. Focusing
specifically on preserving privacy from out-of-group nodes, we enhance DP-OGL to DP-OGL+ using an alternative DP mechanism.

We analyze privacy leakage propagation. The propagation of worker privacy to another worker is delayed based on the propagation paths' lengths and is also degraded due to the number of noisy mechanisms encountered along that path. This paper analyzes these effects. We introduce novel privacy amplification bounds due to {\em propagation delay} (Theorem~\ref{thm:scenario1}) and to {\em information degradation} (Theorem~\ref{thm:scenario2}). Our proof strategies for analyzing propagation delay draw on the composition tools of Renyi-DP (RDP)~\cite{mironov2017renyi} to combine mechanisms assigned to different groups without double-counting. For analyzing information degradation, we build on ideas from~\cite{ye2022differentially}. We decouple each propagation route into a sequence of noisy DP mechanisms, each followed by additive noises, which function as randomized post-processing. Our experimental results validate our theoretical bounds and benchmark privacy leakage effects for different group structures.

An outline for the rest of the paper is as follows. In Sec.~\ref{sec:problem_Setting} we detail our problem setting. In Sec.~\ref{sec:algorithm} we detail our algorithm design. In Sec.~\ref{sec:privacy} we present our privacy analysis. In Sec.~\ref{sec:benchmarkAlg} we present the discussions of our algorithms. In Sec.~\ref{sec5_simulations} we present experimental results. We conclude in Sec.~\ref{sec6}. The detailed proofs for our privacy analysis, plus a summary of our important notation, are given in the appendices.

\section{Problem Setting}\label{sec:problem_Setting}
In this section, we introduce notations to formulate the problem setting for our collaborative training approach. 

\subsection{System Design and Optimization Framework}\label{sec:system_optimization}
We consider a distributed system with $N$ workers. We use $\calN$ to denote the set of workers. Each worker $n\in \calN$ owns a personal dataset $\calD_n$ that follows distribution $\wdistr_{n}$. Each data sample in $\calD_n$ is a pair $\dsetelement$, where $\mathbf{x}\in \mathbb{R}^u$ denotes a feature vector and $y$ denotes the associated label. For example, for regression tasks, $y$ is continuous, while for classification tasks $y$ takes on categorical value. The combined set of datasets forms a {\em data pool}, denoted as $\calD:=\bigcup_{n\in \calN} \calD_n$. Letting $|\calD_n|$ denote the number of data points in worker $n$'s dataset, and assuming all workers have distinct data points, $|\calD|=\sum_{n\in \calN}|\calD_n|$. 
Individual workers employ loss functions to measure the cost of error in a supervised learning task. Worker $n$'s loss function ${\lossfunc}_{n}:\mathbb{R}^{v}\times\dset_{{n}}\rightarrow\mathbb{R}$ maps a model parameter $\mper_{n}\in\mathbb{R}^{v}$ and a data point $\dsetelement\in\dset_{{n}}$ to a cost\footnote{For the ease of notation, we generally regard $\mper_{n}$ as a real-valued vector. However, in a broader context, $\mper_{n}$ can be a vector of tensors, where each tensor represents parameters for a specific neural network layer.}. For any $n\in \calN$, the expected loss of worker $n$ is $\func_{n}(\mper_{n}):=\mathbb{E}_{\dsetelement\sim \wdistr_{n}}\left[{\lossfunc}_{n}(\mper_{n};\dsetelement)\right]$.

We assume workers are divided into $M$ subsets, each termed a {\em group}. We use $\calM$ to denote the set of groups. Workers collaborate within these groups to train machine learning models. The formation of groups is motivated by common objectives and privacy constraints which restrict certain workers from participating in real-time information exchange with others, a concept termed {\em direct} collaboration. We presume static groups, indicative of scenarios in which workers are either not permitted or unwilling to alter their group membership (e.g., in cases involving hospital data~\cite{xu2021federated}, and exclusively connect to the corresponding regional healthcare institutes). Group membership can {\em overlap}. Overlaps enable workers to participate simultaneously in multiple groups. For worker $n\in \calN$, we denote the groups to which they belong by $\calM_n\subseteq \calM$. Similarly, for group $m\in \calM$, $\calN_m\subseteq \calN$ indicates the set of workers that belong to the $m$th group. Workers within groups communicate through a master node.

For example, in Fig.~\ref{fig:structure} two systems are illustrated. Fig.~\ref{fig:structure}(a) depicts a system with $M=1$ group and $N=7$ workers. Each worker $n\in [7]$ is part of a singleton group set $\calM_n=\{1\}$, and group $1$ consists of the entire worker set $\calN_1=[7]$. Alternatively, in the case of $M>1$, Fig.~\ref{fig:structure}(b) depicts an example for $N=7$ workers and $M=5$ groups of smaller size, where each group enables direct collaboration between either two or three workers. 

In this system, our training problem is to find the $M$ model parameters $\model_{1}^{*},\ldots,\model_{M}^{*}$ which represent the solution to
\begin{mini}|s|
{\model_{1},\ldots,\model_{M}}{ \func(\model_{1},\ldots,\model_{M}):=\sum_{m\in \calM} \sum_{n\in \calN_m}\func_{n}\left(\mper_n\right)}
{}{}
\addConstraint{\mper_n=\frac{1}{|\calM_n|}\sum_{m\in\calM_n}\model_{m}, \forall n\in \calN}
\addConstraint{\model_{m}\in\mathbb{R}^{v}, \forall m\in \calM}{}.
\label{eq:fred}
\end{mini}
Each $\model_{m}^{*}$ denotes the optimal model for the $m$th collaboration group ($m\in \calM$). Each worker $n$ directly contributes to solving for $\model_{m}^{*}$ if and only if $m\in\mathcal{M}_{n}$. In other words, to train $\model_{m}^{*}$ only each worker $n\in \calN_m$ directly participates. Each worker $n\in \calN$ creates a personalized model by averaging optimal models from groups $m\in\mathcal{M}_{n}$: $\mper_{n}^{*}:=\frac{1}{|\calM_n|}\sum_{m\in\mathcal{M}_{n}}\model_{m}^{*}$. The weights for each worker's personalized model are assumed to be equal, i.e., $\frac{1}{|\calM_n|}$ for worker $n$. The study of more complex averaging methods is reserved for future research.

\subsection{Threat Model and Privacy Leakage}\label{sec:threat_privacy}
The possibility of overlaps offers the potential for broader cooperation through indirect interaction.
However, such broader cooperation comes at a cost. It leads to unintended leakage of local data across groups. While we defer presenting the leakage analysis to Sec.~\ref{sec:privacy}, to set the stage for our analysis, we now state our privacy-related assumptions and notation.

We assume that information exchanged between workers and group masters remains confidential through secure communication methods, such as channel encryption. In our setting, the group masters are trustworthy. We posit that the master can function as a central entity (e.g., a cloud) distinct from workers, which all workers trust sufficiently to share their updates. The master refrains from any curiosity-driven actions beyond what is necessary for their tasks. Potential adversaries are workers who are assumed to be ``honest but curious'' (HbC).
HbC workers target other workers' data. Workers access the set of models of the groups they belong to. We formally introduce this set in Assumption~\ref{assumption:observation_domain} in Sec.~\ref{sec:privacy}. We consider two threat models. 

\begin{definition}[Threat Model 1 (TM~1)]\label{threat1}
For any $n\in \calN$, the dataset of worker $n$ may be targeted by any HbC worker $i\in \calN\backslash {n}$, whether in-group or out-of-group.
\end{definition}
Since workers often group with trusted peers, a worker's privacy concern may be greater with respect to (w.r.t.) out-of-group HbC workers. Thus, we define the next threat model.

\begin{definition}[Threat Model 2 (TM~2)]\label{threat2}
Workers strictly avoid engaging in curiosity-driven actions concerning datasets of in-group workers. The set $\hat{\calN}_n=\bigcup_{m\in \calM_n}\bigcup_{n'\in \calN_m} {n'}$ encompasses neighboring workers who share at least one common group with worker $n\in \calN$, including worker $n$ itself. The dataset of worker $n$ is vulnerable to potential leakage to every out-of-group HbC worker $i\in \calN\backslash \hat{\calN}_{n}$.
\end{definition}

Workers have certain privacy requirements that determine their willingness to share data with fellow group members. To satisfy the workers' privacy requirements, we use DP mechanisms~\cite{dwork2014algorithmic}. DP mechanisms ensure the inclusion of any worker’s dataset is, to some extent, indistinguishable. We rely on the Gaussian mechanism~\cite{dwork2014algorithmic} to implement DP. Roughly, this mechanism first clips the updates' sensitivity of workers. The sensitivity measures the maximum potential change in the update's value due to the inclusion of a single worker's dataset. Gaussian-distributed noise is then added to the clipped update. A more formal description of the Gaussian mechanism is given in App.~\ref{app:preliminaries}.
 To quantify the level of privacy protection, we rely on $(\alpha, \epsilon)$-Rényi DP (RDP)~\cite{mironov2017renyi}, where $\alpha$ is the order of the Renyi divergence and $\epsilon$ is privacy budget. We use RDP as it has a linear composability property that allows us for a straightforward $\epsilon$ combination in our iterative training algorithm (detailed in Sec.~\ref{sec:algorithm}). 
In further preparation for our privacy analysis, we next provide some definitions.

\begin{definition}\label{def:neighboring_datapool}
Consider two data pools $\calD$ and $\calD'$, which are identical except for the presence of worker $n$'s dataset in one of them. We denote such neighboring datapools as $\calD \stackrel{n}{\equiv} \calD'$. Either $\calD = \bigcup_{i\in \calN} \calD_i$ and $\calD' = \bigcup_{i\in \calN\backslash \{n\}} \calD_i$ holds, or $\calD = \bigcup_{i\in \calN\backslash \{n\}} \calD_i$ and $\calD' = \bigcup_{i\in \calN} \calD_i$ holds.
\end{definition}

This notion of neighboring is worker-specific, varying by the inclusion (or exclusion) of a specific worker's dataset. 

\begin{definition}[Per-Worker Privacy (PwP)]
Assume $\calD \stackrel{n}{\equiv} \calD'$ and the same algorithm is executed on $\calD$ and on $\calD'$. Throughout, each HbC worker $i$ accesses the output of a random mechanism $\randfunc_i$, denoted as $\randfunc_i(\calD)$ or as $\randfunc_i(\calD')$. The distributions of mechanisms $\randfunc_i(\calD)$ and $\randfunc_i(\calD')$ are denoted as $\distr_{\randfunc_i(\calD)}$ and $\distr_{\randfunc_i(\calD')}$, respectively. The algorithm guarantees $(\alpha, \epsilon)$-PwP against any potential HbC worker $i$ targeting the dataset of worker $n \in \calN$ if, under Threat Model~\ref{threat1}, $\forall i \in \calN\backslash {n}$, or under Threat Model~\ref{threat2}, $\forall i \in \calN\backslash \hat{\calN}_{n}$,  
\begin{align}\label{eq:per_worker_group_condition}
R_{\alpha}(\distr_{\randfunc_i(\calD)} \| \distr_{\randfunc_i(\calD')}) \leq \epsilon.
\end{align}
The left-hand-side of (\ref{eq:per_worker_group_condition}) is the Renyi divergence of order $\alpha$ between distributions $P=\distr_{\randfunc_i(\calD)} $ and $P'=\distr_{\randfunc_i(\calD')}$, defined as $R_{\alpha}(P \| P') := \frac{1}{\alpha - 1}\log \mathbb{E}_{x\sim P'}\left(\frac{P(x)}{P'(x)}\right)^{\alpha}$. 
\end{definition}

In our privacy analysis in Sec.~\ref{sec:privacy}, we use the concept of group distance, which we next introduce. Let $\adj$ be an adjacency matrix in which non-zero entries represent a pair of groups in $\calM$ that share at least one worker. Denoting $\adj_{m,m'}$ as the $(m, m')$th entry of $\adj$, $\adj_{m,m'}=1$ when groups $m$ and $m'$ share a worker; otherwise, $\adj_{m,m'} = 0$.  Interpreting $\adj$ as the adjacency matrix of a graph, the graph's nodes represent groups that correspond to rows and columns of $\adj$. The graph's edges represent the connections between neighboring groups. Inspired by Eq.~6.24 of \cite{NewmanNetworks2018}, we now define the shortest distance between any two groups based on matrix $\adj$.
\begin{definition}\label{def:group_distance}
The distance between groups $m,m'\in \calM$ is termed {\em group} distance, denoted as $\rho_{m,m'}$, and defined as $\rho_{m,m'} := \min_{t=0,1,\ldots} t$ st. $ [\adj^t]_{m,m'} > 0$,
where $[\adj^t]_{m,m'}$ represents the $(m, m')$th entry of $\adj^t$ (the $t$-th power of $\adj$). 
\end{definition}
\section{Algorithm Design}\label{sec:algorithm}
To solve for the $\model_1^{*},\ldots,\model_{m}^{*}$ that minimize~(\ref{eq:fred}), we propose the DP-OGL algorithm under Threat Model~\ref{threat1}, and propose the DP-OGL+ algorithm under Threat Model~\ref{threat2}. In Sec.~\ref{sec:DP-OGL}, we introduce DP-OGL. In Sec.~\ref{sec:DP-OGL+}, we introduce DP-OGL+.

\begin{algorithm}
\caption{{Differential Private} Overlapping Grouped Learning  ({DP}-OGL)}\label{alg:Fplus}
\begin{algorithmic}[1]
\State \textbf{Inputs:} $\model_{1}^{1},\ldots,\model_{M}^{1}$
\State \textbf{Results:} $\model_{1}^{T},\ldots,\model_{M}^{T}$
\State \textbf{For} epoch $\round =1,2,\ldots, T$ \textbf{do}
\State  \hspace{1em} \textbf{For} group $m\in \calM$ in parallel \textbf{do}
\State \hspace{1em}\hspace{1em} $\model_{m}^{\round+1}$ =  \texttt{Master\_coordinates}$\left(m,t,\{\model_{m'}^{\round}\}_{\substack{ m'\in \calM_n \\   n\in \calM_m}}\right)$
\State \hspace{1em} \textbf{end}
\State \textbf{end}

\Statex

\State \textbf{Def} \texttt{Master\_coordinates} $\left(m, t, \{\model_{m'}^{\round}\}_{\substack{ m'\in \calM_n\\  n\in \calN_m}}\right)$
\State \hspace{1em} \textbf{If} $(t-1\mod S)=0$
\State \hspace{1em} \hspace{1em} Select all workers in $\calN_m$
\State \hspace{1em} \hspace{1em} \textbf{For} worker $n\in\calN_m$ in parallel \textbf{do}
\State \hspace{1em} \hspace{1em} \hspace{1em} $\mper_{n,m}^{\round,0}= \texttt{Worker\_merges}\left(n,\{ \model_{m'}^{\round}\}_{m'\in \calM_n}\right)$
\State \hspace{1em} \hspace{1em} \textbf{end}
\State \hspace{1em} \textbf{else}
\State \hspace{1em} \hspace{1em} Initialize $\mper_{n,m}^{\round, 0}=\model_m^{\round}$
\State \hspace{1em} \textbf{end}
\State \hspace{1em} Sample workers in $\calW_m^{\round }\subseteq \calN_m$ w.r.t. $\pi_{m}$
\State \hspace{1em} \textbf{For} worker
$n\in\calW_m^{\round }$ in parallel \textbf{do}
\State \hspace{1em} \hspace{1em} $\Delta\mper_{n,m}^{\round}=\texttt{Worker\_trains}\left(n, m, \mper_{n,m}^{\round, 0}\right)$
\State \hspace{1em} \textbf{end}
\State \hspace{1em} Compute $\Delta\model_m^{\round} = \sum_{n\in\calW_m^t}\Delta\mper_{n,m}^{\round}$
\State \hspace{1em} Sample noise $z_{m}^{\round}\sim \calN(0,{c}_m^2\sigma_m^2  \calI_v)$
\State \hspace{1em} Perturb $\Delta\model_m^{\round} \gets z_{m}^{\round}+\Delta\model_m^{\round}$
\State \hspace{1em} $\model_m^{\round+1}=\model_m^{\round} + \frac{1}{\pi_m|\calN_m|}\Delta\model_m^{\round}$
\State  \hspace{1em} \textbf{Return} $\model_m^{\round+1}$ 

\Statex

\State \textbf{Def} \texttt{Worker\_merges} $\left(n,\{\model_{m}^{\round }\}_{m\in \calM_n}\right)$
  \State \hspace{1em} Compute $\mper_n^{\round}=\frac{1}{|\calM_n|}\sum_{m\in\calM_n}\model_{m}^{\round }$
\State \hspace{1em} \textbf{Return} $\mper_n^{\round}$

\Statex

\State \textbf{Def} \texttt{Worker\_trains} $\left(n, m, \mper_{n,m}^{\round, 0}\right)$
	\State \hspace{1em} Select mini-batches $\calB_{n,m}^{\round,\iteration}$ for $l\in [L]$
	\State \hspace{1em} \textbf{For} iteration $\iteration\in[\maxiter]$ \textbf{do}
	\State \hspace{1em} \hspace{1em} $\mper_{n,m}^{\round, \iteration} = \mper_{n,m}^{\round, \iteration-1}-\lrloc\nabla\tilde{\lossfunc}_n(\mper_{n,m}^{\round, \iteration-1};\calB_{n,m}^{\round ,\iteration})$
	\State\hspace{1em}  \textbf{end}
   \State \hspace{1em} Compute $\Delta\mper_{n,m}^{\round}=\mper_{n, m}^{\round, \maxiter} - \mper_{n, m}^{\round, 0}$ 
  \State \hspace{1em} {Clip} $\Delta\mper_{n,m}^{\round} \gets \frac{\Delta\mper_{n,m}^{\round }}{{\max \left(1,\frac{\|\Delta\mper_{n,m}^{\round} \|_2}{c_{m}}\right)}}$  
\State \hspace{1em} \textbf{Return} $\Delta\mper_{n,m}^{\round }$ 

\end{algorithmic}
\end{algorithm}

\subsection{DP-OGL}
\label{sec:DP-OGL} 
DP-OGL is iterative and spans multiple communication epochs. Within each epoch, workers conduct local training in parallel, each involving performing ${\maxiter}$ iterations. Alg.~\ref{alg:Fplus} is the pseudocode for DP-OGL. For iteration ${\iteration}\in [{\maxiter}]$ within epoch ${\round}\in\{1,2,\cdots\}$, we use $\mper_{n,m}^{{\round},{\iteration}}$ to denote the local model of worker $n$ in group $m$. Similarly, $\model_m^{\round}$ denotes the model of group $m$. In epoch $t$, $\calW_{m}^{\round} \subseteq \calN_m$ denotes the set of workers contributing to local training to group $m\in \calM$. The union of all workers being selected to contribute in at least one group in epoch $\round$ is denoted as $\calW^{\round} \subseteq \calN$. I.e., $\calW^{\round} = \bigcup_{m\in\calM} \calW_{m}^{\round}$.

\textbf{Workers model initialization:} 
Under the masters' coordination, workers initialize their local models using the shared information within their groups. As shown in Lines 9-16 in Alg.~\ref{alg:Fplus}, epochs are divided into intervals of $S\geq 1$ epochs. During the first epoch of each interval (Lines 9-14), worker $n\in \calN_m$ performs model averaging as
\begin{align}\label{eq:workerinit}
\mper_{n,m}^{\round,0}=\frac{1}{|\calM_n|}\sum_{m'\in\mathcal{M}_{n}}\model_{m'}^{\round}.
\end{align}

In the subsequent $S-1$ epochs in each interval, workers do not engage in inter-group model averaging as in (\ref{eq:workerinit}). Instead, within group $m\in \calM$, workers initialize their models with the previous epoch's group model (Line 15). Specifically, for worker $n\in \calN_m$ in group $m$, the initial model is set as $\mper_{n,m}^{\round ,0}=\model_m^{\round}$. We refer to these epochs that confine models within groups as ``intra-group'' epochs. During inter-group epochs the initial model $\mper_{n,m}^{\round,0}$ of worker $n$ is equal across all groups $m\in \calM_n$, obtained as (\ref{eq:workerinit}). Conversely, during intra-group epochs, these initial models vary across different groups. The parameter $S$ controls the frequency of inter-group model sharing. A smaller $S$ improves the group model's ability to generalize, but, as we will discuss in Sec.~\ref{sec:privacy}, accelerates privacy leakage propagation across overlapping
groups. Therefore, $S$ is a parameter that allows us to balance model generalization and privacy leakage propagation.

\textbf{Workers local training:} 
In group $m\in \calM$, $\calW_m^{\round}$ is randomly chosen using Poisson sampling (Line 17). This choice is made independently across $t$ and regardless of choices made in other groups. Worker $n\in \calN_m$ has a fixed probability $\pi_{m}\in [0,1]$, of being selected in group $m$. The sampled worker $n\in \calW_m^t$ then performs local training (Line 19). The worker selects $L$ mini-batches $\calB_{n,m}^{t,1},\dots,\calB_{n,m}^{t,L}\subseteq \calD_n$. For each $\calB_{n,m}^{t,l}$, worker $n$ computes the empirical loss:
\begin{align}
\func_{n}(\mper_{n})\approx \emfunc_{n}(\mper_{n}; \calB_{n,m}^{t,l}):= \frac{1}{|\calB_{n,m}^{t,l}|}\sum_{\dsetelement\in \calB_{n,m}^{t,l}} {\lossfunc}_{n}(\mper_{n};\dsetelement).\label{eq:eloss}
\end{align}

As shown in Lines 29-34, and following FedAvg's approach~\cite{mcmahan2017communication}, worker $n\in \calW^{\round}$ updates $\mper_{n,m}^{{\round},{\iteration}}$, using a gradient-based techniques, such as mini-batch SGD. Selecting mini-batch $\calB_{n,m}^{{\round},{\iteration}}$, the $n$th worker updates $\mper_{n,m}^{\round,\iteration}$ as 
\begin{align}\label{eq:workermodel}
\mper_{n,m}^{\round,{\iteration}}=\mper_{n,m}^{\round,{\iteration}-1}-\lrloc\nabla\tilde{\lossfunc}_{n}(\mper_{n,m}^{\round,{\iteration}-1};\calB_{n,m}^{\round,\iteration}),
\end{align}
where $\lrloc$ is the learning rate and $\nabla\tilde{\lossfunc}_{n}(\mper_n;\calB_{n,m}^{\round,\iteration})$ estimates the gradient $\nabla\func_{n}(\mper_n)$ as 
\begin{align}\label{eq:unbiased}
\nabla\tilde{\lossfunc}_{n}(\mper_n;\calB_{n,m}^{\round,\iteration})=\frac{1}{|\calB_{n,m}^{\round,\iteration}|}\sum_{\dsetelement\in\calB_{n,m}^{\round,\iteration}}\nabla{\lossfunc}_{n}(\mper_n;\dsetelement).
\end{align}

In each group $m\in \calM$, each worker $n\in \calW_m^t$ repeats (\ref{eq:workermodel}) for $\maxiter$ iterations, and computes the model update $\Delta \mper_{n,m}^{{\round}}=(\mper_{n,m}^{\round,\maxiter}-\mper_{n,m}^{\round,0})$. The first phase of integrating DP in DP-OGL is to assign the worker the task of clipping $\Delta\mper_{n,m}^{\round }$ (Line 35):
\begin{align}\label{eq:dpclipping}
\Delta\mper_{n,m}^{\round}\leftarrow\text{Clip}(\Delta\mper_{n,m}^{\round }, c_m) := \frac{\Delta\mper_{n,m}^{\round }}{{\max \left(1,\frac{\| \Delta\mper_{n,m}^{\round } \|_2}{c_{m}}\right)}},
\end{align}
where $c_m>0$ is the clipping parameter. Clipping limits $\Delta\mper_{n,m}^{\round}$ to an $L_2$-sensitivity of at most $c_m$. The worker then sends the clipped $\Delta\mper_{n,m}^{\round}$ to the group $m$'s master (Line 36).

\textbf{Masters model updating:}
The group $m$ master aggregates the workers updates $\Delta\mper_{n,m}^{\round}$, for all $n\in \calW_m^t$. As shown in Line 21, this master then linearly combines the updates as 
\begin{align}\label{eq:groupupdate}
\Delta\theta_m^{\round}=\sum_{n\in \calW_m^{\round}}\Delta \mper_{n,m}^{{\round}}.
\end{align}

With each $\Delta\mper_{n,m}^{\round}$ having an $L_2$-sensitivity limit of $c_m$, the maximum change of the $\Delta\model_{m}^{\round}$ value between any neighboring data pools is also $c_m$. This is called the worker-level $L_2$-sensitivity of $\Delta\model_{m}^{\round}$. We define this formally in App.~\ref{app:preliminaries}. Another phase of integrating DP is to have the master add Gaussian noise to each entry of the computed sum (Line 23). This noise has mean 0 and variance ${c}_m^2\sigma_m^2$. The noise is selected independently for all $m\in \calM$ and in each epoch $t$. Algebraically, given independent noise $z_m^{\round} \sim \calN(0,{c}_m^2\sigma_m^2\calI_v)$, 
\begin{align}\label{eq:dpnoise}
\Delta\model_{m}^{\round}\leftarrow \Delta\model_{m}^{\round} +  z_m^{\round}.
\end{align}
As shown in Line 24, the group $m$ master computes 
\begin{align}\label{eq:masterupdate}
\model_m^{\round+1}=\model_m^{\round}+ \frac{\Delta\theta_m^{\round}}{\pi_m|\calN_m|}.
\end{align} 
In (\ref{eq:masterupdate}), the master updates $\model_m^t$ by adding to it the value of $\frac{\Delta\theta_m^{\round}}{\pi_m|\calN_m|}$. This is an unbiased estimate of $\frac{\sum_{n\in \calN_m}\Delta \mper_{n,m}^{{\round}}}{|\calN_m|}$.

\subsection{DP-OGL+}
\label{sec:DP-OGL+} 
We modify the three-step clipping (Line 35), noise addition (Line 23), and worker sampling (Line 17), used in DP-OGL, to design DP-OGL+. We can thereby improve the privacy leakage guarantees, as we will discuss in Sec.~\ref{sec:privacy}. In contrast to DP-OGL, in each epoch DP-OGL+ refrains from clipping, adding noise, and renewing the sampled worker set. Instead, in every epoch $\tau$ that $(\tau\mod S) \neq 1$, DP-OGL+ updates $\model_m^{\tau-1}$ as
\begin{align}\label{eq:masterupdate+}
\model_m^{\tau}=\model_m^{\tau-1}+ \frac{\sum_{n\in\calW_m^{\tau-1}} (\mper_{n,m}^{\tau-1,L}-\mper_{n,m}^{\tau-1,0})}{\pi_m|\calN_m|}.
\end{align}
For any epoch $\tau>1$ that $(\tau\mod S) = 1$, let $\tau=S\tau'+1$. In these epochs, DP-OGL+ uses a Gaussian mechanism with clipping parameter $\sqrt{S}c_m$ and noise $z_m^{\text{inter},\tau}$ to update
\begin{align}\label{eq:method2}
\model_m^{\tau} = \model_m^{\tau-S} +  \frac{{\Delta}\model_m^{\tau-S:\tau} +z_m^{\text{inter},\tau} }{\pi_m|\calN_m|},
\end{align}
where
\begin{align}\label{eq:method2_0}
{\Delta}\model_m^{\tau-S:\tau} =  \sum_{n\in\calW_m^{\tau-S}}\text{Clip}\left(\sum_{t=\tau-S}^{\tau-1}(\mper_{n,m}^{t,L}-\mper_{n,m}^{t,0}),\sqrt{S}c_m\right).
\end{align}

The noise is i.i.d. across $\tau$, $z_m^{\text{inter},\tau}\sim\calN(0,Sc_m^2\sigma_m^2)$. The variance we use here is the same as the variance of the summed noise added during every $S$ epochs in DP-OGL (of (\ref{eq:dpnoise})). As we will show in Sec.~\ref{sec5_simulations}, this choice ensures a fairer comparison of the convergence performance between DP-OGL and DP-OGL+ when subjected to the same noise but different clipping functions. Additionally, in contrast to DP-OGL, DP-OGL+ samples the workers that participate once per inter-group epoch. In other words, in any group $m\in \calM$, the sampled workers in the inter-group epoch $\tau'\geq 1$ (i.e., the workers in $\calW_m^{S(\tau'-1)+1}$) continue participating in group $m$'s training for the next subsequent $S-1$ epochs. The other, non-sampled, workers remain inactive during this period (during epochs $t \in \{S(\tau'-1)+1,\ldots,S\tau'\}$). Next, during the following inter-group epoch $\tau'+1$, another set of workers $\calW_m^{S\tau'+1}$ (possibly different from the previous one) is sampled independently according to $\pi_m$. This worker sampling approach fixes the sampled set during the operation of any Gaussian mechanism in DP-OGL+.

\section{Privacy Analysis}\label{sec:privacy}
In this section, we examine worker $n$'s privacy leakage as it propagates through overlapping groups. The HbC worker $i$ under Threat Model~\ref{threat1} satisfies $i \in \calN\backslash {n}$, and under Threat Model~\ref{threat2} satisfies $ i \in \calN\backslash \hat{\calN}_{n}$. For Threat Model~\ref{threat1}, we present PwP guarantees for the DP-OGL algorithm, and for Threat Model~\ref{threat2}, we do the same thing for DP-OGL+. We next describe the data the HbC workers have access to and define the mechanism formed by combining these data.

\begin{assumption}\label{assumption:observation_domain}
In the first $\round$ epochs, all group models $\model_m^{\tau}$ where $m\in \calM_i$ and $\tau \leq \round$ are revealed to Hbc worker $i$.
\end{assumption}

Per (\ref{eq:masterupdate}) in DP-OGL, each group model $\model_m^{\tau}$ contains two terms: the prior $\model_m^{\tau-1}$ and $\frac{\Delta\model_{m}^{\tau-1}}{\pi_m|\calN_m|}$. Per (\ref{eq:groupupdate}), $\Delta\model_{m}^{\tau-1}$ aggregates information about workers that belong to group $m$. If group $m$ includes the targeted worker, $\Delta\model_{m}^{\tau-1}$ can be conceptualized as a Gaussian mechanism. This is because $\Delta\model_{m}^{\tau-1}$ is masked through clipping and additive Gaussian noise. Similarly, per (\ref{eq:method2}) in DP-OGL+, $\model_m^{S\tau'+1}$ contains $\model_m^{S(\tau'-1)+1}$ and a Gaussian mechanism term. We next combine the group models described in Assumption~\ref{assumption:observation_domain} for both DP-OGL and DP-OGL+. 

\begin{definition}[Combined mechanism]
The mechanism $\randfunc_i^{1:\round}$ is the concatenation of group models ${\model_{m}^{\tau}}$ for all $m\in\calM_i$ and across all $\tau\leq t$ that involves a Gaussian mechanism. This includes the values $\tau\in [t]$ in DP-OGL and $\tau=S\tau'+1, $ where $\tau'\in \left[\lfloor\frac{t-1}{S}\rfloor\right]$ in DP-OGL+. Mathematically, the combined mechanism $\randfunc_i^{1:\round}$ is defined as
\begin{align}\label{eq:access_set_of_HbC_worker}
\randfunc_i^{1:\round}:= \begin{cases}\left(\bigcup_{\tau \in [\round]}\bigcup_{m\in \calM_i} \model_m^{\tau}\right) & \text{in DP-OGL}\\
\left(\bigcup_{\tau' \in \left[\lfloor\frac{t-1}{S}\rfloor\right]}\bigcup_{m\in \calM_i} \model_m^{S\tau'+1}\right) & \text{in DP-OGL+}
\end{cases}.
\end{align}
\end{definition}

In Sec.~\ref{sec:privacy:module}, we calculate PwP bounds. In Sec.~\ref{sec:privacy:submodule}, we analyze privacy leakage propagation. In particular, we analyze {\em propagation delay} and {\em information degradation}, two effects that influence privacy leakage. In Sec.~\ref{sec:privacy:examples}, we overview fundamental aspects of our analysis with illustrative examples.

\subsection{Per-worker Privacy Analysis}\label{sec:privacy:module}
Here, we calculate PwP bounds for any (possibly targeted) worker $n\in \calN$. Alg.~\ref{alg:privacy} is the pseudocode of the analysis.  Consider any neighboring data pools $\calD \stackrel{n}{\equiv} \calD'$. We consider any HbC worker $i$ who accesses $\randfunc_i^{1:t}(\calD)$ or $\randfunc_i^{1:t}(\calD')$ at epoch $t$ (Line 6). We derive a privacy leakage bound that ensures indistinguishability between $\randfunc_i^{1:t}(\calD)$ and $\randfunc_i^{1:t}(\calD')$ by meeting condition (\ref{eq:per_worker_group_condition}) for $\randfunc_i=\randfunc_i^{1:t}$ (Line 7). We denote the privacy leaking of worker $n$ w.r.t. HbC worker $i$ as $\epsilon_{n,i}^{1:t}$. As shown in Line 9, for any given $\alpha$, the PwP bound for worker $n$ is 
\begin{align}\label{eq:worstcase}
\epsilon_n^{1:t}=\begin{cases}\max_{i\in\calN\backslash \{n\}} \epsilon_{n,i}^{1:t} & \text{Under Threat Model~\ref{threat1}} \\ \max_{i\in\calN\backslash \hat{\calN}_{n} }\epsilon_{n,i}^{1:t} & \text{Under Threat Model~\ref{threat2}} \end{cases}.
\end{align}

In (\ref{eq:worstcase}), $\epsilon_n^{1:t}$  is contingent on the worst-case HbC worker(s) who has access to the most possible information about worker $n$. That said, the possibility of overlaps motivates the careful analysis of $\epsilon_{n,i}^{1:t}$ as it may vary concerning HbC workers at different distances and with different group memberships.

\begin{algorithm}
\caption{PwP Analysis}
\label{alg:privacy} 
\begin{algorithmic}[1]
\State \textbf{Results:} $\epsilon_1^{1:\round}, \epsilon_2^{1:\round}, \ldots, \epsilon_N^{1:\round}$, $\round =1,2,\ldots ,T$
\State \textbf{For} epoch $\round =1,2,\ldots ,T$ \textbf{do}
\State  \hspace{1em} \textbf{For} targeted worker
$n\in\calN$
\State \hspace{1em} \hspace{1em} Consider any neighboring $\calD \stackrel{n}{\equiv} \calD'$
\State \hspace{1em} \hspace{1em} \textbf{For} HbC worker $i$ ($i\in \calN\backslash \{n\}$ in TM~\ref{threat1}, $i\in \calN\backslash \hat{\calN}_{n}$ in TM~\ref{threat2}) \textbf{do}
\State \hspace{1em} \hspace{1em} \hspace{1em} Worker $i$ accesses $\randfunc_i^{1:t}(\calD)$ and $\randfunc_i^{1:t}(\calD')$
\State \hspace{1em} \hspace{1em} \hspace{1em} Find $\epsilon_{n,i}^{1:t}$ st. $R_{\alpha}(\distr_{\randfunc_i^{1:t}(\calD)} \| \distr_{\randfunc_i^{1:t}(\calD')}) \leq \epsilon_{n,i}^{1:t}$
\State \hspace{1em} \hspace{1em} \textbf{end}
\State \hspace{1em} \hspace{1em} Compute $\epsilon_n^{1:t}$ as in (\ref{eq:worstcase})
\State \hspace{1em} \textbf{end}
\State \textbf{end}
\end{algorithmic}
\end{algorithm}
\vspace{-3.5ex}

 \subsection{Privacy Boost from Distant and Noisy Groups}\label{sec:privacy:submodule}
We now analyze $\epsilon_{n,i}^{1:t}$ for each pair $(n,i)$ of targeted and HbC workers. This details the calculation associated with Line 7 in Alg.~\ref{alg:privacy}. In the following, we first analyze propagation delays that result from the (group) distance that information traverses along a path from the source group(s) in which the targeted worker is a member to the destination group(s) that involve the HbC worker. Due to this delay, HbC workers in distant groups don't get the most recent information from the targeted worker. On the other hand, as it propagates, the information degrades. We then analyze information degradation which is a function of the noise that is added in the processing that occurs in the intermediate groups along the propagation path(s). 
 
\subsubsection{Propagation delay} 
We consider the distance between the groups containing the targeted worker $n$ and the groups containing the HbC worker $i$. Specifically, for any group $m'\in \calM_n$ that contains worker $n$, define the distance between $m'$ and the groups of HbC worker $i$ as 
\begin{align}\label{eq:GtoGdist}
\tilde{\groupdist}_{m',i}=\min_{m\in \calM_i} \groupdist_{m',m}.
\end{align}

To distinguish between group distances, we refer to $\groupdist_{m',m}$ as group-to-group (GtoG) distance, and $\tilde{\groupdist}_{m',i}$ as group-to-HbC-worker (GtoH) distance. We use GtoH distance to compute PwP bounds in Thm.~\ref{thm:scenario1}, with the proof given in App.~\ref{sec:thm:scenario1}.

\begin{theorem}\label{thm:scenario1}
Setting $\epsilon_{m'}=\frac{2\pi_{m'}^2\alpha}{\sigma_{m'}^2}$, targeting worker $n\in \calN$, and running for $t$ epochs, our algorithms achieve $(\alpha,\epsilon_n^{1:t})$-PwP with $\epsilon_n^{1:t}$ defined as ($\ref{eq:worstcase}$). Under Threat Model~\ref{threat1}, in the DP-OGL algorithm $\epsilon_{n,i}^{1:t}$ satisfies 
\begin{align}\label{eq:theorem}
\epsilon_{n,i}^{1:t} &= \sum_{m'\in \calM_n\backslash \calM_i}S\epsilon_{m'} \left(\left\lfloor \frac{\round-1}{S}\right\rfloor-\tilde{\groupdist}_{m',i} \right)\mathbf{1}_{\left\lfloor \frac{\round-1}{S}\right\rfloor>\tilde{\groupdist}_{m',i}}\nonumber 
\\ &+ \sum_{m'\in \calM_n\bigcap \calM_i} \epsilon_{m'}(t-1), \;\; \forall i\in \calN\backslash \{n\}.
\end{align}
Under Threat Model~\ref{threat2}, in the DP-OGL+ algorithm, for every $i\in \calN\backslash \hat{\calN}_{n}$, $\epsilon_{n,i}^{1:t}$ satisfies
\begin{align}\label{eq:theorem+}
&\epsilon_{n,i}^{1:t} = \sum_{m'\in \calM_n\backslash \calM_i} \epsilon_{m'} \left(\left\lfloor \frac{\round-1}{S}\right\rfloor-\tilde{\groupdist}_{m',i} \right)\mathbf{1}_{\left\lfloor \frac{\round-1}{S}\right\rfloor>\tilde{\groupdist}_{m',i}}.
\end{align}
\end{theorem}

We now make two comments about Thm.~\ref{thm:scenario1}. First, in (\ref{eq:theorem}) $\epsilon_{n,i}^{1:t}$ has two terms. The first sums the privacy leakage bound of groups that contain worker $n$ but do not contain HbC worker $i$. The second term sums up the privacy leakage bound of groups that contain both workers. In the first term in (\ref{eq:theorem}), $S\epsilon_{m'}$ corresponds to the accumulated privacy leakage bounds of the group $m'$ between two consecutive inter-group epochs. The multiplier $\left(\left\lfloor \frac{\round-1}{S}\right\rfloor-\tilde{\groupdist}_{m',i} \right)$  shows the number of inter-group epochs required for information from worker $n$ to propagate to groups that contain worker $i$. A larger GtoH distance $\tilde{\groupdist}_{m',i}$ implies a longer delay in the privacy leakage propagation, resulting in better (lower) $\epsilon_{n,i}^{1:t}$. 
The worst-case bounds $\epsilon_n^{1:t}$ depend on HbC workers' proximity to the target worker. When both workers are in the group $m'$, $\epsilon_{m'}$ accumulates linearly over the $t$ epochs, as is seen in the second sum in (\ref{eq:theorem}). 

Our second comment is about DP-OGL+ which removes the $S$ multiplier in the first term in (\ref{eq:theorem}). This ensures better (lower) privacy bounds between any two workers without a common group. Examining (\ref{eq:theorem+}) when applied to the targeted/HbC worker pair $(n,i)$ where $i\notin \hat{\calN}_{n}$, we observe (\ref{eq:theorem+}) lacks the second term in (\ref{eq:theorem}). While (\ref{eq:theorem+}) strengthens the bounds for HbC workers $i\in \calN\backslash \hat{\calN}_{n}$, it also implies that worker $n$ fully discloses data to workers $i\in \hat{\calN}_{n}$.

 \subsubsection{Information degradation}
 We now describe the effect of information degradation that, once incorporated, will improve on the privacy bounds of (\ref{eq:theorem}) for DP-OGL and (\ref{eq:theorem+}) for DP-OGL+. We first present Lem.~\ref{lem:scenario2}. This lemma computes the upper bound of the Renyi divergence between the group $m$'s model across $\calD \stackrel{n}{\equiv} \calD'$. Such computation is expressed as a recursive function of the Renyi divergence between the previous epoch's group models when using $\calD $ and $ \calD'$.

\begin{lem}\label{lem:scenario2}
Consider data pools $\calD \stackrel{n}{\equiv} \calD'$. Assume the loss functions in Alg.~\ref{alg:Fplus} are convex and $\smooth$-smooth, and that the aggregated model within group $m$ satisfies LSI with a constant $\bar{h}_m^{\tau}$ (obtained as (\ref{app:eq:lsi:sumClips}) in App.~\ref{app:corollaries_dpogl} under Threat Model~\ref{threat1} and as (\ref{app:eq:lsi:sumClips+}) in App.~\ref{app:corollaries_dpogl+} under Threat Model~\ref{threat2}). Let $\check{\epsilon}_{n,m,\tau}:= R_{\alpha}\left(\distr_{\check{\randfunc}_{m}^{\tau}(\calD)} \| \distr_{\check{\randfunc}_{m}^{\tau}(\calD')}\right)$, where $\check{\randfunc}_{m}^{\tau}$ combines group models from epoch $\tau$ that are given as inputs to group $m$'s model in epoch $\tau+1$. In other words, $\check{\randfunc}_{m}^{\tau}=\left( \bigcup_{m'\in \calM, \groupdist_{m,m'}\leq 1}\model_{m'}^{\tau}\right)$ if $(\tau\mod S)=1 $. Otherwise, $\check{\randfunc}_{m}^{\tau}=(\model_{m}^{\tau})$.
Under these assumptions, the following recursive bounds hold. In DP-OGL and under Threat Model~\ref{threat1}, 
\begin{align}\label{eq:lem:scenario2}
R_{\alpha}\left(\distr_{\model_m^{\tau+1}}(\calD) \| \distr_{\model_m^{\tau+1}}(\calD')\right) \leq \begin{cases} \frac{\alpha \check{\epsilon}_{n,m,\tau}}{\alpha+\bar{h}_m^{\tau}{c}_m^2\sigma_m^2} & \text{ if } n\notin \calN_m \\ \check{\epsilon}_{n,m,\tau} +\frac{\alpha}{2\sigma_m^2} & \text{ if } n\in \calN_m \end{cases}.
\end{align}
In DP-OGL+ and under Threat Model~\ref{threat2}, 
\begin{align}\label{eq:lem:scenario2+}
&R_{\alpha}\left(\distr_{\model_m^{S\tau'+1}}(\calD) \| \distr_{\model_m^{S\tau'+1}}(\calD')\right)  \nonumber
\\
&\leq  \begin{cases} \frac{\alpha \check{\epsilon}_{n,m,S(\tau'-1)+1}}{\alpha+\bar{h}_m^{\tau}S{c}_m^2\sigma_m^2} & \text{ if } n\notin \calN_m \\ \check{\epsilon}_{n,m,S(\tau'-1)+1} +\frac{\alpha}{2\sigma_m^2} & \text{ if } n\in \calN_m \end{cases}.
\end{align}
\end{lem}

The proof of Lem.~\ref{lem:scenario2} is similar to that of Lem.~3.2. from~\cite{ye2022differentially}. Yet, incorporating new LSI constants compatible with DP-OGL and DP-OGL+ and adjusting the lemma to adapt recursive Renyi divergence of group models introduces some added technical complexity. The complete proof of Lem.~\ref{lem:scenario2} is given in App.~\ref{app:corollaries}. In the remainder of this section, we present Thm.~\ref{thm:scenario2}. The proof of this theorem is given in App.~\ref{app:thm:scenario2}. Theorem~\ref{thm:scenario2} solves the recursive formulas (\ref{eq:lem:scenario2}) and (\ref{eq:lem:scenario2+}) for {\em string}-like group structure. String structures ensure a single simple path (no repeated groups) between any pair of groups. This is a key property for proving Thm.~\ref{thm:scenario2}.

\begin{definition}[String]\label{def:string}
A group structure is a ``string'' if it satisfies two conditions. First, each worker belongs to a maximum of two groups. Secondly, groups can be ordered such that group $m$, where $m\notin \{1,M\}$, shares at least one worker with group $m-1$ and one with group $m+1$. Group 1, located at one end of the string, only shares workers with group 2, while group $M$ at the other end only shares workers with group $M-1$. The string has an adjacency matrix that is Toeplitz with only non-zero components on two alternative diagonals immediately adjacent to the main diagonal. 
\end{definition}

\begin{theorem}\label{thm:scenario2}
Consider data pools $\calD \stackrel{n}{\equiv} \calD'$. Let $\epsilon_{m'}=\frac{\alpha}{2\sigma_{m'}^2}$ assuming $\pi_{m'}=1$. Within a string structure, 
running for $t$ epochs, our algorithms achieves $(\alpha,\epsilon_n^{1:t})$-PwP for worker $n\in \calN$ with $\epsilon_n^{1:t}$ defined as ($\ref{eq:worstcase}$). Under Threat Model~\ref{threat1}, in the DP-OGL algorithm $\epsilon_{n,i}^{1:t}$ satisfies (\ref{eq:thm:scenario2_1}),
\begin{figure*}[!t]
\begin{align}\label{eq:thm:scenario2_1}
\epsilon_{n,i}^{1:t} &= \sum_{m'\in \calM_n \backslash \calM_i} \sum_{\tau=\tilde{\groupdist}_{m',i}}^{\left\lfloor \frac{t-1}{S}\right\rfloor}\mathbf{1}_{\left\lfloor \frac{\round-1}{S}\right\rfloor>\tilde{\groupdist}_{m',i}}\left(S\epsilon_{m'} \prod_{\tau' \in (S(\tau - \tilde{\groupdist}_{m,n}) , S\tau ]}\mu_{m-S\left(\tau -\left\lfloor\frac{\tau'}{S}\right\rfloor\right)\text{sgn}(m-m')}^{\tau'}\right)\nonumber 
\\& + \sum_{m'\in \calM_n\bigcap \calM_i} \epsilon_{m'}(t-1), \;\; \forall i\in \calN\backslash \{n\}, 
\\
\epsilon_{n,i}^{1:t} &= \sum_{m'\in \calM_n \backslash \calM_i} \sum_{\tau=\tilde{\groupdist}_{m',i}}^{\left\lfloor \frac{t-1}{S}\right\rfloor}\mathbf{1}_{\left\lfloor \frac{\round-1}{S}\right\rfloor>\tilde{\groupdist}_{m',i}} \left(\epsilon_{m'} \prod_{\tau' \in (\tau - \tilde{\groupdist}_{m,n} , \tau ]}\bar{\mu}_{m-S\left(\tau -\tau'\right)\text{sgn}(m-m')}^{S\tau'+1}\right),\;\;  \forall i\in \calN\backslash \hat{\calN}_{n}, \label{eq:thm:scenario2_1+}
\end{align}
\end{figure*}
where $\mu_{j}^{\tau'}=\frac{\alpha}{\alpha+\bar{h}_{j}^{\tau'}{c}_{j}^2\sigma_{j}^2}$ if $j\notin \calM_n$, otherwise $\mu_{j}^{\tau'}=1$. Additionally, when $\calM_n\backslash \calM_i$ is not empty, $m =\text{arg min}_{j\in \calM_i}\min_{m'\in\calM_n \backslash \calM_i}\groupdist_{j,m'}$. Under Threat Model~\ref{threat2}, in DP-OGL+ $\epsilon_{n,i}^{1:t}$ satisfies (\ref{eq:thm:scenario2_1+}),
where $\bar{\mu}_{j}^{\tau'}=\frac{\alpha}{\alpha+\bar{h}_{j}^{\tau'}S{c}_{j}^2\sigma_{j}^2}$ if $j\notin \calM_n$, otherwise $\bar{\mu}_{j}^{\tau'}=1$.
\end{theorem}

We now make three comments. First, compared to Thm.~\ref{thm:scenario1}, in Thm.~\ref{thm:scenario2}  factors of $\mu_{m-S\left(\tau -\left\lfloor\frac{\tau'}{S}\right\rfloor\right)\text{sign}(m-m')}^{\tau'}$ in (\ref{eq:thm:scenario2_1}) and $\bar{\mu}_{m-S\left(\tau -\tau'\right)\text{sgn}(m-m')}^{S\tau'+1}$ in (\ref{eq:thm:scenario2_1+}) multiply $\epsilon_{m'}$. These factors are in $(0,1]$, and so reduce (strengthen) $\epsilon_{n,i}^{1:t}$. For example, under Threat Model~\ref{threat1} and when $j\notin \calM_n$, each $\mu_{j}^{\tau'}$ quantifies information degradation due to clipping and noise in group $j$. In this case, group $j$ functions as randomized post-processing~\cite{ye2022differentially}. When $j\in \calM_n$, $\mu_{j}^{\tau'}=1$, implying a Gaussian mechanism that adds a privacy bound $\epsilon_j$ to the privacy leakage calculation.

Our second comment contrasts (\ref{eq:thm:scenario2_1}) and (\ref{eq:thm:scenario2_1+}). In (\ref{eq:thm:scenario2_1+}), the removal of the factor of $S$, compared to (\ref{eq:thm:scenario2_1}), aligns with Thm.\ref{thm:scenario1}. DP-OGL+ reduces the number of DP mechanisms between targeted and HbC workers. This results in fewer degradation blocks when compared to DP-OGL. However, each degradation block in DP-OGL+ is expected to have a smaller $\bar{\mu}_{j}^{\tau'}$ than ${\mu}_{j}^{\tau'}$ in DP-OGL. This has the potential to lead to larger degradation. Consequently, the privacy improvement expected by DP-OGL+ in Thm.~\ref{thm:scenario2} is less straightforward than in Thm.~\ref{thm:scenario1}. It depends on factors such as the values of $\bar{\mu}_{j}^{\tau'}$, ${\mu}_{j}^{\tau'}$, and $S$. Combining DP mechanisms from both DP-OGL and DP-OGL+ offers a path for future exploration, as this combination has the potential to further reduce the privacy bounds by simultaneously removing the $S$ multiplier and lowering the degradation term.

Lastly, Thm.~\ref{thm:scenario2} applies exclusively to the string structure. Extending Thm.~\ref{thm:scenario2} to more generalized group structures that include at least one pair of workers with multiple distinct simple paths connecting them is challenging. The challenge lies in efficiently combining the independently gathered information from the targeted worker across various propagation paths that reach the HbC worker's groups. Generalizing Thm.~(\ref{thm:scenario2}) is a direction of future research. The proof of Thm.~\ref{thm:scenario2} forms the basis for extending these bounds to various group structures.

\subsection{Illustrative Examples}\label{sec:privacy:examples}

\begin{figure*}[ht]
\centering
\input{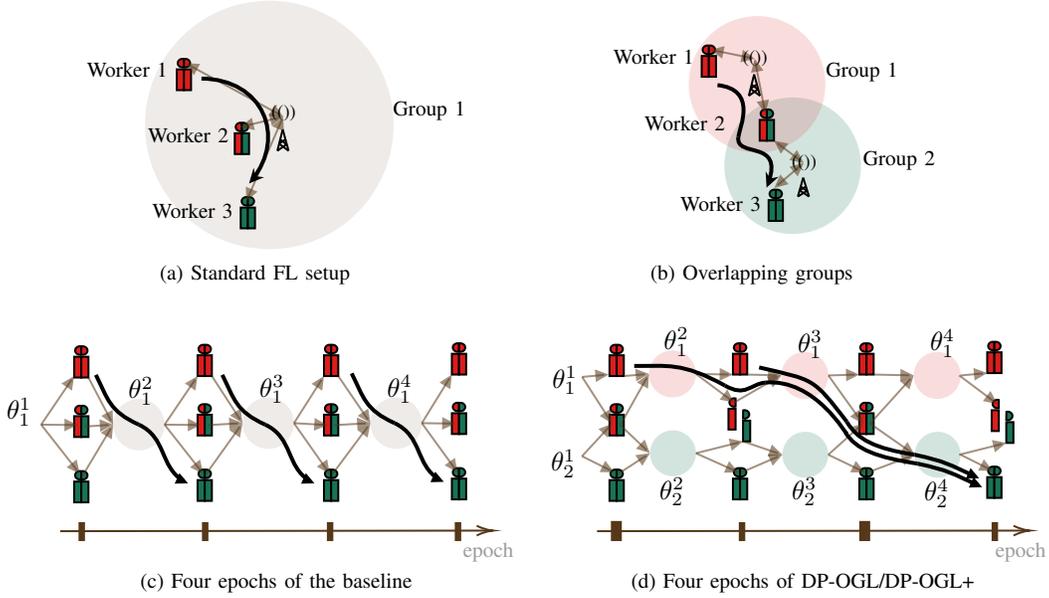}
\caption{Group structure involves $N=3$ workers, collaborating in (a) $M=1$ group (considered as baseline), (b) $M=2$ overlapping groups (string structure). Learning occurs across epochs $\tau \in [3]$, with the results of learned models distributed to workers in epoch 4, and collaborative learning structured in (c) a single group, and (d) overlapping groups.}\label{fig:group_structure_compoistion}\vspace{-2ex}
\end{figure*}

We now provide two illustrative systems with $N=3$ workers. Worker 1 is colored red, Worker 2 is a combination of red and blue, and Worker 3 is colored blue. Figure~\ref{fig:group_structure_compoistion}(a) depicts a baseline single-group FL structure. This is a classic hub-and-spoke system, where all workers connect to a single master. In Fig.~\ref{fig:group_structure_compoistion}(b), we present an example of string structure with $M=2$ overlapping groups. Group 1 includes workers 1 and 2. Group 2 includes workers 2 and 3.


 Figures~\ref{fig:group_structure_compoistion}(c) and (d) illustrate a training procedure that spans four epochs for both the baseline and string structures. Figure~\ref{fig:group_structure_compoistion}(c) consists of three gray circles. Each circle from left to right corresponds to Group 1 and the respective group model $\model_1^{\tau+1}$ that the master computes in epoch $\tau\in [3]$. Figure~\ref{fig:group_structure_compoistion}(d) consists of $3M$ circles. These circles, ordered from left to right, represent groups 1 and 2, in which the masters compute $\model_1^{\tau+1}$ and $ \model_2^{\tau+1}$, respectively, in epoch $\tau$. In Figs.~\ref{fig:group_structure_compoistion}(c) and (d), the gray arrow lines illustrate the communication links either from worker to master or from master to worker across epochs. The black arrow lines depict privacy leakage paths originating from the red worker and propagating to the blue worker.

 To simplify the notation in this section, we fix $\pi_m=1, c_m=c$ and $\sigma_m=\sigma$. Thus, $\epsilon_m$ are fixed across all groups $m\in [M]$ at $\epsilon = \frac{\alpha}{2\sigma^2}$.
In the following, we compute privacy bounds $\epsilon_{n,i}^{1:4}$ for all targeted-HbC worker pair ($n,i$). Using $\epsilon_{n,i}^{1:4}$, we compute PwP bounds $\epsilon_{n}^{1:4}$ via (\ref{eq:worstcase}).

\begin{example}[Baseline]\label{example:baseline} In this example, all workers collaborate to train a single model. The combined mechanism revealed to HbC worker $i$ across epochs $\tau \in [4]$ is 
\begin{align}\label{eq:example0_combined_mechanism}
&\randfunc_i^{1:4} =  \left(\theta_1^1, \theta_1^2, \theta_1^3, \theta_1^4\right). 
\end{align}

As the training procedure used for the baseline structure is a special case of our algorithms with $M=1$ and $S=1$, Thm.~\ref{thm:scenario2} applies. In this structure, since $M=1$ neither propagation delay nor information degradation affects, and thus privacy leakage accumulates across epochs, linearly in the number of sequentially combined Gaussian mechanisms. Under Threat Model~\ref{threat1}, this leads to privacy bounds for any $n \in [3]$ as 
\begin{align}\label{eq:example0}
\{\epsilon_{n,i}^{1:4}\}_{i\in [3]\backslash \{n\}} = 
 \{3\epsilon, 3\epsilon\}.
\end{align}
The PwP bounds are $\epsilon_{n}^{1:4}=\max_{i\in [3]\backslash \{n\}} \epsilon_{n,i}^{1:4}=3\epsilon$ for all $n\in [3]$. We note that under Threat Model~\ref{threat2}, each worker $n$ is assumed to trust any other worker $i\in \calN\backslash \{n\}$. Thus, the set $\{i\in [3]\backslash [3]\}$ is empty and both $\epsilon_{n,i}^{1:4}$ and $\epsilon_{n}^{1:4}$ are undefined. 
\end{example}

\begin{example}[DP-OGL]\label{example:overlapping_groups} 
In contrast to the single-group structure in Example~\ref{example:baseline}, we now consider $M=2$ groups. The Worker 2 that is common to groups 1 and 2 merges models from both groups every $S=2$ epochs. This is shown in Fig.~\ref{fig:group_structure_compoistion}(d) where, from left to right, during odd epochs, workers engage in inter-group collaboration by merging models from their groups, and during even epochs workers engage in intra-group collaboration. The combined mechanism revealed to the HbC worker $i$ across epochs $\tau \in [4]$ is
\begin{align}\label{eq:example_combined_mechanism}
&\randfunc_i^{1:4} = \begin{cases} \left(\theta_1^1, \theta_1^2, \theta_1^3, \theta_1^4\right) & \text{if } i=1 \\ 
\left(\theta_1^1, \theta_1^2, \theta_1^3, \theta_1^4, \theta_2^1, \theta_2^2, \theta_2^3, \theta_2^4\right) & \text{if } i=2 \\
\left(\theta_2^1, \theta_2^2, \theta_2^3, \theta_2^4\right) & \text{if } i=3
\end{cases}.
\end{align}


Due to the string structure, Thm.~\ref{thm:scenario2} can be applied. Therefore, Worker 1's privacy leakage to HbC Worker 3 is bounded by $\epsilon_{1,3}^{1:4}=2\epsilon\mu_2^4$. For Worker 1's privacy leakage to HbC Worker 2, the bound is $\epsilon_{1,2}^{1:4}=3\epsilon$. Comparing these bounds, Worker 1 mostly leaks privacy to HbC Worker 2, as both belong to Group 1 and Worker 1 is not part of any other group. Consequently, the Worker 1's worst-case PwP bound remains the same as in Example~\ref{example:baseline}, $\epsilon_1^{1:4}=\max(3\epsilon,2\epsilon\mu_2^4)=3\epsilon$. However, Worker 1's privacy leakage to the distant HbC Worker 3 is more than $1.5$ times lower than $\epsilon_{1,3}^{1:4}=3\epsilon$ in Example~\ref{example:baseline}. The more groups a worker is part of, the greater the potential for its privacy to leak. Thus we see that Worker 2 who is a member of two groups faces a larger PwP bound compared to workers 1 and 3. Applying Thm.~\ref{thm:scenario2}, Worker 2's privacy leakage to HbC Worker 1 is bounded by $\epsilon_{2,1}^{1:4}=5\epsilon$. Symmetrically, the bound on privacy leakage of Worker 2 to HbC Worker 3 is also $\epsilon_{2,3}^{1:4}=5\epsilon$. Thus, Worker 2's PwP bound is $\epsilon_{2}^{1:4}=\max(5\epsilon,5\epsilon)=5\epsilon$, which exceeds $\epsilon_{2}^{1:4}$ in Example~\ref{example:baseline}. Applying Thm.~\ref{thm:scenario2} to other targeted-HbC worker pairs, the bounds on privacy leakage are obtained as
\begin{align}
\{\epsilon_{n,i}^{1:4}\}_{i\in [3]\backslash \{n\}} = 
\begin{cases}
 \{3\epsilon,2\epsilon\mu_2^4\}& \text{if } n=1
 \\ \{5\epsilon,5\epsilon\}& \text{if } n=2 
 \\\{2\epsilon\mu_1^4,3\epsilon \} & \text{if } n=3
 \end{cases}. \label{eq:example2}
\end{align}
\end{example}

Comparing (\ref{eq:example2}) and (\ref{eq:example0}), one can see that under Threat Model~\ref{threat1} the baseline structure provides uniform privacy leakage guarantees of $\epsilon_{n,i}^{1:4}=3\epsilon$ for all $n\in [4]$ and $i\in [4]\backslash \{n\}$. In contrast, the string structure offers non-uniform guarantees. The latter is better suited to accommodate the varied privacy requirements of workers based on their group memberships.

\begin{example}[DP-OGL+]\label{example:overlapping_groups+}
In the given system configuration as in Example~\ref{example:overlapping_groups}, and considering Threat Model~\ref{threat2} for DP-OGL+, privacy leakage computation is restricted to out-of-group HbC workers. Given that Worker 2 shares one group with all other workers in the system, we only need to calculate the privacy leakage from the targeted Worker 1 to the out-of-group HbC Worker 3, and vice versa. Across epochs $\tau\in [4]$, the combined mechanism revealed to HbC Worker 1 is $\randfunc_1^{1:4} = \left(\theta_1^3\right) $, and to HbC Worker 3 is $\randfunc_3^{1:4} =\left(\theta_2^3\right) $. Due to the string structure, Thm.~\ref{thm:scenario2} again applies. 
Under Threat Model~\ref{threat2}, the privacy leakage bounds for any $n \in \{1,3\}$ are as follows:
\begin{align}
\{\epsilon_{n,i}^{1:4}\}_{i\in [3]\backslash \{2,n\}} =  \{\epsilon\}\label{eq:example2+}
\end{align}
Comparing (\ref{eq:example2+}) and (\ref{eq:example2}) shows that DP-OGL+ exhibits better (lower) $\epsilon_{1,3}^{1:4}$ than DP-OGL if $\epsilon \leq 2\epsilon \mu_2^4$. Similarly, for DP-OGL+ to have lower $\epsilon_{3,1}^{1:4}$ than DP-OGL, the condition is $\epsilon \leq 2\epsilon \mu_1^4$. In the case of $S=2$, $N=3$, and $M=2$, these conditions are satisfied if $\mu_1^4,\mu_2^4 > 0.5$.
\end{example}

\section{Discussions of DP-OGL and DP-OGL+}\label{sec:benchmarkAlg}
In this section, we discuss the highlights of DP-OGL and DP-OGL+ compared to prior works in the literature.

\subsection{DP-OGL and Other Multi-server FL Methods}
The recent work~\cite{qu2022convergence} introduces an optimization framework that shares some similarities with ours. However, their framework (and algorithm) lacks any considerations of privacy. Their algorithm, named multi-server FedAvg (MS-FedAvg), is designed for a multi-server FL system. In MS-FedAvg, each server takes on a role similar to that of the group masters in our setup, coordinating tasks among workers in order to enhance communication efficiency. However, MS-FedAvg differs from DP-OGL  (and DP-OGL+) in three aspects. 

The first distinction is the use of DP. Our algorithms use DP to inject randomness into shared models, thereby limiting privacy leakage propagation. In contrast, MS-FedAvg does not use DP. Secondly, our algorithms combine the model updates $\Delta \mper_{n,m}^{{\round}}$ as shown in (\ref{eq:groupupdate}). In contrast, MS-FedAvg averages the models $ \mper_{n,m}^{{\round},{\maxiter}}$ themselves. Another distinction is in the choice of $S$. In MS-FedAvg, workers who belong to multiple groups combine their groups' models every epoch. Thus, in MS-FedAvg $S=1$. In contrast, our algorithms are adaptable to a range of values for $S\geq 1$ which is important in controlling privacy leakage propagation. Our privacy analysis in Sec.~\ref{sec:privacy} substantiates that setting  $S > 1$ reduces such propagation.      

\subsection{DP-OGL and Other DP Variants of FL}
The way that we integrate DP in DP-OGL is known as central DP (CDP)~\cite{ramaswamy2020training}. CDP involves local clipping of updates by workers, and the noise addition by master. DP-OGL+ is a variant of CDP where masters are tasked with both clipping and noise addition. While CDP and its variants suit our threat models with HbC workers and trusted masters, it lacks DP guarantees against a malicious master who may extract data before noise addition. Extending our privacy considerations to an alternative threat model with both workers and masters being HbC poses novel challenges. Although we refrain from this extension to maintain our focus on introducing the privacy leakage propagation issue and accounting for corresponding bounds under our current threat models, in the following we discuss potential modifications for future work to consider in order to extend our algorithm designs to other threat models. 

In contrast to CDP, local DP (LDP)~\cite{ramaswamy2020training} reduces trust in the master as each worker adds local noise to their updates. However, the accumulation of the local noises after aggregation leads to significant overall noise, impoverishing the privacy-utility trade-off and making LDP unpopular in practice~\cite{wei2020federated}. In such cases, we would have to introduce some cryptography methods, such as secure aggregation~\cite{bonawitz2017practical} into our algorithm designs. This would ensure that HbC masters only observe aggregated results rather than individual updates from workers. Distributed DP (DDP) combines the advantages of CDP and LDP. For example, the DDP approach in~\cite{wei2020federated} adds two layers of noise: one centrally applied by the master and another locally by workers. This would ensure privacy protection against external malicious nodes who observe communication between workers and the master in both directions. While these DP variants can be adapted to DP-OGL, the careful calibration of such adaptations is left for future research.

\subsection{DP-OGL and Other FL Subroutines}
DP-OGL can serve as a versatile meta-algorithm applicable to different FL subroutines in the literature. While we highlight the benefits of DP-OGL in combining overlapping groups with the standard FedAvg algorithm~\cite{mcmahan2017communication}, there are different ways to extend DP-OGL beyond FedAvg. One way is to substitute Line 24 of Alg.~\ref{alg:Fplus} with an alternative master optimizer (e.g.,~\cite{reddi2020adaptive}) and Lines 30-33 with alternative worker optimizers beyond mini-batch SGD. This can open more possibilities for improved convergence performance for DP-OGL. However, as long as the masters aggregate the same number of worker updates as the standard FedAvg, our privacy bounds remain intact. DP-OGL can also use different aggregation rules, other than the standard mean aggregation. Robust aggregation methods replace the averaging calculation in FL with, e.g., median aggregation~\cite{pillutla2022robust}. The goal is to maintain the effectiveness of the aggregated model in the presence of outliers or malicious behavior. 
Our privacy bounds remain valid even if robust aggregation is used because the group models are still a function of the same updates as in the standard averaging aggregation method used in FedAvg.
\section{Experimental Evaluation}\label{sec5_simulations} 
In this section, we empirically evaluate the performance of DP-OGL and DP-OGL+ for several grouping structures. We first detail the experimental setup. We then provide our results. 

\subsection{Experimental Setup}
We consider an image classification task on the MNIST~\cite{lecun1998mnist} and Fashion-MNIST (FMNIST) datasets~\cite{xiao2017fashion}. We divide the MNIST and FMNIST datasets across $100$ workers. We simulate data heterogeneity using the Dirichlet distribution~\cite{zhang2023fedala} with a default parameter value of $ 0.1$. Our experiments involve two scenarios: employing all 100 workers (i.e., $N=100$) or exclusively employing the first set of 40 workers (i.e., $N=40$). Experiments are coded in PyTorch and conducted on an NVIDIA GeForce RTX 3080 GPU with 10 GB of memory. For both datasets, we train a 4-layer convolutional neural network (CNN), also used in~\cite{mcmahan2017communication}. For the loss function, we use negative log-likelihood. We use SGD as the optimization routine. We set the learning rate as 0.001, and mini-batch size as 200. We use $L=10$ local iterations across experiments. Our codebase is built on the code used in~\cite{zhang2023fedala, Zhang2023fedcp, zhang2023gpfl}.
 
To assess privacy, we used the Ocapus library. We used the function $\texttt{rdp.compute.rdp}$ to compute PwP bounds for various choices of $\alpha$. The optimal $\alpha$ is selected from the default set in the function $\texttt{RDPAccountant}$. We used the function $\texttt{rdp.get.privacy.spent}$ to select the optimal $\alpha$ and to convert the $\epsilon$ value from RDP to DP, assuming that $\delta=1e-5$ in DP. We use the Poisson sampling parameter $\pi_m=0.7$ and clipping parameters $c_m=0.05$, uniformly across groups $m\in [M]$. The noise multiplier $\sigma_m$ is also fixed across all groups $m\in [M]$, with its values varied among $\{1.3, 2, 3\}$ in different experiments. 
 
We explore four group structures. The first is fully global (GL) with $M=1$ group, where all $N$ workers collaborate within this group. This is akin to the example in Fig.~\ref{fig:group_structure_compoistion}(a) and emulates the DP variant of the baseline algorithm FedAvg~\cite{geyer2017differentially}. We consider $M>1$ groups in the other examples: label-based (LB), clustered (CL), and ring (RI) structures. In the LB structure, workers containing labels $y\in [10]$ are assigned to group $(y\mod M)$. In the CL structure, workers are grouped into $M$ disjoint clusters of equal size. Finally, in the RI structure, each worker participates in at most two groups (akin to Fig.~\ref{fig:structure}(b)). Each group consists of $\lfloor N/M+1 \rfloor$ workers, with the property that group $m$ shares a worker with both group $m-1$ and group $m+1$. This results in an overlap between adjacent groups and creates a closed RI structure. 

We benchmark the algorithms DP-OGL, DP-OGL+, and the baseline DP-FedAvg. DP-OGL uses RI with different values of $M\in\{4, 10\}$, LB with $M=5$, and CL with $M=4$. DP-OGL experiments evaluate different $S$ choices ($S\in \{1,2,10,25\}$). DP-OGL+ uses RI with $M\in \{4, 10\}$ and $S\in \{2,4,10\}$. The baseline comprises the GL structure ($M=1$ and $S=1$).

\subsection{Experimental Results}

\begin{figure*}[ht!]
\centering \vspace{-3ex}
\subfloat[Varied structures]{\includegraphics[width=0.25\textwidth]{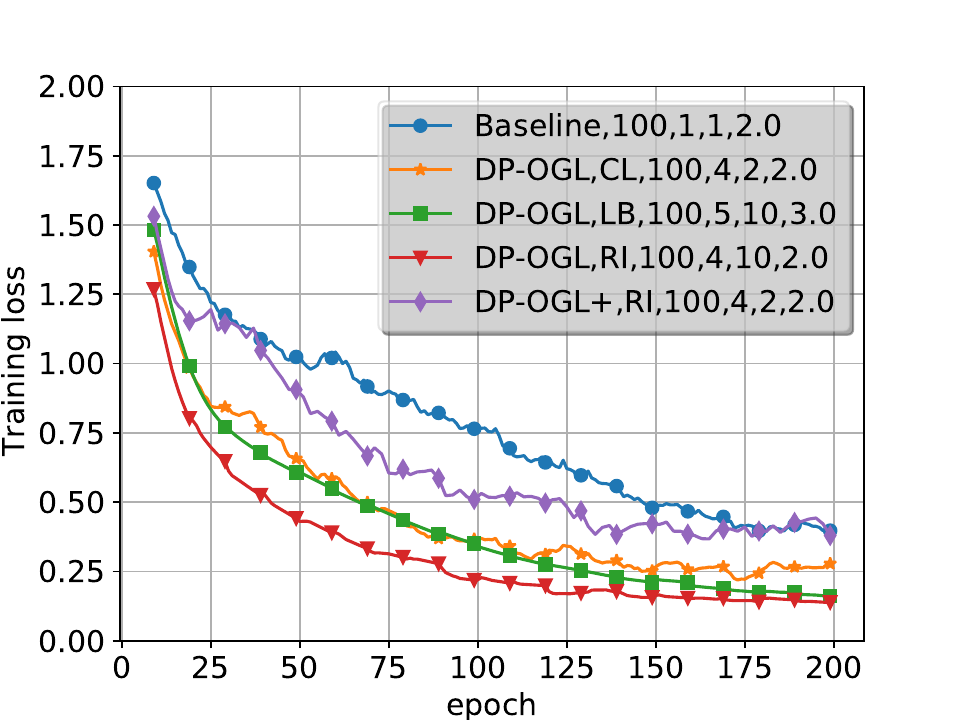
}}
\subfloat[Varied structures]{\includegraphics[width=0.25\textwidth]{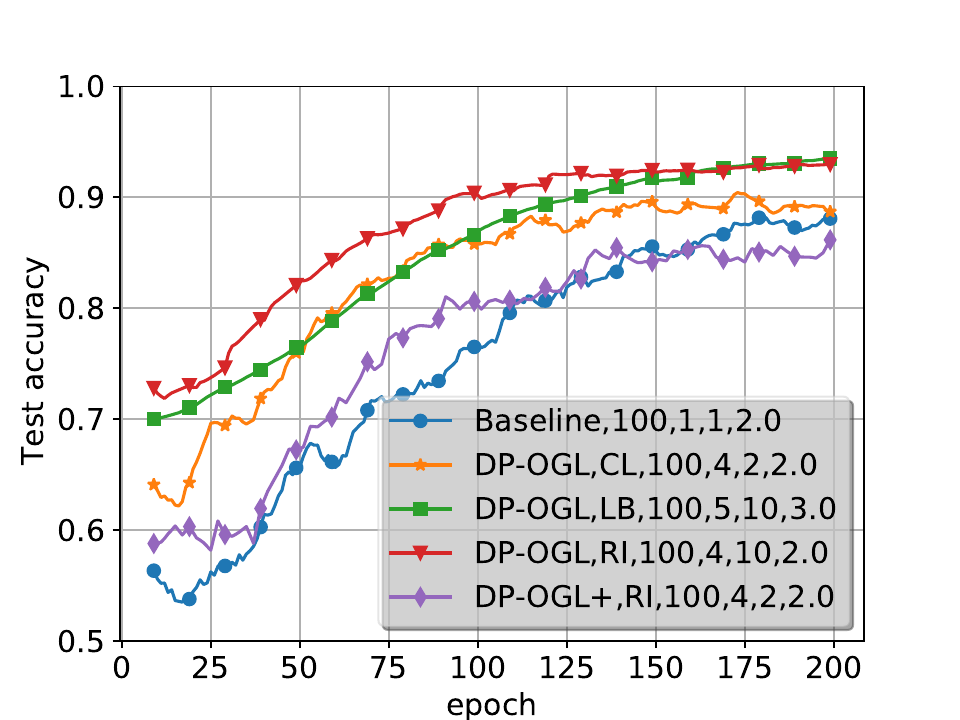
}}
\subfloat[Varied $N/M$ ratio]{\includegraphics[width=0.25\textwidth]{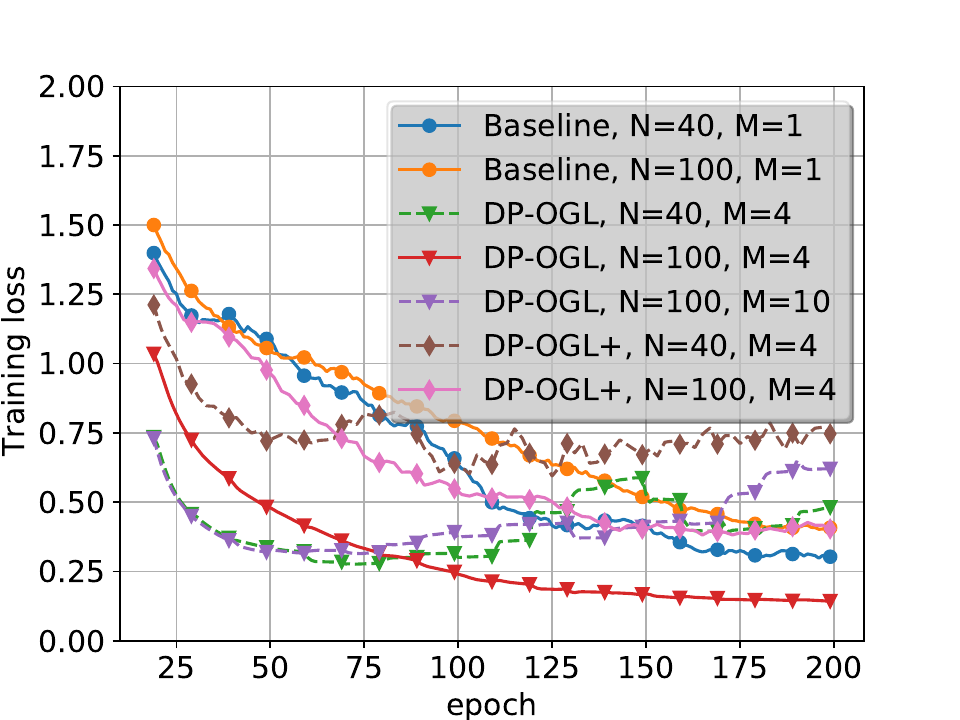
}}
\subfloat[Varied $N/M$ ratio]{\includegraphics[width=0.25\textwidth]{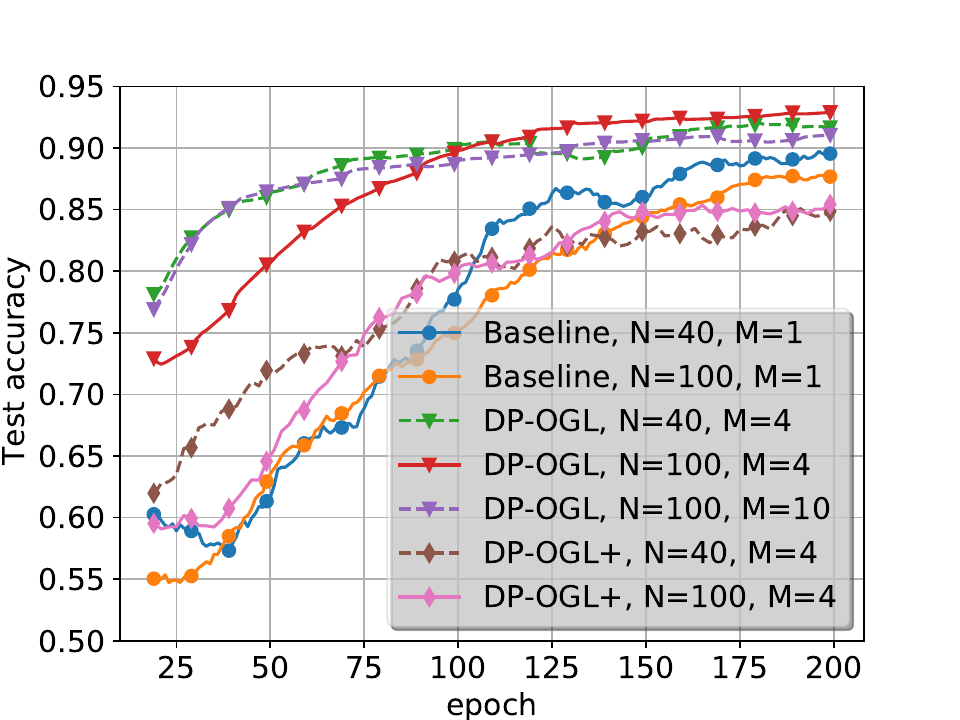
}} 

\subfloat[Varied $S$]{\includegraphics[width=0.25\textwidth]{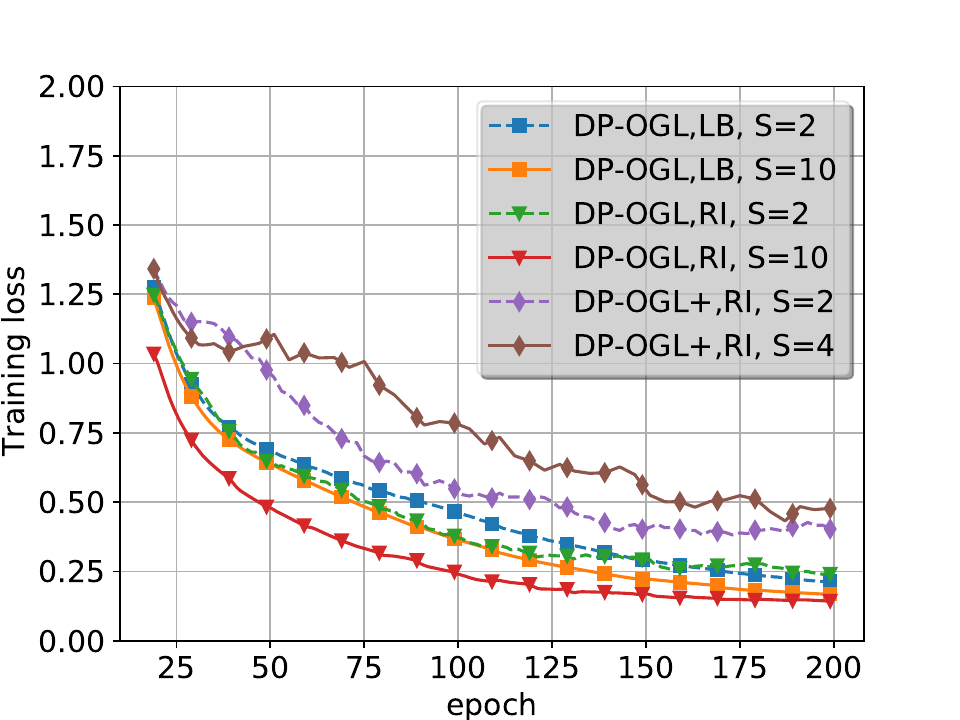
}}
\subfloat[Varied $S$]{\includegraphics[width=0.25\textwidth]{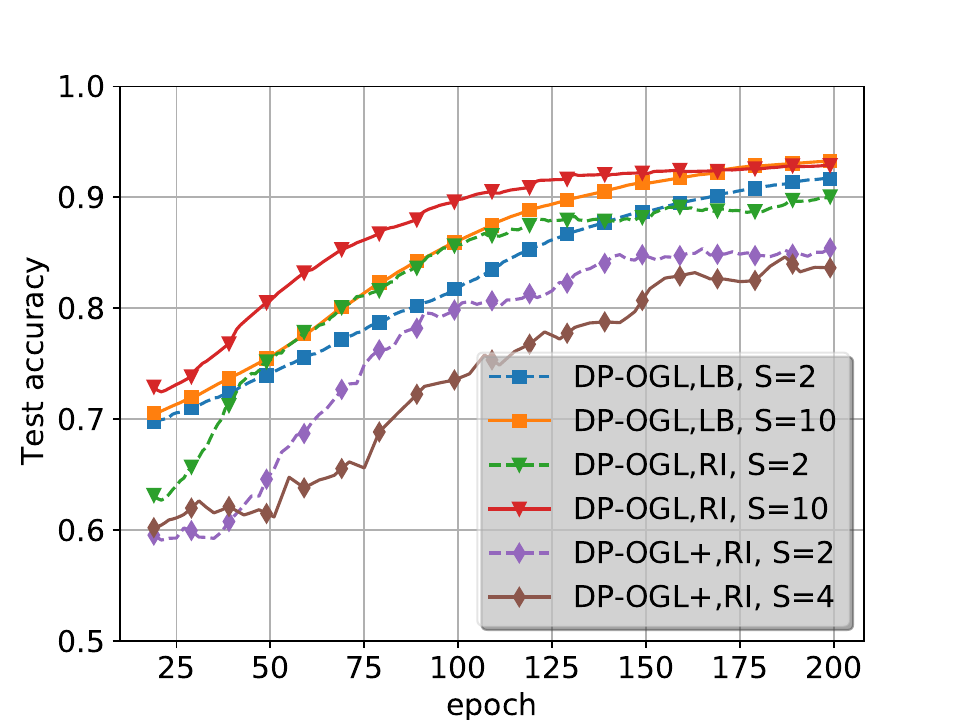
}}
\subfloat[Varied noise]{\includegraphics[width=0.25\textwidth]{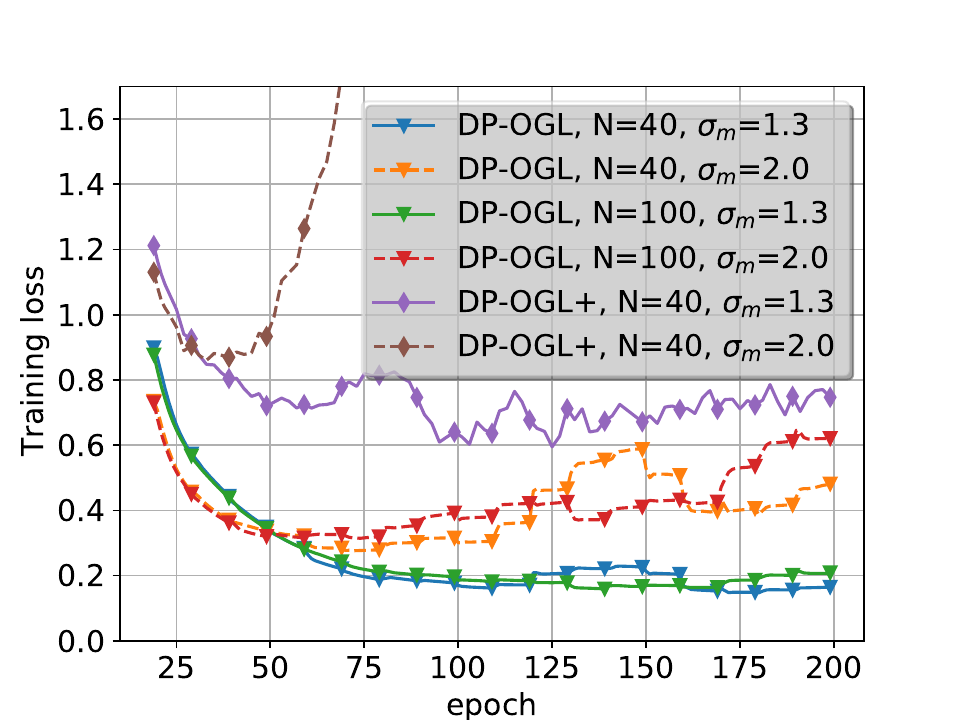
}}
\subfloat[Varied noise]{\includegraphics[width=0.25\textwidth]{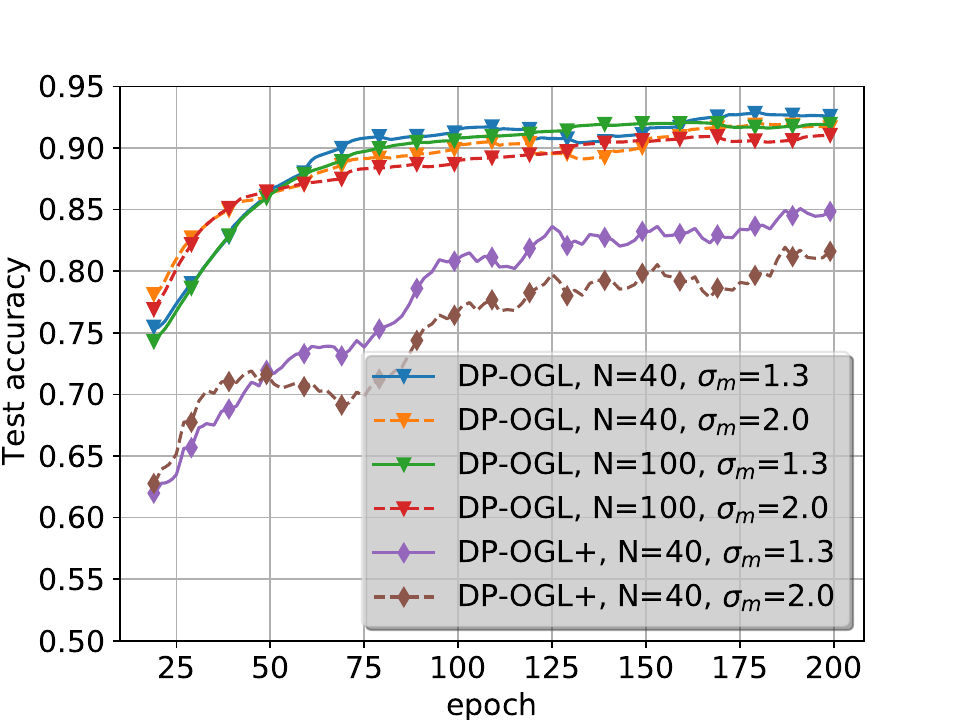
}} 
 \caption{We plot the average training loss vs. epoch in (a), (c), (e), and (g), and the average test accuracy vs. epoch in (b), (d), (f), and (h). In legends, ``LB'', ``CL'', and ``RI'' stand for label-based, clustered, and ring group structures. In (a) and (b), the legends also include four parameters: the number of workers $N$, the number of groups $M$, the value of $S$, and the noise multiplier $\sigma_m$, listed left to right. In (c), (d), (g), and (h), both DP-OGL and DP-OGL+ adopt an RI structure, with $S=10$ for DP-OGL and $S=2$ for DP-OGL+. In (c) and (d), $\sigma_m=2$ is fixed across all curves, except for DP-OGL+ with $N=40$, where $\sigma_m=1.3$. In (g) and (h), $N/M=10$ is fixed. All subfigures maintain $c_m=0.05, \pi_m=0.7,$ and $ L=10$. The baseline corresponds to the GL structure with $M=1$ and $S=1$.}
\label{fig:sim:acc_loss} \vspace{-2ex}
\end{figure*}

\begin{figure*}[ht!]
\centering  \vspace{-3ex}
\subfloat[{\footnotesize PwP $\epsilon$}]{\includegraphics[width=0.25\textwidth]{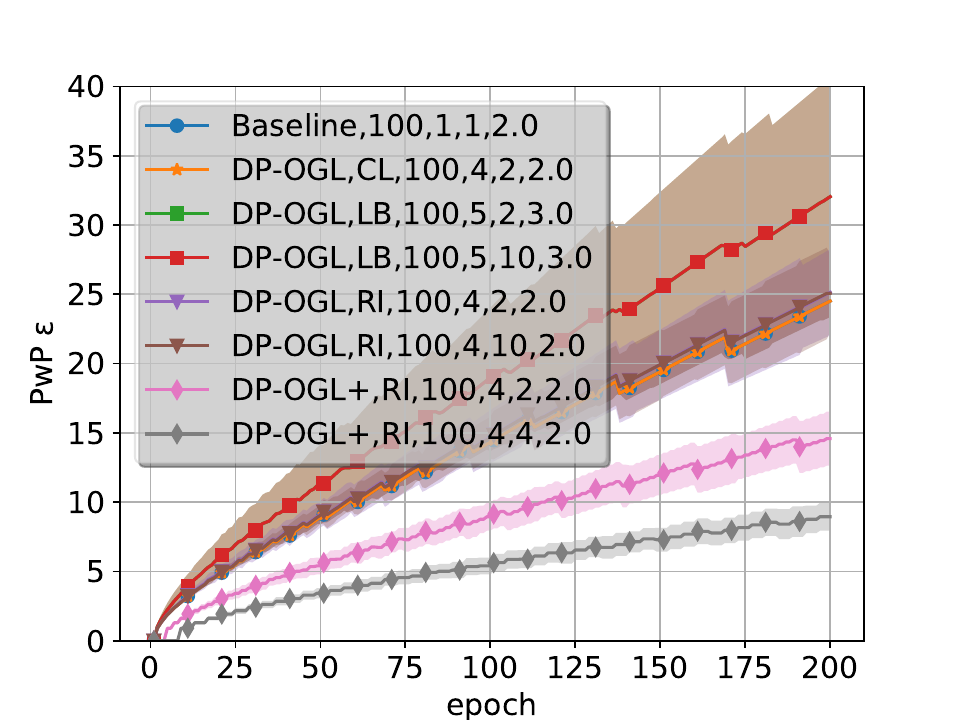
}}
\subfloat[{\footnotesize Baselin, GL $(M,S)=(1,1)$}]{\includegraphics[width=0.25\textwidth]{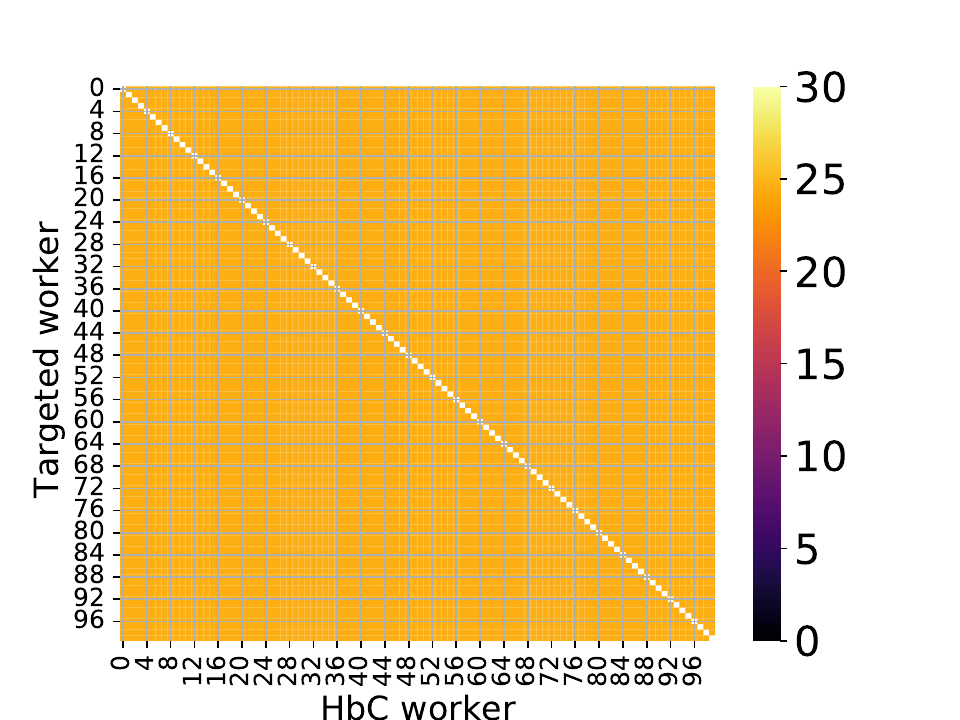
}}
\subfloat[{\footnotesize DP-OGL, CL $(M,S)=(4,2)$}]{\includegraphics[width=0.25\textwidth]{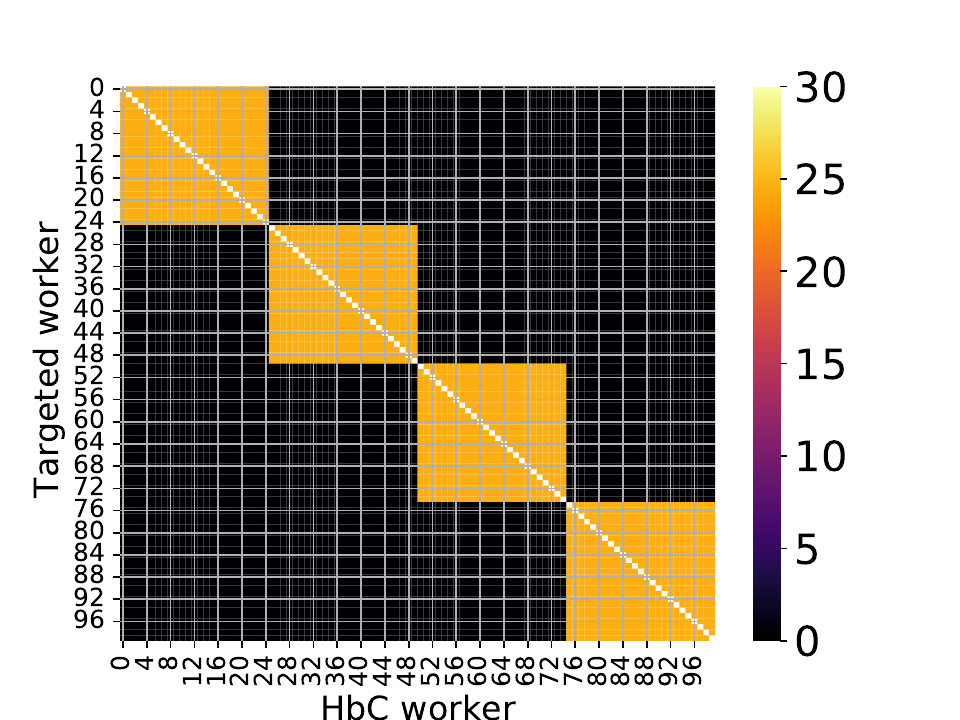
}}
\subfloat[{\footnotesize DP-OGL, LB $(M,S)=(5,10)$}]{\includegraphics[width=0.25\textwidth]{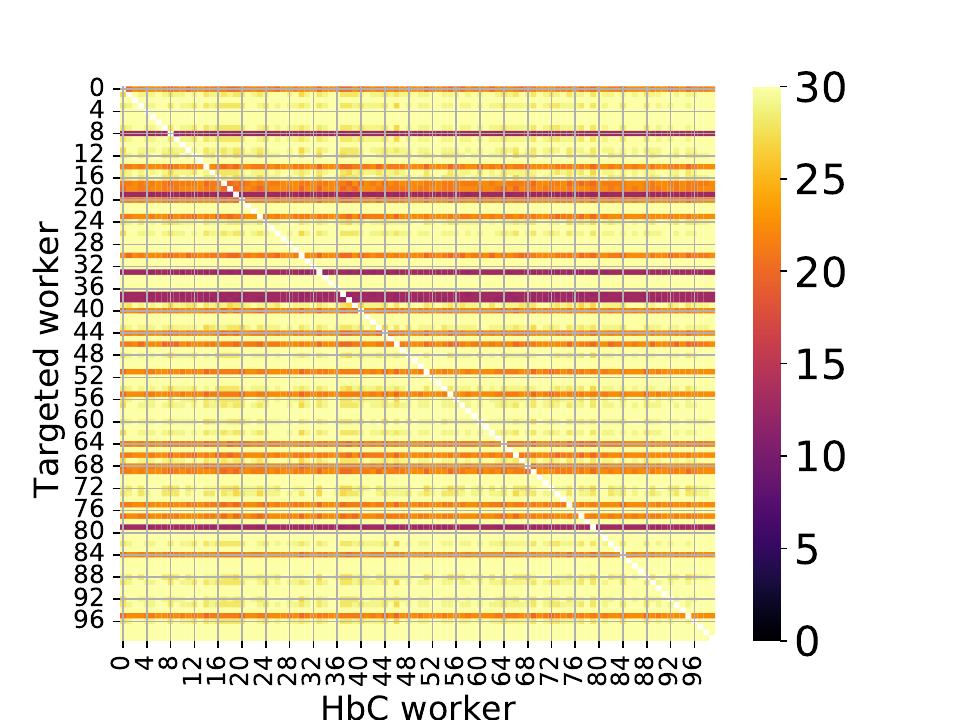
}}

\subfloat[{\footnotesize DP-OGL, RI $(M,S)=(4,10)$}]{\includegraphics[width=0.25\textwidth]{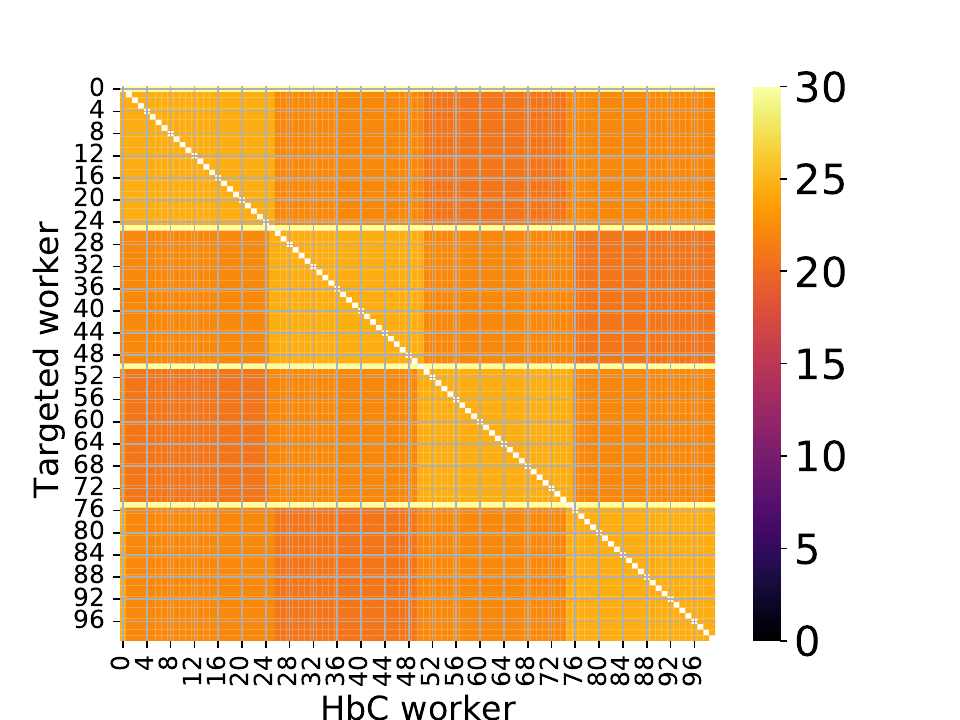
}}
\subfloat[{\footnotesize DP-OGL, RI $(M,S)=(4,25)$}]{\includegraphics[width=0.25\textwidth]{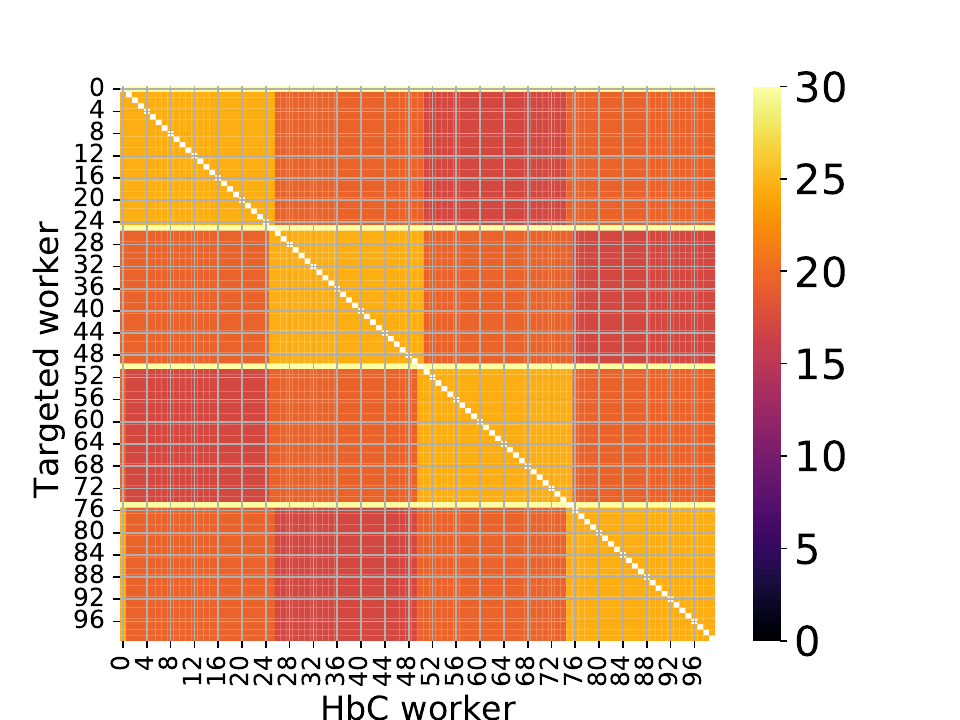
}}
\subfloat[{\footnotesize DP-OGL, RI $(M,S)=(10,10)$}]{\includegraphics[width=0.25\textwidth]{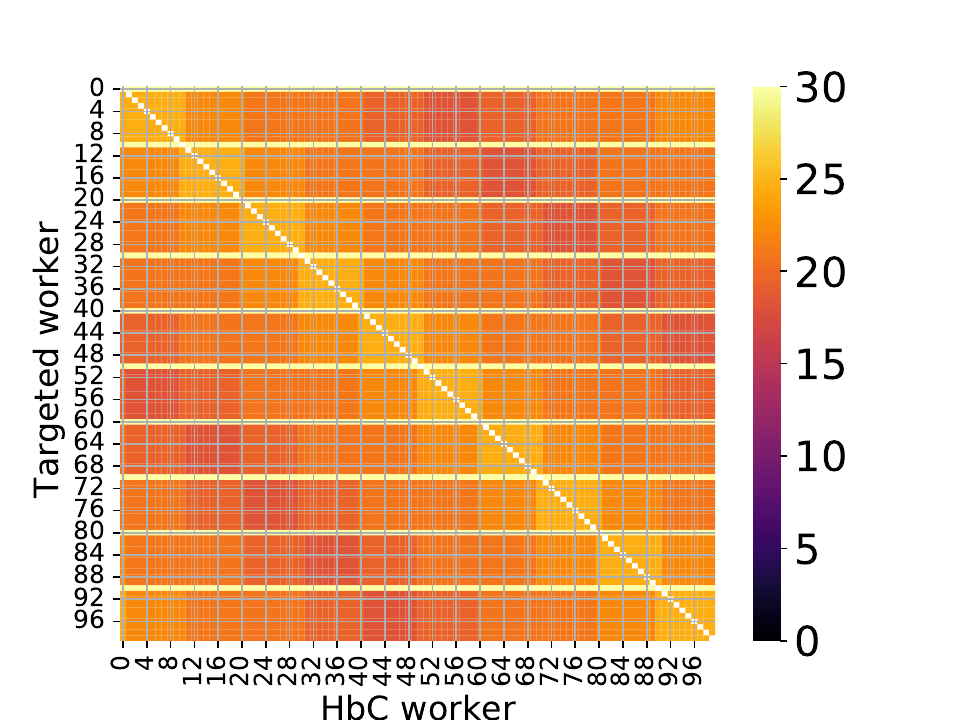
}}
\subfloat[{\footnotesize DP-OGL+, RI $(M,S)=(4,10)$}]{\includegraphics[width=0.25\textwidth]{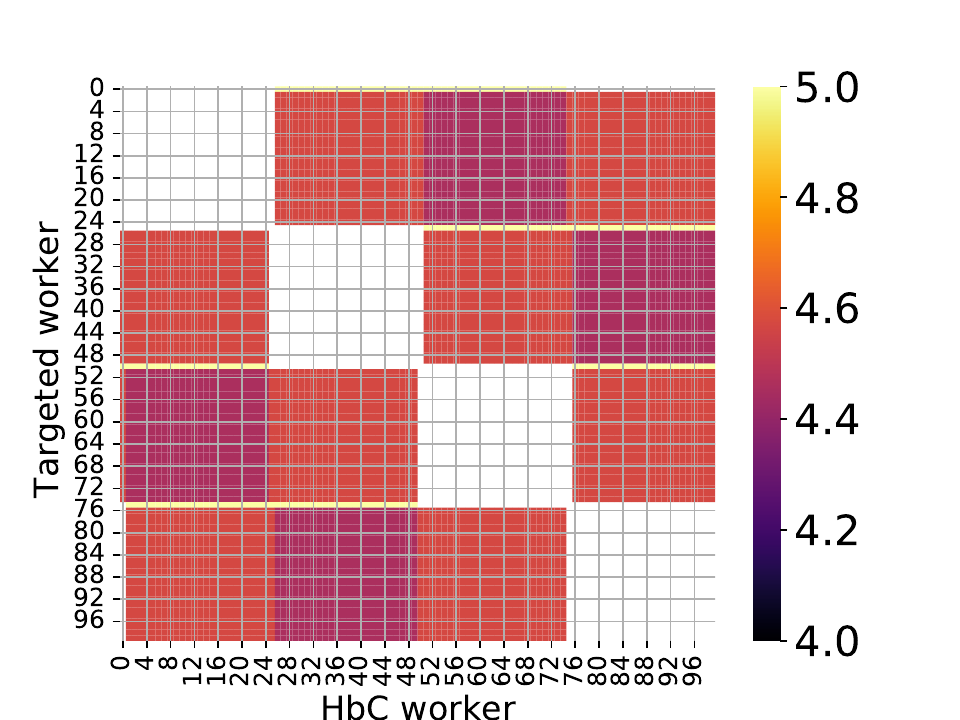
}}
 \caption{PwP bounds are displayed in (a). Heatmaps are displayed for $\epsilon_{n,i}^{1:200}$ across $n,i\in [100]$ under group structures: (b) fully global (GL) with $(M,S)=(1,1)$, (c) clustered (CL) with $(M,S)=(4,2)$, (d) label-based (LB) with $(M,S)=(5,10)$, (e) ring (RI) with $(M,S)=(4,10)$, (f) RI with $(M,S)=(4,25)$, (g) RI with $(M,S)=(10,10)$, and (h) RI with $(M,S)=(4,10)$. In (a)-(g), DP-OGL is run. In (h), DP-OGL+ is run. 
 In heatmaps, darker colors (near 0 in (b)-(g) and near 4 in (h)) indicate lower privacy leakage. Lighter colors (near to 30 in (b)-(g) and near to 5 in (h)) indicate higher leakage. Omitted (white) cells correspond to the pairs of workers with mutual trust.}
\label{fig:sim:heatmaps} \vspace{-2ex}
\end{figure*}

Results for the MNIST dataset are provided below. Due to space constraints, FMNIST dataset's results, which demonstrate the adaptability of our algorithms to other datasets, are presented in App.~\ref{App:sim}. Figure~\ref{fig:sim:acc_loss} shows average training loss and test accuracy versus epoch. Figure~\ref{fig:sim:heatmaps} shows privacy results.

\textbf{Training loss and test accuracy}:
In Figs.~\ref{fig:sim:acc_loss}(a) and (b), we benchmark different grouping structures, while fixing $N=100$. Results for the GL, CL, and LB structures are plotted using circle, star, and square marks, respectively. Results for the RL structures with DP-OGL are plotted by triangular marks and for DP-OGL+ by diamond-shaped marks. For GL, we run the baseline scheme and set $\sigma_m=2$. For CL, we run DP-OGL with $M=4$, $S=2$, and $\sigma_m=2$. For LB, we run again DP-OGL but with $M=5$, $S=10$, and $\sigma_m=3$. For RI, we run both DP-OGL with $S=10$ and DP-OGL+ with $S=2$. In both RI cases, $M=4$ and $\sigma_m=2$.

Figures~\ref{fig:sim:acc_loss}(a) and (b) underscore the advantages of multi-group structures, as they result in lower training loss and higher test accuracy compared to the baseline GL structure with $M=1$. Group structures with $M>1$ enable workers to experience personalized training to some extent due to unique group memberships, resulting in improved convergence performance. As shown in Figs.~\ref{fig:sim:acc_loss}(a) and (b), DP-OGL with the RI structure achieves the lowest training loss and highest test accuracy. The superiority of DP-OGL with  RI compared to LB and CL indicates the importance of optimizing the level of overlaps among groups. The overlap level is minimum (zero) for CL, and it is larger for LB than RI. In these figures, RI outperforms the other two structures. While DP-OGL+ convergence is worse than DP-OGL, for most epochs it is better than the baseline convergence. As we will show later in the section, the advantage of DP-OGL+ lies in its lower privacy bounds compared to both DP-OGL and the baseline.

In Figs.~\ref{fig:sim:acc_loss}(e) and (f), we show the impact of $S$ on the convergence of both DP-OGL and DP-OGL+. With $N=100$ workers fixed, we plot results for DP-OGL using the LB structure with $S=2$ and $S=10$. We also plot results for DP-OGL with the RI structure with the same $S$ values. DP-OGL+ results are plotted for the RI structures with $S=2$ and $S=4$. In these experiments, we set $M=5$ and $\sigma_m=3$ for LB, and $M=4$ and $\sigma_m=2$ for RI. The dashed lines represent the lower $S$ versions of the solid lines with the same marks.

Figures~\ref{fig:sim:acc_loss}(e) and (f) show that increasing $S$ from 2 to 10 enhances DP-OGL convergence. However, the effectiveness of larger $S$ varies across structures. Notably, for LB experiments, the performance difference between $S=2$ and $S=10$ is less pronounced due to significant overlaps among groups, unlike the RI structure where such differences are more evident. In contrast, DP-OGL+ exhibits an inverse relationship with $S$—lower training loss and higher test accuracy are observed with $S=2$ compared to $S=4$. This behavior stems from the summation of the $S$ terms in DP-OGL+ (as of (\ref{eq:method2})) before applying a clipping function. This leads to increased errors and degraded convergence performance with larger $S$. Lastly, for the baseline structure (a single group), $S=1$ is always used, as other values have no impact. Therefore, the baseline curve is not depicted in Figs.~\ref{fig:sim:acc_loss}(e) and (f).

Figures~\ref{fig:sim:acc_loss}(c) and (d) show how $N$, $M$, and more significantly, the $N/M$ ratio affect convergence. We vary $N\in\{4,100\}$ and $M\in\{4,10\}$. In DP-OGL experiments, we plot results for the RI structures with $S=10$. In DP-OGL+, we plot results for RI with $S=2$. Dashed curves correspond to $N/M=10$, while solid lines indicate larger $N/M$ values.

Figures~\ref{fig:sim:acc_loss}(c) and (d) show that both DP-OGL and DP-OGL+ with $N/M=10$ (approximately 11 workers per group) exhibit less favorable performance towards the end of epochs compared to other RI structures with $N/M=25$ and the baseline GL structures with $N/M\in \{40,100\}$. This is attributed to the fact that the sum of worker updates is perturbed by a fixed amount of noise with $\sigma_m=2$ (per (\ref{eq:dpnoise})). This noise perturbation is independent of the group size, thereby hindering convergence when using a small $N/M$ value. As shown in Figs.~\ref{fig:sim:acc_loss}(c) and (d), DP-OGL with $N/M=25$ achieves the best convergence performance towards the last epochs. This highlights the superiority of multi-group structures as $N$ increases while $M$ remains relatively low. Identifying the optimal $N/M$ ratio is an interesting avenue for future work. In the baseline experiments, a larger value of $N$ (and thereby a larger $N/M$) has the opposite impact on the convergence when compared to DP-OGL and DP-OGL+. The baseline experiment with $N=40$ exhibits slightly better performance in some epochs than the baseline with $N=100$. This can be explained by the more personalized models 40 workers can train compared to the scenario where 100 workers collaborate within a Dirichlet-based heterogeneous setting.

In Figs.~\ref{fig:sim:acc_loss}(g) and (h), we show how the noise multiplier $\sigma_m$ affects convergence. Dashed curves correspond to $\sigma_m=2$, while solid lines indicate $\sigma_m=1.3$. Across all curves, we use RI structures and maintain $N/M=10$. As depicted in Figs.~\ref{fig:sim:acc_loss}(g) and (h), reducing $\sigma_m$ from 2 to $1.3$ resolves convergence issues for the RI structures with $N/M=10$, allowing them to achieve better accuracy-loss performance.

\textbf{Per worker privacy}: We compute the average of PwP privacy bounds $\epsilon_{n}^{1:t}$ across workers $n\in [N]$. In Fig.~\ref{fig:sim:heatmaps}(a), we plot the average of $\epsilon_{n}^{1:t}$ vs. epoch $t$. Shaded regions around the curves represent the standard deviation of $\epsilon_{n}^{1:t}$. The specifics of group structure used in Fig.~\ref{fig:sim:heatmaps}(a) match those in Figs.~\ref{fig:sim:acc_loss}(a), (b), (e), and (f). Under Threat Model~\ref{threat1}, Fig.~\ref{fig:sim:heatmaps}(a) shows that among the group structures that use $c_m=0.05$ and $\sigma_m=2.0$, the CI structure (with $M=4$) and the baseline structure (with $M=1$) achieve the strongest average PwP bound. The two RI structures (with $M=4$ and $S\in\{2,10\}$) closely follow and offer the second-best average PwP bounds. On the other hand, the LB structure (with $M=5$ and $S\in\{2,10\}$), even with using a higher noise multiplier $\sigma_m=3.0$, offers a weaker average PwP bound due to more overlaps in its structure. 

Under Threat Model~\ref{threat2}, Fig.~\ref{fig:sim:heatmaps}(a) shows that DP-OGL+ with RI structure ($S=2$) reduces PwP bounds by at least $40\%$ compared to DP-OGL. Increasing $S$ from 2 to 4 further reduces the PwP bound by about $40\%$, showing the benefit of larger $S$. Comparing Fig.~\ref{fig:sim:heatmaps}(a) with Figs.~\ref{fig:sim:acc_loss}(a) and (b) illustrates tradeoffs between PwP bounds and accuracy-loss performance. For example, the baseline GL structure ($M=1$) excels in average PwP but lags behind RI and LB structures in training accuracy and test loss. Figure~\ref{fig:sim:heatmaps}(a) also shows that the CL structure achieves a similar average PwP bound as the baseline structure but outperforms it in convergence. 

\textbf{Heatmaps:} Figures~\ref{fig:sim:heatmaps}(b)-(h) present heatmaps of the privacy bounds $\epsilon_{n,i}^{1:t}$ for different group structures at epoch $t=200$. For the LB structure, we set $(\sigma_m,c_m,\pi_m)=(3.0,0.05,0.7)$, and for others, we set $(\sigma_m,c_m,\pi_m)=(2.0,0.05,0.7)$. Each heatmap in Fig.~\ref{fig:sim:heatmaps} has a vertical y-axis for the targeted worker $n\in [100]$ and an x-axis for the HbC worker $i$. Omitted (white) cells correspond to the pairs of workers with mutual trust. For example, in Figs.~\ref{fig:sim:heatmaps}(b)-(g), $\epsilon_{n,i}^{1:t}$ are defined when $n\neq i$ due to each worker's inherent self-trust.

Comparing the heatmaps of the baseline structure in Fig.~\ref{fig:sim:heatmaps}(b) with that of the CL structure in Fig.~\ref{fig:sim:heatmaps}(c), we observe that the worst-case privacy bounds, which are equivalent to the PwP bounds, match and are colored in orange. However, in CL's heatmap we see black regions that represent zero privacy leakage (i.e., $\epsilon_{n,i}^{1:200}=0$). This highlights the superior privacy of CL compared to the baseline, a distinction not apparent in the PwP bounds in Fig.~\ref{fig:sim:heatmaps}(a). 

The heatmaps in Figs.~\ref{fig:sim:heatmaps}(d)-(h) correspond to the group structures that contain overlapping groups. Specifically, Fig.~\ref{fig:sim:heatmaps}(d) depicts the heatmap of the LB structure with $M=5$ and $S=10$. Figures~\ref{fig:sim:heatmaps}(e)-(h) depict the heatmaps of the RI structures with $M\in \{4,10\}$ and $S\in \{10,25\}$. Figure~\ref{fig:sim:heatmaps}(d) exhibits a scattered pattern, with many cells in the heatmap displaying larger $\epsilon_{n,i}^{1:200}$ compared to the baseline in Fig.~\ref{fig:sim:heatmaps}(b). However, certain cells in this heatmap in Fig.~\ref{fig:sim:heatmaps}(d) offer lower $\epsilon_{n,i}^{1:200}$ values than those of the baseline. These cells correspond to the pair of targeted and HbC workers who are further from each other when compared to the baseline.

We now compare Figs.~\ref{fig:sim:heatmaps}(e)-(h). When $S=10$ in Fig.~\ref{fig:sim:heatmaps}(e) increases into $S=25$ in Fig.~\ref{fig:sim:heatmaps}(f), the darker regions in the heatmap that correspond to better (lower) privacy bounds appear. This occurs because as we increase $S$, the privacy leakage faces more propagation delay. Thereby, the privacy leakage to distant HbC workers decreases at a given epoch 200. Furthermore, as we increase the number of groups from $M=4$ in Fig.~\ref{fig:sim:heatmaps}(e) to $M=10$ in Fig.~\ref{fig:sim:heatmaps}(g), propagation delay further grows. Comparing Figs.~\ref{fig:sim:heatmaps}(e) and (g), we see that a larger value of $M$ in the RI structure results in better (lower) privacy bounds in specific regions. The colors in these regions tend to shift towards red or become darker as the HbC workers locate in further groups than the targeted workers. However, we note that the worst-case bounds among $\epsilon_{n,i}^{1:200}$ (representing the PwP leakage) are depicted in orange and slightly exceed the bounds of the baseline structure.

In DP-OGL+'s heatmap in Fig.~\ref{fig:sim:heatmaps}(h), white regions represent full privacy leakage to in-group HbC nodes. Excluding these regions, cell colors in this heatmap are scaled down by around $\frac{1}{S}=0.1$ compared to the corresponding cells in Fig.~\ref{fig:sim:heatmaps}(e). This is due to the improvement we show in Thm.~\ref{thm:scenario1} when comparing (\ref{eq:theorem}) with (\ref{eq:theorem+}). However, this comes at the cost of a convergence delay in DP-OGL+, as is shown in Fig.~\ref{fig:sim:acc_loss}.

\section{Conclusion and Future Work}\label{sec6}
We propose the novel DP-OGL and DP-OGL+ algorithms that apply DP to collaborative learning with overlapping group structures. These algorithms run periodic inter-group epochs, where workers ``mix'' information about their respective groups into one model parameter. In between consecutive inter-group epochs, multiple intra-group epochs are operated that confine information within each group. The longer the gap between these inter-group epochs, the more it delays privacy leakage to workers in other groups. DP-OGL controls privacy leakage propagation between any worker pair, while DPOGL+ controls stronger privacy leakage propagation w.r.t. out-of-group HbC nodes. In this setup, we provide a tight privacy analysis that quantifies privacy leakage propagation across overlapping groups between arbitrary pairs of workers instead of having one single privacy bound over all workers. In our analysis, we discover and characterize two effects. First, we establish guarantees that consider \textit{propagation delay}, showing an inverse relationship between privacy leakage and the distance between worker pairs. Secondly, by incorporating the \textit{information degradation} effect into our analysis, we are able to further strengthen privacy guarantees through noise addition to intermediate model updates. Our theoretical and numerical analysis highlights the individual per-pair privacy guarantees among the HbC workers who may have multiple group memberships at a time.

Our paper suggests promising future research directions. The first is broadening information-degradation analysis beyond string-like group structures. The second is expanding the privacy leakage propagation analysis for dynamically evolving group structures. The third is enhancing worker scheduling within DP-OGL by exploring scenarios where workers strategically engage in various groups, rather than participating randomly and independently. Future research can also explore ways to optimize groups' privacy parameters and mixing coefficients based on trust levels within and across groups.

\section{Acknowledgments}
We thank our sponsors. We would also acknowledge helpful feedback from and discussions with Dr. Haider AL-Lawati. 

\bibliographystyle{IEEEtran} 

\bibliography{literature}

\appendix
\section{Appendix}\label{sec:appendix}

\subsection{Notations}\label{app:notation}
We use the calligraphic font to denote sets, e.g., $\calD, \calN$, and $\calM$. The cardinality of a finite set $\calD$ is denoted $|\calD|$. We also use the notation $[n]=\{1,2,\ldots,n\}$. Vectors are denoted using bold lowercase, e.g., $\mathbf{x}$. We use bold upper case to denote randomized mechanisms, e.g., $\mathbf{F}$ and $\mathbf{G}$. For any sample $x$ drawn from distribution $\distr$ (represented as $x\sim \distr$), we denote the probability density function (PDF) of $x$ as $\distr(x)$. The distribution of a random variable $X$ is represented as $\distr_X$.  Table~\ref{app:summary_table} summarizes all important letters used in this paper. 
\begin{table*}[ht]
\centering
\begin{threeparttable}
\caption{A Summary of Notations\tnote{1}}
\centering\label{app:summary_table}
\begin{tabular}{|c|l||c|c||c|c|}
\hline
$N$ & No. of Wr. & $\calD$ & Datapool & $\lossfunc$ & Loss function of Wr. $n$ 
\\ \hline
$\calN$ & Wrs. & $\calD_n$ & Dataset of Wr. $n$ & $\func$ & Expected loss of Wr. $n$ 
\\ \hline
$\calN_m$ & Wrs. in Gr. $m$ & $\dsetelement$ & Data sample & $\emfunc$ & Empirical loss of Wr. $n$ 
\\ \hline
$\calW_m^t$ & Wrs. in Gr. $m$, E. $t$ & $\mathbf{x}$ & Feature vector & $\lrloc$ & Learning rate 
\\ \hline
$M$ & No. of Gr. & $y$ & Scalar label & $\alpha$ & Renyi divergence order \\ \hline
$\calM$ & Grs. & $\wdistr_n$ & Samp. Dist. of Wr. $n$ & $\epsilon_{n,i}^{1:t}$ & (Wr. $n$, HbC $i$)-PB in E. $t$ \\ \hline
$\calM_n$ & Grs. of Wr. $n$ & $\calB_n^{t,l}$ &   MB of Wr. $n$, I. $l$, E. $t$ & $\epsilon_{n}^{1:t}$ & PwPB of Wr. $n$, E. $t$
\\ \hline
$T$ & No. of E. & $\model_m^{t}$ & Model of Gr. $m$, E. $t$ & $\beta$ & Smoothness degree of $\lossfunc$ 
\\ \hline
$L$ &  No. of Local I. & $\mper_n^{t,l}$ & Model of Wr. $n$, I. $l$, E. $t$ & $c_m$ & Clipping Par.
\\ \hline
$\pi_m$ & Poi. Par. in Gr. $m$ & $\Delta\model_m^t$ & $\pi_m|\calN_m|(\model_m^t-\model_m^{t-1})$ & $\sigma_m$ & Noise multiplier \\ 
\hline
$\groupdist_{m,m'}$ & Dis. of Gr. $m$, $m'$ & $\Delta\mper_n^t$ & $\mper_n^{t,L}-\mper_n^{t,0}$ & $\randfunc_i^{1:t}$ & Wr. $i$ Mech. in E. $t$ \\ 
\hline
\end{tabular}
 \begin{tablenotes}
	\item[1] Table's abbreviations: ``No.'' for ``Number'', ``Wr.'' for ``Worker'', ``Wrs.'' for ``Set of workers'', ``Gr.'' for ``Group'', ``Grs.'' for ``Set of groups'', ``E.'' for ``Epochs'', ``I.'' for ``Iteration(s)'', ``Samp.'' for ``Sampling'', ``Poi.'' for ``Poisson'', ``Dis.'' for ``Distance'', ``Dist.'' for ``Distribution'',  ``MB'' for ``Mini Batch'', ``Par.'' for ``Parameter'', ``Agg.'' for ``Aggregate'', ``PB.'' for ``Privacy Bound'', and ``PwPB'' for ``Per-worker privacy Bound''.     
   \end{tablenotes}
\end{threeparttable}\vspace{-3ex}
\end{table*}

\subsection{Preliminaries}\label{app:preliminaries}

We next restate some definitions from~\cite{mironov2017renyi},~\cite{mironov2019r}, and~\cite{vempala2019rapid}. We also introduce the concept of worker-level $L_2$-sensitivity, building upon the definition of $L_2$-sensitivity.



\begin{definition}[Renyi DP~\cite{mironov2017renyi}]\label{def:RDP}
A random function $\randfunc: \chi\rightarrow \mathbb{R}^w$ is $(\alpha,\epsilon)$-Renyi DP (RDP) if for any neighboring inputs ${\calX}, {\calX'} \subseteq \chi$, $R_{\alpha}(\distr_{\randfunc({\calX})} \| \distr_{\randfunc({\calX'})}) \leq \epsilon$.
\end{definition}

\begin{definition}[$L_2$-sensitivity~\cite{mironov2017renyi}]
The $L_2$-sensitivity of a function $f: \chi\rightarrow \mathbb{R}^w$ given any input datasets $\calX,{\calX}'\subseteq \chi$ with dissimilarity $\left| \calX\cup \calX  \right| - \left| \calX\cap\calX'\right|=1$ is $\max_{{\calX},{\calX}',\; \datadist(\calX,\calX')=1} \left\|f({\calX})-f({\calX}') \right\|_2$.
\end{definition}

\begin{definition}[Worker-level $L_2$ sensitivity]
In a system with $N$ workers, the worker-level $L_2$-sensitivity of a function $f: \chi\rightarrow \mathbb{R}^w$ is defined for neighboring input data pools $\calX,{\calX}'\subseteq \chi$ (where $\calX = \bigcup_{n\in [N]} \calX_n$ and $\calX'= \bigcup_{n\in [N]} \calX_n'$) as $\max_{\text{neighboring }{\calX},{\calX}'} \left\|f({\calX})-f({\calX}') \right\|_2$.
\end{definition}

\begin{definition}[Gaussian Mechanism~\cite{mironov2017renyi}]
The Gaussian mechanism for function $f: \chi\rightarrow \mathbb{R}^w$ with $L_2$-sensitivity $c$ and input $\calX\subseteq \chi$ is $\randgauss_{\sigma, c}(\calX) := f(\calX) + \calN (0,c^2\sigma^2\calI_w)$. 
\end{definition}

\begin{definition}[Sampled Gaussian Mechanism~\cite{mironov2019r}]
The function $f_q$, parameterized by $q \in [0,1]$, maps a subset of the input set $\calX\in\chi$ to $\mathbb{R}^w$, with elements in $\calX$ included with probability $q$. Applying the Gaussian mechanism to $f_q$ with $L_2$-sensitivity  $c$, the sampled Gaussian mechanism is 
\begin{align}
\randgauss_{\sigma, c, q}(\calX) &:= f_q(\{x \;| \; x\in \calX \text{ is sampled with Prob. } q \}) \nonumber \\ &+ \calN (0,c^2\sigma^2\calI_w).
\end{align}
\end{definition}

\begin{definition}[Log-Sobolev inequality (LSI)~\cite{vempala2019rapid}]\label{app:def:LSI}
A distribution $Q$ over $\mathbb{R}^{w}$ satisfies the LSI with constant $\lsicont$ if for all smooth functions $g: \mathbb{R}^w \rightarrow \mathbb{R}$ with $\mathbb{E}_{q\sim Q}(g^2(q)) < \infty$,
\begin{align*}
&\mathbb{E}_{q\sim Q}\left(g^2(q)\log g^2(q) \right) - \mathbb{E}_{q\sim Q} \left(g^2(q)\right) \log \mathbb{E}_{q\sim Q} (g^2(q)) \\ & \leq \frac{2}{\lsicont} \mathbb{E}_{q\sim Q} \left( \|\nabla g(q) \|^2 \right).
\end{align*} 
\end{definition}

We next recall a few lemmas from~\cite{mironov2017renyi},~\cite{mironov2019r}, and~\cite{vempala2019rapid} that will prove useful for our analysis. 

\begin{lem}[\cite{mironov2017renyi}]\label{lem:rdpGM} 
The Gaussian mechanism $\randgauss_{\sigma,c}$ guarantees $(\alpha, \epsilon)$-RDP with $\epsilon = \frac{\alpha}{2\sigma^2}$.
\end{lem}

\begin{lem}[\cite{mironov2019r}]\label{lem:rdpsGM} 
The sampled Gaussian mechanism $\randgauss_{\sigma,c,q}$ guarantees $(\alpha, \frac{2q^2\alpha}{\sigma^2})$-RDP.
\end{lem}

\begin{lem}[Post-processing~\cite{mironov2017renyi}]\label{lem:postproc} 
Let the randomized mechanism $\randfunc:\chi\rightarrow \calR_1$ be $(\alpha,\epsilon)$-RDP and consider a mapping $g:\calR_1\rightarrow \calR_2$. By the analogue of the data processing inequality, $g(\randfunc_1(.))$ also satisfies $(\alpha,\epsilon)$-RDP.   
\end{lem}

\begin{lem}[Sequential composition~\cite{mironov2017renyi}]\label{lem:composition}
Let the mechanism $\randfunc_1:\chi\rightarrow \calR_1$ be $(\alpha,\epsilon_1)$-RDP and $\randfunc_2:\calR_1\times \chi\rightarrow \calR_2$ be $(\alpha,\epsilon_2)$-RDP. The joint mechanism $(\randfunc_1,\randfunc_2)$ with $Y_1\sim \distr_{\randfunc_1(\calX)}$ and $Y_2\sim \distr_{\randfunc_2(Y_1,\calX)}$ satisfies $(\alpha, \epsilon_1+\epsilon_2)$-RDP.
\end{lem}

\begin{lem}[Joint convexity~\cite{ye2022differentially}]\label{lem:jointconvexity}
Let $\randfunc_1,\ldots,\randfunc_n$ be randomized mechanisms where $\randfunc_i: \chi\rightarrow \calR$. Then for any $\alpha\geq 1$, $\calX,\calX'\in \chi$, and $a_1,\ldots,a_n\geq 0$ that satisfies $\sum_{i=1}^na_i=1$,
\begin{align}
&e^{(\alpha-1)R_{\alpha}\left(\sum_{i=1}^na_i\randfunc_i(\calX)\| \sum_{i=1}^n a_i\randfunc_i(\calX')\right)} \nonumber  \\ & \leq \sum_{i=1}^n a_ie^{(\alpha-1)R_{\alpha}(\randfunc_i(\calX)\| \randfunc_i(\calX'))}. 
\end{align}
\end{lem}

\begin{lem}[$\lipcont$-Lipschitz mapping~\cite{vempala2019rapid}]\label{app:lem:Lipschitzmap} If there exists  a probability distribution $Q$ over $\mathbb{R}^w$ that satisfies the LSI with a constant $\lsicont>0$, and if $g:\mathbb{R}^w\rightarrow \mathbb{R}$ is a differentiable $\lipcont$-Lipschitz mapping function, then the distribution $g(Q)$ also satisfies the LSI with a constant of $\frac{\lsicont}{\lipcont^2}$.
\end{lem}

\begin{lem}[Gaussian convolution mapping~\cite{vempala2019rapid}]\label{app:lem:gaussmap} For a probability distribution $Q$ over $\mathbb{R}^w$ that satisfies the LSI with constant $\lsicont>0$, the distribution $Q\ast \calN(0,\sigma^2 \calI_w)$ satisfies the LSI with constant of $\left(\frac{1}{\lsicont}+\sigma^2\right)^{-1}$.
\end{lem}

\begin{lem}[Convolution mapping~\cite{vempala2019rapid}]\label{app:lem:convmap} For distributions $Q_1$ and $Q_2$ over $\mathbb{R}^w$ that both satisfy the LSI with, respectively, constants $\lsicont_1>0$ and $\lsicont_2>0$, the distribution $Q_1\ast Q_2$ satisfies the LSI with constant $\left(\frac{1}{\lsicont_1}+\frac{1}{\lsicont_2}\right)^{-1}$.
\end{lem}

\begin{lem}\label{app:lem:moreinfo} Let $\alpha>1$. Consider joint random variable $(X,Y)\sim \distr_{X,Y}$ and $(X',Y')\sim \distr_{X',Y'}$, where $X\sim \distr_X$, $X'\sim \distr_{X'}$, $Y\sim \distr_{Y|X}$, and $Y'\sim \distr_{Y'|X'}$. Assuming $X$ and $X'$ are over an output space $\calR_1$, and $Y$ and $Y'$ are over $\calR_2$, 
\begin{align}
R_{\alpha}\left(\distr_{X,Y}\| \distr_{X',Y'}\right) \geq R_{\alpha}\left(\distr_{X}\| \distr_{X'}\right).
\end{align} 

\textit{Proof:} By Renyi divergence definition, we have
\begin{align*}
&\exp\left((\alpha-1)R_{\alpha}\left(\distr_{X,Y}\| \distr_{X',Y'}\right)\right) 
\\ &= \int_{\calR_1, \calR_2} \distr_{X}(x)^{\alpha}\distr_{X'}(x)^{1-\alpha} 
\distr_{Y|X}(x,y)^{\alpha}\distr_{Y'|X'}(x,y)^{1-\alpha} dy dx \nonumber
\\ & = \int_{\calR_1}  \distr_{X}(x)^{\alpha}\distr_{X'}(x)^{1-\alpha}
\int_{\calR_2} \distr_{Y|X}(y|x)^{\alpha}\distr_{Y'|X'}(y|x)^{1-\alpha} dy dx 
\\ &= \int_{\calR_1} \distr_{X}(x)^{\alpha}\distr_{X'}(x)^{1-\alpha} 
 e^{(\alpha-1)R_{\alpha}\left(\distr_{Y|X=x} \| \distr_{Y'| X'=x} \right)} dx 
\\ & \geq  \int_{\calR_1} \distr_{X}(x)^{\alpha}\distr_{X'}(x)^{1-\alpha} \exp \left( (\alpha-1)0 \right) dx
\\ & =  \int_{\calR_1} \distr_{X}(x)^{\alpha}\distr_{X'}(x)^{1-\alpha} dx  = e^{(\alpha-1)R_{\alpha}\left(\distr_{X}\| \distr_{X'}\right)}.\; \blacksquare
\end{align*}

\end{lem}

\subsection{Layered Composition Lemma}\label{app:lem:gcomposition}
We consider a sequence of two sets of mechanisms. Each set is referred to as a layer. In Lem.~\ref{lem:gcomposition}, we extend Lem.~\ref{lem:composition} from App.~\ref{app:preliminaries} to layers. Within each layer, multiple mechanisms operate in parallel, with the layers executing sequentially. For $i\in [2]$, the $i$th layer contains $K_i$ mechanisms. We use $\calK_i$ to denote the set $[K_i]$. These two mechanism sets operate on a data pool $\calX \in \chi$. We also use $\randfunc_{k}^{(\calK_i)}$ to denote the $k$th mechanism in the $i$th layer, for  $i\in [2]$ and $k\in \calK_i$. For each $k\in \calK_1$, $\randfunc_{k}^{(\calK_{1})}:\chi \rightarrow \calR_{1,k}$. For each $k\in \calK_2$, $\randfunc_{k}^{(\calK_2)}: \chi  \times\left(\prod_{i\in \calH_{k}} \calR_{1,i}\right) \rightarrow \calR_{2,k}$. Mechanisms in the second layer depend on outcomes from a subset of mechanisms in the first layer. This subset is denoted as $\calH_{k}\subseteq \calK_1$. We define $\randfunc^{(\calK_1)}:=\left(\bigcup_{{k\in \calK_1}} \randfunc_{k}^{(\calK_1)}\right)$, $\randfunc^{(\calK_2)}:=\left(\bigcup_{k\in \calK_2} \randfunc_{k}^{(\calK_2)}\right)$, and $\randfunc^{(\calK_1, \calK_2)}:=\left(\randfunc^{(\calK_1)}, \randfunc^{(\calK_2)}\right)$. Given $\calX \in \chi$, $\randfunc^{(\calK_1, \calK_1)}$ outputs $\left(\{z_{1,k}\}_{k\in \calK_1}, \{z_{2,j}\}_{j\in \calK_2}\right)$, where $z_{1, k}\sim \distr_{\randfunc_{k}^{(\calK_1)}(\calX)}$ and $z_{2,j}\sim \distr_{\randfunc_{j}^{(\calK_2)}(\calX, \{z_{1, i}\}_{i\in \calH_{j}})}$. 

\begin{lem}[Layered composition]\label{lem:gcomposition}
Consider $\randfunc_{k}^{(\calK_1)}:\chi\rightarrow \calR_{1,k}$, for $k\in \calK_1$. Let $\alpha>1$, $\bigcup_{j\in \calK_2}\calH_j=\calK_1$, and $\randfunc_{j}^{(\calK_2)}:\left(\prod_{i\in \calH_{j}} \calR_{1,i}\right)\times \chi\rightarrow \calR_{2,j}$ be  $(\alpha,\epsilon_{2,j})$-RDP,  for $j\in \calK_2$. For neighboring data pools $\calX, \calX' \in \chi$, $\randfunc^{(\calK_1, \calK_2)}$ satisfies
\begin{align}\label{app:eq:layer_composition}
& R_{\alpha}\left(\distr_{\randfunc^{(\calK_1, \calK_2)}(\calX)} \|  \distr_{\randfunc^{(\calK_1, \calK_2)}(\calX')} \right) \nonumber \\
& \leq \sum_{j\in \calK_2} \epsilon_{2,j} + R_{\alpha}\left(\distr_{\randfunc^{(\calK_1)}(\calX)} \|  \distr_{\randfunc^{(\calK_1)}(\calX')} \right).
\end{align}

\textit{Proof:} We use $Z_{1,k}, Z_{2,j}$,  $Z$, $Z_{1,k}', Z_{2,j}'$, and $Z'$ to denote, respectively, the distributions $\distr_{\randfunc_k^{(\calK_1)}(\calX)}$, $\distr_{\randfunc_j^{(\calK_2)}(\calX)}$,  $\distr_{\randfunc^{(\calK_1, \calK_1)}(\calX)}$, $\distr_{\randfunc_k^{(\calK_1)}(\calX')}$, $\distr_{\randfunc_j^{(\calK_2)}(\calX')}$, and $\distr_{\randfunc^{(\calK_1, \calK_1)}(\calX')}$. By the Renyi divergence definition, we have
\begin{align}
&\exp \left((\alpha-1)R_{\alpha}\left(Z \| Z')\right)\right) 
\nonumber \\ 
&= \int_{\prod_{k\in\calK_1} \calR_{1,k} \times \prod_{j\in\calK_2}\calR_{2,j}} \left[ Z(\{z_{1,k}\}_{k\in \calK_1}, \{z_{2,j}\}_{j\in \calK_2})^{\alpha} \right. 
\nonumber \\ &  \left. \times Z'(\{z_{1,k}\}_{k\in \calK_1}, \{z_{2,j}\}_{j\in \calK_2})^{1-\alpha}dz_{1,1}\ldots dz_{2,K_2}\right].
\end{align}

The joint PDF of the distribution $Z$ over $\left(\{z_{1,k}\}_{k\in \calK_1}, \{z_{2,j}\}_{j\in \calK_2}\right)$ factors into the individual PDFs of $\{Z_{1,k}\}_{k\in \calK_1}$ over $\{z_{1,k}\}_{k\in \calK_1}$ for $\{z_{1,k}\}_{k\in \calK_1}$, and conditioned PDFs of $\{ Z_{2,j}\}_{j\in \calK_2}$ over $\{z_{2,J}\}_{J\in \calK_2}$ where $ Z_{2,j}$ is conditioned on $\{z_{1,i}\}_{i\in \calH_j}$. Similarly, for $Z'$, the joint PDF factors into PDFs of $\{Z'_{1,k}\}_{k\in \calK_1} $ and $ \{Z'_{2,j}\}_{j\in \calK_2}$. Thus,
\begin{align}
&\exp \left((\alpha-1)R_{\alpha}\left(Z \| Z'\right)\right) 
\nonumber\\ 
&= \int_{\prod_{k\in \calK_1} \calR_{1,k} \times \prod_{j\in \calK_2} \calR_{2,j}} \left[ \prod_{k\in \calK_1} Z_{1,k}(z_{1,k})^{\alpha} \times \right. 
\nonumber \\ & \left.  \prod_{j\in \calK_2} Z_{2,j}(\{z_{1,i}\}_{i\in \calH_k},z_{2,j})^{\alpha} \prod_{k\in \calK_1} Z'_{1,k}(z_{1,k})^{1-\alpha}\times \right. 
\nonumber \\ &  \left. \prod_{j\in \calK_2} Z'_{2,j}(\{z_{1,i}\}_{i\in \calH_j}, z_{2,j})^{1-\alpha}dz_{1,1}\ldots dz_{2,K_2}\right].
\end{align}

By rearranging the nested integrals, we obtain
\begin{align}
&\exp \left((\alpha-1)R_{\alpha}\left(Z \| Z'\right)\right) 
\nonumber \\ 
&= \int_{\prod_{k\in\calK_1} \calR_{1,k}}  \prod_{k\in \calK_1}\left[ Z_{1,k}(z_{1,k})^{\alpha}Z'_{1,k}(z_{1,k})^{1-\alpha}\right] 
\nonumber \\ & \int_{\prod_{j\in\calK_2} \calR_{2,j}} \prod_{j\in \calK_2} \left[Z_{2,j}(\{z_{1,i}\}_{i\in \calH_j},z_{2,j})^{\alpha}\times \right. 
\nonumber \\ & \left. Z'_{2,j}(\{z_{1,i}\}_{i\in \calH_j}, z_{2,j})^{1-\alpha}\right]dz_{2,1}\ldots dz_{2,K_2} \ldots dz_{1,K_1}
\\ &= \int_{\calR_{1,K_1}}\ldots \int_{\calR_{1,1}} \left[ Z_{1,1}(z_{1,1})^{\alpha}Z'_{1,1}(z_{1,1})^{1-\alpha}\right] 
\nonumber \\ &\int_{\calR_{2,K_2}} \ldots \int_{\calR_{2,1}} \left[ Z_{2,1}(\{z_{1,i}\}_{i\in \calH_1},z_{2,1})^{\alpha}\times \right.
\nonumber \\ &  \left. Z'_{2,1}(\{z_{1,i}\}_{i\in \calH_1}, z_{2,1})^{1-\alpha}\right]dz_{2,1}\ldots dz_{1,K_1}.
\end{align}
Since $\randfunc_{j}^{(\calK_{2})}$ is $(\alpha,\epsilon_{2,j})$-RDP for every $j\in \calK_2$, we obtain
\begin{align}
&\exp \left((\alpha-1)R_{\alpha}\left(Z \| Z'\right)\right) 
\nonumber \\ 
& \leq \int_{\prod_{k\in\calK_1} \calR_{1,k}} \left[ \prod_{k\in \calK_1}\left[ Z_{1,k}(z_{1,k})^{\alpha}Z'_{1,k}(z_{1,k})^{1-\alpha}\right] \right. 
\nonumber \\ &\left. \prod_{j\in \calK_2}e^{(\alpha-1)\epsilon_{2,j}}  dz_{1,1}\ldots dz_{1,K_1}\right] 
\\ &= \left(\prod_{j\in \calK_2} e^{(\alpha-1)\epsilon_{2,j}} \right)
e^{(\alpha-1) R_{\alpha}\left(\distr_{\randfunc^{(\calK_1)}(\calX)} \|  \distr_{\randfunc^{(\calK_1)}(\calX')} \right)} 
\\ & = e^{(\alpha-1)\left(\sum_{j\in \calK_2}\epsilon_{2,j}+ R_{\alpha}\left(\distr_{\randfunc^{(\calK_1)}(\calX)} \|  \distr_{\randfunc^{(\calK_1)}(\calX')} \right)\right)}. 
\end{align}
Since $\alpha>1$, the above inequality concludes (\ref{app:eq:layer_composition}). $\blacksquare$
\end{lem}

\subsection{Proof of Thm.~\ref{thm:scenario1}} \label{sec:thm:scenario1}
\subsubsection{Proof of Thm.~\ref{thm:scenario1} for DP-OGL under the Threat Model~\ref{threat1}}\label{sec:thm:scenario1_1}
We solve the theorem by monitoring the contributions of groups to $\randfunc_i^{1:t}$ during the first $t$ epochs. Based on whether a group that contributes to $\randfunc_i^{1:t}$ includes the targeted worker $n$ or not, we either augment $\epsilon_{n,i}^{1:t}$ by adding the privacy budget of the respective group or regard that group's mechanism as post-processing, which ensures it does not increase $\epsilon_{n,i}^{1:t}$. Recall that the mechanism $\randfunc_i^{1:t}$ combines group models $\model_{m}^{\tau}$ for all $ m\in \calM_i$ and $\tau\in [t]$. To find which groups' models contribute to any of $\model_{m}^{\tau}$, we separate the consideration of inter- and intra-group epochs for each $\tau' \in \left[1, \tau\right]$.

\textbf{Intra-group contributors:} When $(\tau'\mod S)\neq 1$, we encounter intra-group epochs. Thus, every $\model_{m}^{\tau'+1}$ is impacted by only the preceding $\Delta\model_{m}^{\tau'}$ and $\model_{m}^{\tau'}$. 

\textbf{Inter-group contributors:}
When $(\tau'\mod S)= 1$, we encounter an inter-group epoch. Thus, $\model_{m}^{\tau'+1}$ is impacted by $\Delta\model_{m}^{\tau'}$ and all $\model_{m'}^{\tau'}$, where $m'\in \calM$ satisfies $\groupdist_{m,m'}\leq 1$. 

We now combine all contributors to $\randfunc_i^{1:t}$ during both intra- and inter-group epochs. Later in our proofs, we will use this new combined mechanism to place an upper bound on $R_{\alpha}\left( \distr_{\randfunc_i^{1:t}(\datapool)} \| \distr_{\randfunc_i^{1:t}(\datapool')} \right)$. At epoch $\tau'\in [\tau]$, consider a set of group models $ \model_{m'}^{\tau'}$ where $m'\in \calM$ satisfies $\tilde{\groupdist}_{m,m'}\leq \left\lfloor \frac{t-1}{S}\right\rfloor-\left\lceil\frac{\tau'}{S}\right\rceil$ for at least one $m\in\calM_i$. We refer to this set as a layer with index $\tau'$. Combining all mechanisms of the first $\tau$ layers, we define $\tilde{\randfunc}_{i}^{1:\tau,t}$ as
\begin{align}\label{eq:app:thm1:ftilde}
\tilde{\randfunc}_{i}^{1:\tau,t}:= \bigcup_{\tau'\in [\tau]}\bigcup_{m'\in \left\{j :\; j\in \calM,\;\exists m\in \calM_i \text{ st. } \groupdist_{m,j}\leq \left\lfloor \frac{t-1}{S}\right\rfloor-\left\lceil\frac{\tau'}{S}\right\rceil\right\}} \model_{m'}^{\tau'}.
\end{align}
 
Note that $\tilde{\randfunc}_{i}^{1:\tau,t}$ is a sequence of $\tau$ layers. Within each layer, mechanisms operate concurrently, while the layers progress sequentially. For example, layer $t$ encompasses all mechanisms $\model_m^t$, $m\in\calM_i$. In another example, if $t-1>S\left\lfloor \frac{t-1}{S} \right\rfloor$, layer $t-1$ encompasses $ \model_{m}^{t-1}$, $m\in \calM_i$. Otherwise (i.e., if $t-1=S\left\lfloor \frac{t-1}{S} \right\rfloor$), layer $t-1$ encompasses all $ \model_{m'}^{t-1}$,
 where $\groupdist_{m,m'}\leq 1$ for at least one $m\in \calM_i$. Every group model that is included in $\randfunc_i^{1:t}$ also appears in $\tilde{\randfunc}_{i}^{1:t,t}$. Moreover, due to overlapping groups, $\tilde{\randfunc}_{i}^{1:t,t}$ may contain group models that are not part of $\randfunc_i^{1:t}$. Thus, by Lem.~\ref{app:lem:moreinfo} in App.~\ref{app:preliminaries}, we can upper bound the Renyi divergence between the distribution of $\randfunc_i^{1:t}$ over neighboring $\calD \stackrel{n}{\equiv} \calD'$ as
\begin{align*}
R_{\alpha}\left( \distr_{\randfunc_i^{1:t}(\datapool)} \| \distr_{\randfunc_i^{1:t}(\datapool')} \right) \leq R_{\alpha} \left(  \distr_{\tilde{\randfunc}_{i}^{1:t,t}(\datapool)} \| \distr_{\tilde{\randfunc}_{i}^{1:t,t}(\datapool')} \right).
\end{align*}

By applying layered composition lemma (Lem.~\ref{lem:gcomposition} in App.~\ref{app:lem:gcomposition}), we derive the recursive formula:
\begin{align}\label{eq:sec:thm:scenario1:recursive}
&\exp\left((\alpha-1)R_{\alpha}(\distr_{\tilde{\randfunc}_{i}^{1:\tau,t}(\datapool)} \| \distr_{\tilde{\randfunc}_{i}^{1:\tau,t}(\datapool'))}\right)\nonumber \\ &\leq e^{(\alpha-1)R_{\alpha}(\distr_{\tilde{\randfunc}_{i}^{1:\tau-1,t}(\datapool)} \| \distr_{\tilde{\randfunc}_{i}^{1:\tau-1,t}(\datapool'))}}
\times \nonumber \\ & 
\prod_{m'\in \left\{j :\; j\in \calM,\;\exists m\in \calM_i \text{ st. } \groupdist_{m,j}\leq \left\lfloor \frac{t-1}{S}\right\rfloor-\left\lceil\frac{\tau}{S}\right\rceil\right\}} e^{(\alpha-1)\epsilon_{m',n}}.
\end{align}

If worker $n$ is targeted, $\epsilon_{m',n}$ is associated to the group update $\Delta\model_{m'}^{\tau-1}$. If $m'\in \calM_{n}$, by Lem.~\ref{lem:rdpsGM} in App.~\ref{app:preliminaries}, for a sampled Gaussian mechanism under $L_2$-sensitivity ${c}_{m'}$, noise $\calN(0,{c}_{m'}^2\sigma_{m'}^2)$, and sampling probability $\pi_{m'}$, $\epsilon_{m',n}=\epsilon_{m'}=\frac{2\pi_{m'}^2\alpha}{\sigma_{m'}^2}$. If $m'\notin \calM_{n}$, $\epsilon_{m',n}=0$ due to post processing, Lem.~\ref{lem:postproc} in App.~\ref{app:preliminaries}. Recursively applying (\ref{eq:sec:thm:scenario1:recursive}) for $\tau\in [t]$, we obtain
\begin{align*}
&\exp\left((\alpha-1)R_{\alpha}(\distr_{\tilde{\randfunc}_{i}^{1:t,t}(\datapool)} \| \distr_{\tilde{\randfunc}_{i}^{1:t,t}(\datapool'))}\right)\nonumber \\ &\leq  \prod_{\tau \in [t]}\prod_{m'\in \left\{j :\; j\in \calM_n,\;\exists m\in \calM_i \text{ st. } \groupdist_{m,j}\leq \left\lfloor \frac{t-1}{S}\right\rfloor-\left\lceil\frac{\tau}{S}\right\rceil\right\}} e^{(\alpha-1)\epsilon_{m'}}.
\end{align*}

Since $\alpha>1$, the above inequality  results in
\begin{align}\label{eq:sec:thm:scenario1:finalbound}
&R_{\alpha}\left(\distr_{\randfunc_i^{1:t}(\datapool)} \| \distr_{\randfunc_i^{1:t}(\datapool')}\right) \nonumber 
\\ &\leq \sum_{\tau \in [t]}\sum_{m'\in \left\{j :\; j\in \calM_n,\;\exists m\in \calM_i \text{ st. } \groupdist_{m,j}\leq \left\lfloor \frac{t-1}{S}\right\rfloor-\left\lceil\frac{\tau}{S}\right\rceil\right\}} \epsilon_{m'}.
\end{align}

In (\ref{eq:sec:thm:scenario1:finalbound}), if $m'\in \calM_n\backslash\calM_i$ and $\left\lfloor \frac{t-1}{S}\right\rfloor>\tilde{\groupdist}_{m',i}$, $\epsilon_{m'}$ appears during $\left\lfloor \frac{t-1}{S}\right\rfloor-\tilde{\groupdist}_{m',i}$ inter-group epochs. After each inter-group epoch, $\epsilon_{m'}$ accumulates $S-1$ times over $S-1$ intra-group epochs. As a result, in this case, $\epsilon_{m'}$ adds up to $S\left(\left\lfloor \frac{t-1}{S}\right\rfloor-\tilde{\groupdist}_{m',i}\right)$ terms to $\epsilon_{n,i}^{1:t}$. In the case of $m'\in \calM_i$, $\epsilon_{m'}$ accumulates in every single epoch $\tau\in [t-1]$, adding up to $\epsilon_{m'}(t-1)$ terms to $\epsilon_{n,i}^{1:t}$. Hence, $\epsilon_{n,i}^{1:t}$ meets the condition in (\ref{eq:theorem}). The worst-case bound equals $\epsilon_n^{1:t} = \max_{i\in \calN\backslash \{n\}} \epsilon_{n,i}^{1:t}$. $ \blacksquare$

\subsubsection{Proof of Thm.~\ref{thm:scenario1} for DP-OGL+ under the Threat Model~\ref{threat2}}\label{sec:thm:scenario1_2}
The proof of the theorem for this scenario is similar to the previous one, except for the following changes. Since DP-OGL+ addresses only out-of-group HbC workers under Threat Model~\ref{threat2}, we need to track only inter-group contributions introduced in App.~\ref{sec:thm:scenario1_1}. At each inter-group epoch $\tau'\in [\tau]$, we consider a set of group models $\model_{m'}^{\tau'S+1}$ where $m'\in \calM$ satisfies $\tilde{\groupdist}_{m,m'}\leq \left\lfloor \frac{t-1}{S}\right\rfloor-\tau'$ for at least one $m\in\calM_i$. We refer to this set as a layer with index $\tau'$.
Mixing all mechanisms from the first $\tau$ layers in this scenario, the combined mechanism $\tilde{\randfunc}_{i}^{1:\tau,t}$ differs from (\ref{eq:app:thm1:ftilde}). Here, $\tilde{\randfunc}_{i}^{1:\tau,t}$ is
\begin{align}\label{eq:app:thm1:ftilde+}
\tilde{\randfunc}_{i}^{1:\tau,t}:= \bigcup_{\tau'\in [\tau]}\bigcup_{m'\in \left\{j :\; j\in \calM,\;\exists m\in \calM_i \text{ st. } \groupdist_{m,j}\leq \left\lfloor \frac{t-1}{S}\right\rfloor-\tau'\right\}} \model_{m'}^{\tau'S+1}.
\end{align}

Recall that the mechanism $\randfunc_i^{1:t}$ combines group models $\model_{m}^{\tau S+1}$ for all $ m\in \calM_i$ and $\tau\in \left[\lfloor \frac{t-1}{S} \rfloor \right]$. Every group model in $\randfunc_i^{1:t}$ also appears in $\tilde{\randfunc}_{i}^{1:\lfloor \frac{t-1}{S} \rfloor ,t}$. Thus, by Lem.~\ref{app:lem:moreinfo} in App.~\ref{app:preliminaries}
\begin{align*}
R_{\alpha}\left( \distr_{\randfunc_i^{1:t}(\datapool)} \| \distr_{\randfunc_i^{1:t}(\datapool')} \right) \leq R_{\alpha} \left(  \distr_{\tilde{\randfunc}_{i}^{1:t,t}(\datapool)} \| \distr_{\tilde{\randfunc}_{i}^{1:t,t}(\datapool')} \right).
\end{align*}

By applying layered composition lemma (Lem.~\ref{lem:gcomposition} in App.~\ref{app:lem:gcomposition}), we derive the recursive formula:
\begin{align}\label{eq:sec:thm:scenario1:recursive+}
&\exp\left((\alpha-1)R_{\alpha}(\distr_{\tilde{\randfunc}_{i}^{1:\tau,t}(\datapool)} \| \distr_{\tilde{\randfunc}_{i}^{1:\tau,t}(\datapool'))}\right)\nonumber \\ &\leq \exp\left((\alpha-1)R_{\alpha}(\distr_{\tilde{\randfunc}_{i}^{1:\tau-1,t}(\datapool)} \| \distr_{\tilde{\randfunc}_{i}^{1:\tau-1,t}(\datapool'))}\right)\times \nonumber \\ 
& \prod_{m'\in \left\{j :\; j\in \calM,\;\exists m\in \calM_i \text{ st. } \groupdist_{m,j}\leq \left\lfloor \frac{t-1}{S}\right\rfloor-{\tau}\right\}} \exp\left((\alpha-1)\epsilon_{m',n}\right),
\end{align}
where $\epsilon_{m',n}$ is associated to the group model $\model_{m'}^{\tau S+1}$ if worker $n$ is targeted. If $m'\in \calM_{n}$, by Lem.~\ref{lem:rdpsGM} in App.~\ref{app:preliminaries}, for a sampled Gaussian mechanism under $L_2$-sensitivity $\sqrt{S}{c}_{m'}$, noise $\calN(0,S{c}_{m'}^2\sigma_{m'}^2)$, and sampling probability $\pi_{m'}$, $\epsilon_{m',n}=\epsilon_{m'}=\frac{2\pi_{m'}^2\alpha}{\sigma_{m'}^2}$. If $m'\notin \calM_{n}$,  $\epsilon_{m',n}=0$ due to post processing, Lem.~\ref{lem:postproc} in App.~\ref{app:preliminaries}. Recursively applying (\ref{eq:sec:thm:scenario1:recursive+}) for $\tau\in [t]$,
\begin{align}
&
\exp\left((\alpha-1)R_{\alpha}(\distr_{\tilde{\randfunc}_{i}^{1:\lfloor \frac{t-1}{S} \rfloor ,t}(\datapool)} \| \distr_{\tilde{\randfunc}_{i}^{1:\lfloor \frac{t-1}{S} \rfloor ,t}(\datapool'))}\right)
\nonumber \\ &
\leq  \prod_{\tau \in \left[\lfloor \frac{t-1}{S} \rfloor \right]}\prod_{m'\in \left\{j :\; j\in \calM_n,\;\exists m\in \calM_i \text{ st. } \groupdist_{m,j}\leq \left\lfloor \frac{t-1}{S}\right\rfloor-\tau\right\}} e^{(\alpha-1)\epsilon_{m'}}.
\end{align}

Since $\alpha>1$, the above inequality  results in
\begin{align}
&
R_{\alpha}\left(\distr_{\randfunc_i^{1:t}(\datapool)} \| \distr_{\randfunc_i^{1:t}(\datapool')}\right) 
\nonumber 
\\ &
\leq \sum_{\tau \in [t]}\sum_{m'\in \left\{j :\; j\in \calM_n,\;\exists m\in \calM_i \text{ st. } \groupdist_{m,j}\leq \left\lfloor \frac{t-1}{S}\right\rfloor-\left\lceil\frac{\tau}{S}\right\rceil\right\}} \epsilon_{m'}. \label{eq:sec:thm:scenario1:finalbound+}
\end{align}

In (\ref{eq:sec:thm:scenario1:finalbound+}), if $\left\lfloor \frac{t-1}{S}\right\rfloor>\tilde{\groupdist}_{m',i}$, $\epsilon_{m'}$ appears during $\left\lfloor \frac{t-1}{S}\right\rfloor-\tilde{\groupdist}_{m',i}$ epochs. Hence, $\epsilon_{n,i}^{1:t}$ meets the condition in (\ref{eq:theorem+}). The worst-case bound equals $\epsilon_n^{1:t} = \max_{i\in \calN\backslash \hat{\calN}_{n}} \epsilon_{n,i}^{1:t}$. $ \blacksquare$

\subsection{LSI Constant Sequence in DP-OGL and DP-OGL+}
In the following two subsections, we show that the distributions of both $\Delta \model_{m}^{t}$ in DP-OGL (with pseudocode presented in Alg.~\ref{alg:Fplus}) and $\model_{m}^{S\tau+1}-\model_{m}^{S(\tau-1)+1}$ in DP-OGL+ satisfy the log-Sobolev inequality (LSI, as defined in Def.~\ref{app:def:LSI} in App.~\ref{app:preliminaries}).

\subsubsection{LSI Constant Sequence in DP-OGL}\label{app:lem:LSI}

\begin{lem}\label{app:lem:LSIalg}
Let loss functions $\lossfunc_n$ be convex and $\smooth$-smooth ($n\in\calN$). Assume full participation of workers in their groups, i.e., $\pi_m=1$ for all $m\in \calM$. For each $m\in \calM$ and epoch $t$, the distribution of $\Delta \model_{m}^{t}$ in DP-OGL satisfies LSI with constant $\lsicont_m^{t}$ that follows the recursive formulas
\begin{align}\label{app:eq:LSI_constants}
\begin{cases}
b_{m}^{t} = \left(\frac{1}{b_m^{t-1}}+\frac{\pi_m^2|\calN_m|^2}{e_m^{t-1}} \right)^{-1} \\ 
a_{n,m}^t = \left(\sum_{m\in \calM_n} \frac{1}{b_{m}^{t}}\right)^{-1} \;\; \text{if } (t\mod S)\neq 1 \\
a_{n,m}^t = b_{m}^t \;\; \text {if } (t\mod S) =1 \\
h_{n,m}^{t} = \frac{a_{n,m}^t}{\left(1+(1+\lrloc \smooth )^L\right)^2} \\
e_m^{t} = \left({c}_m^2\sigma_m^2 + \sum_{n\in \calN_m} \frac{1}{h_n^t} \right)^{-1}
\end{cases}.
\end{align}
 
\textit{Proof:} We next prove that the empirical loss, as derived from Eq.~(\ref{eq:eloss}), is both $\smooth$-smooth and convex. Using the triangle inequality and considering the $\smooth$-smoothness of $\lossfunc_n$, we obtain:
\begin{align*}
&\left| \nabla \tilde{\lossfunc}_n(z) - \nabla \tilde{\lossfunc}_n(z') \right| \nonumber
\\  &= \left| \frac{1}{|\calB_n|}\sum_{\dsetelement \in \calB_n} \nabla \lossfunc_n(z; \dsetelement) - \nabla \lossfunc_n(z'; \dsetelement) \right| 
\\ &\leq \frac{1}{|\calB_n|}\sum_{\dsetelement \in \calB_n}  \left| \nabla \lossfunc_n(z; \dsetelement) - \nabla \lossfunc_n(z'; \dsetelement) \right| 
\\ &\leq \frac{\smooth}{|\calB_n|}\sum_{\dsetelement \in \calB_n} \left|z-z' \right| = \smooth \left|z-z' \right| .
\end{align*}
This validates the $\smooth$-smoothness of $\tilde{\lossfunc}_n$. Based on the triangle inequality and the convexity of $\lossfunc_n$, for any $\lambda \in [0,1]$ we have
\begin{align*}
& \tilde{\lossfunc}_n(\lambda z + (1-\lambda) z')  \nonumber
\\  &=  \frac{1}{|\calB_n|}\sum_{\dsetelement \in \calB_n} \lossfunc_n\left(\lambda z + (1-\lambda) z'; \dsetelement \right)
\\ & \leq \frac{1}{|\calB_n|}\sum_{\dsetelement \in \calB_n} \left(  \lambda  \lossfunc_n(z; \dsetelement) + (1-\lambda)  \lossfunc_n(z'; \dsetelement) \right)  
\\ & = \lambda \tilde{\lossfunc}_n(z; \dsetelement)  +  (1-\lambda) \tilde{\lossfunc}_n(z'; \dsetelement) .
\end{align*}

This confirms the convexity of $\tilde{\lossfunc}_n$. 
We next prove that the worker model $\mper_{n,m}^{t,l}$, as derived from Eq.~(\ref{eq:workermodel}), is $(1+\lrloc \smooth )$-Lipschitz given the input $\mper_{n,m}^{t,l-1}$. When we consider the input choices $\mper_{n,m}^{t,l-1}\in \{z,z'\}$, the $\smooth$-smoothness of $\tilde{\lossfunc}_n$ and the triangle inequality ensure that $\mper_{n,m}^{t,l}$ satisfies
\begin{align}
& \left|\mper_{n,m}^{t,l}(z)-\mper_{n,m}^{t,l}(z')\right| \nonumber
\\ &= \left| z - \lrloc \nabla \tilde{\lossfunc}(z; \calB_{n,m}^{t,l}) - z' + \lrloc \nabla\tilde{\lossfunc}(z'; \calB_{n,m}^{t,l}) \right|
\\ & \leq \left| z-z' \right| + \lrloc \left| \nabla \tilde{\lossfunc}(z; \calB_{n,m}^{t,l}) - \nabla\tilde{\lossfunc}(z'; \calB_{n,m}^{t,l}) \right| 
\\ & \leq \left| z-z' \right|(1+\lrloc \smooth ). \label{app:lem:LSIalg:lipmper:recursive}
\end{align}
By applying~(\ref{app:lem:LSIalg:lipmper:recursive}) recursively, we prove that $\Delta\mper_{n,m}^{t}$ is Lipschitz with a constant of $1+(1+\lrloc \smooth )^L$, given the input $\mper_{n,m}^{t,0}$. Setting the input choices $\mper_{n,m}^{t,0}\in \{z,z'\}$,
\begin{align}
& \left|\Delta\mper_{n,m}^{t}(z)-\Delta\mper_{n,m}^{t}(z')\right| \nonumber
\\ &= \left|\mper_{n,m}^{t,L}(z)-z - \mper_{n,m}^{t,L}(z')-z' \right|
\\ &\leq \left|z-z' \right| + \left|\mper_{n,m}^{t,L}(z) - \mper_{n,m}^{t,L}(z')\right|
\\ & \leq \left|z-z' \right| + (1+\lrloc\smooth)\left|\mper_{n,m}^{t,L-1}(z) - \mper_{n,m}^{t,L-1}(z')\right|
\\ & \leq \left| z-z' \right|\left(1+(1+\lrloc \smooth )^L\right). \label{app:lem:LSIalg:lipmper}
\end{align}
We note that clipping workers' updates do not increase the Lipschitz constant of $\Delta\mper_{n,m}^{t}$. This is because for any $n\in \calN_m$, 
\begin{align*}
&\left|\text{Clip}(\Delta\mper_{n,m}^{t}(z),c_m)-\text{Clip}(\Delta\mper_{n,m}^{t}(z'),c_m)\right| \nonumber \\ &\leq \left|\Delta\mper_{n,m}^{t}(z)-\Delta\mper_{n,m}^{t}(z')\right|.
\end{align*}

 Let $\Delta\model_m^{t-1}$ and $\model_m^{t-1}$ satisfy LSI with constants $\lsicont_m^{t-1}$ and $b_m^{t-1}$, respectively. By Lem.~\ref{app:lem:Lipschitzmap} (App.~\ref{app:preliminaries}), $\frac{1}{\pi_m|\calN_m|}\Delta\model_m^{t-1}$ satisfies LSI with constant $\frac{\lsicont_m^{t-1}}{\pi_m^2|\calN_m|^2}$. Applying Lem.~\ref{app:lem:convmap} (App.~\ref{app:preliminaries}) to $\frac{1}{\pi_m|\calN_m|}\Delta\model_m^{t-1}$ and $\model_m^{t-1}$ yields a LSI constant for $\model_m^{t}$: 
\begin{align} \label{app:eq:LSI_constants_b}
 b_{m}^{t} := \left(\frac{1}{b_m^{t-1}}+\frac{\pi_m^2|\calN_m|^2}{\lsicont_m^{t-1}} \right)^{-1}.
\end{align} 

To initiate $\mper_{n,m}^{t,0}$, there is two cases. If $(t\mod S)=1$, as shown in (\ref{eq:workerinit}) we linearly combine models $\model_m^{t}$, $m\in \calM_n$ to derive $\mper_{n,m}^{t,0}$. By Lem.~\ref{app:lem:convmap}, $\mper_{n,m}^{t,0}$ maintains LSI with constant 
\begin{align}\label{app:eq:LSI_constants_a}
a_{n,m}^t := \left(\sum_{m\in \calM_n} \frac{1}{b_{m}^{t}}\right)^{-1}.
\end{align}
If $(t\mod S)\neq 1$, $\mper_{n,m}^{t,0} = \model_m^{t}$. Thus, $a_{n,m}^t=b_m^t$.
Furthermore, using LSI under Lipchitz mapping (per Lem.~\ref{app:lem:Lipschitzmap} in App.~\ref{app:preliminaries}), $\Delta\mper_{n,m}^{t}$ satisfies LSI with constant $a_{n,m}^t / \left(1+(1+\lrloc \smooth )^L\right)^2$. Thereby, $\text{Clip}(\Delta\mper_{n,m}^{t},c_m)$ follows LSI with constant 
\begin{align}\label{app:eq:LSI_constants_h}
h_{n,m}^t := \frac{a_{n,m}^t}{\left(1+(1+\lrloc \smooth )^L\right)^2}.
\end{align} 

Using Gaussian convolution and Lem.~\ref{app:lem:gaussmap} (cf. App.~\ref{app:preliminaries}), $\Delta\model_m^{t}$ satisfies LSI with constant 
\begin{align}\label{app:eq:LSI_constants_e}
\lsicont_m^{t} := \left({c}_m^2\sigma_m^2 + \sum_{n\in \calW_m^t} \frac{1}{h_{n,m}^t} \right)^{-1},
\end{align}
where the subset $\calW_m^t$ includes all workers in $\calN_m$ owing to the assumption of $\pi_m=1$. $\blacksquare$

\end{lem}

\subsubsection{LSI Constant Sequence in DP-OGL+}\label{app:lem:LSI+}
\begin{lem}\label{app:lem:LSIalg+}
Let the loss functions $\lossfunc_n$ be convex and $\smooth$-smooth for all $n\in\calN$. Let $\pi_m=1$ for all $m\in \calM$. For any $m\in \calM$ and epoch $t$, the distribution of $(\model_{m}^{S\tau+1}-\model_{m}^{S(\tau-1)+1})\pi_m|\calN_m|$ in DP-OGL+ satisfies LSI with constant $\lsicont_m^{\tau}$ that follows the recursive formulas
\begin{align}\label{app:eq:LSI_constants+}
\begin{cases}
 b_{m}^{S\tau+1} = \left(\frac{1}{b_m^{S(\tau-1)+1}}+\frac{\pi_m^2|\calN_m|^2}{\lsicont_m^{\tau-1}} \right)^{-1} \\
b_{m}^{t} =  \left(\frac{1}{b_{m}^{t-1}} + \sum_{n\in \calW_m^{t-1}} \frac{\pi_m^2|\calN_m|^2}{h_{n,m}^{t-1}} \right)^{-1}  \;\; \text{if } (t\mod S)\neq 1 \\ 
a_{n,m}^t = \left(\sum_{m\in \calM_n} \frac{1}{b_{m}^{t}}\right)^{-1} \;\; \text{if } (t\mod S)= 1 \\
a_{n,m}^t = b_{m}^t \;\; \text {if } (t\mod S) \neq 1 \\
h_{n,m}^{t} = \frac{a_{n,m}^t}{\left(1+(1+\lrloc \smooth )^L\right)^2} \\
\lsicont_m^{\tau} = \left(S{c}_m^2\sigma_m^2 +\sum_{t=S\tau+1}^{S(\tau+1)}\sum_{n\in \calW_m^{t}} \frac{1}{h_{n,m}^t}\right)^{-1}
\end{cases}.
\end{align}
 
\textit{Proof:} As is shown in Lem.~\ref{app:lem:LSIalg}'s proof, $\tilde{\lossfunc}_n$ is convex, and $\Delta\mper_{n,m}^{t}$ is Lipschitz with  constant $1+(1+\lrloc \smooth )^L$, given the initial model $\mper_{n,m}^{t,0}$. Let $(\model_{m}^{S\tau+1}-\model_{m}^{S(\tau-1)+1})\pi_m|\calN_m|$ and $\model_{m}^{t}$ satisfy LSI with constants $\lsicont_m^{\tau-1}$ and $b_m^{t}$, respectively. 
Applying Lem.~\ref{app:lem:convmap} (App.~\ref{app:preliminaries}) to $\model_{m}^{S\tau+1}=\frac{(\model_{m}^{S\tau+1}-\model_{m}^{S(\tau-1)+1})\pi_m|\calN_m|}{\pi_m|\calN_m|}+\model_{m}^{S(\tau-1)+1}$ yields a LSI constant for $\model_{m}^{S\tau+1}$: 
\begin{align} \label{app:eq:LSI_constants_b+}
 b_{m}^{S\tau+1} = \left(\frac{1}{b_m^{S(\tau-1)+1}}+\frac{\pi_m^2|\calN_m|^2}{\lsicont_m^{\tau-1}} \right)^{-1}.
\end{align} 

When $(t\mod S)\neq 1$ in DP-OGL+, the group model $\model_{m}^{t}$ is obtained as (\ref{eq:masterupdate+}). Assume that the aggregated worker updates $\sum_{n\in \calW_m^{t-1}}(\mper_{n,m}^{t-1,L}-\mper_{n,m}^{t-1,0})$ satisfies LSI with constant $\tilde{\lsicont}_{m}^{t-1}$. Therefore, when $(t\mod S)\neq 1$ and by reapplying Lem.~\ref{app:lem:convmap}, $\model_{m}^{t}$ satisfies LSI with constant
\begin{align}\label{app:eq:LSI_constants_b+2}
b_{m}^{t} =  \left(\frac{1}{b_{m}^{t-1}} + \frac{\pi_m^2|\calN_m|^2}{\tilde{\lsicont}_m^{t-1}} \right)^{-1}.
\end{align}

Using Lem.~\ref{app:lem:gaussmap} (cf. App.~\ref{app:preliminaries}), $(\model_m^{S(\tau+1)+1}-\model_m^{S\tau+1})\pi_m|\calN_m|$ satisfies LSI with constant 
\begin{align}\label{app:eq:LSI_constants_e+}
\lsicont_m^{\tau} = \left(S{c}_m^2\sigma_m^2 +\sum_{t=S\tau+1}^{S(\tau+1)} \frac{1}{\tilde{\lsicont}_m^{t}} \right)^{-1}.
\end{align}

Let's assume that each worker update $\mper_{n,m}^{t,L}-\mper_{n,m}^{t,0}$ satisfies LSI with constant $h_{n,m}^{t}$. Therefore, applying Lem.~\ref{app:lem:convmap}, the LSI constant of 
$\sum_{n\in \calW_m^{t}}(\mper_{n,m}^{t,L}-\mper_{n,m}^{t,0})$ satisfies
\begin{align}\label{app:eq:LSI_constants_e+2}
\tilde{\lsicont}_{m}^t = \left(\sum_{n\in \calW_m^{t}} \frac{1}{h_{n,m}^t}\right)^{-1}
\end{align}

As is shown in (\ref{eq:workerinit}), we initiate $\mper_{n,m}^{S\tau+1,0}$ by linearly combining models $\model_{m'}^{S\tau+1}$, for all $m'\in \calM_n$. By Lem.~\ref{app:lem:convmap}, $\mper_{n,m}^{S\tau+1,0}$ maintains LSI with constant 
\begin{align}\label{app:eq:LSI_constants_a+}
a_{n,m}^{S\tau+1} := \left(\sum_{m'\in \calM_n} \frac{1}{b_{m'}^{S\tau+1}}\right)^{-1}.
\end{align}
When $(t\mod S)\neq 1$, $\mper_{n,m}^{t,0}$ is initialized by $\model_{m}^t$, and thus satisfies LSI with constant $a_{n,m}^{t}=b_{m}^t$. As shown in the proofs of Lem.~\ref{app:lem:LSIalg+}, using LSI under Lipchitz mapping (per Lem.~\ref{app:lem:Lipschitzmap} in App.~\ref{app:preliminaries}), $\Delta \mper_{n,m}^{t}$ satisfies LSI with constant
\begin{align}\label{app:eq:LSI_constants_h+}
h_{n,m}^{t} := \frac{a_{n,m}^t}{\left(1+(1+\lrloc \smooth )^L\right)^2}.\; \blacksquare
\end{align} 
\end{lem}

\subsection{Proof of Lem.~\ref{lem:scenario2}}\label{app:corollaries}

\subsubsection{Proof of Lem.~\ref{lem:scenario2} for DP-OGL under the Threat Model~\ref{threat1}}\label{app:corollaries_dpogl}
Here, we apply Lemmas~3.1. and 3.2. from~\cite{ye2022differentially}. These lemmas prove how Rényi privacy loss decays after post-processing with additive Gaussian noise. We make notational adjustments to adapt Lemmas' notation to fit our narrative and use them to prove Lem.~\ref{lem:scenario2}. In Lem.~\ref{lem:scenario2}, we assume $\pi_m=1$ for every $m\in \calM$. We also consider $\calD \stackrel{n}{\equiv} \calD'$.
We represent the model updates for worker $i\in \calW_m^{\tau}$ in group $m\in \calM_i$ as $\Delta\mper_{i,m}^{\tau}(\calD)$ and ${\Delta\mper_{i,m}^{\tau}}(\calD')$ for datasets $\calD$ and $\calD'$, respectively. Similarly, $\model_m^{\tau}(\calD)$ and $\model_{m}^{\tau}(\calD')$ represent the group $m$'s models using $\calD$ and $\calD'$. 
For any group $m\in \calM$, we consider two cases.

\textbf{Case 1 ($n\notin\calN_m$):} Consider group $m\in \calM$ where the targeted worker $n\in \calN$ is not a member. We first define 
\begin{align}\label{app:lemm1:umD}
u_m^{\tau}(s,\calD) &:= \model_m^{\tau}(\calD)+\frac{\sum_{j\in \calN_m}\text{Clip}\left(\Delta\mper_{j,m}^{\tau}(\calD),c_m\right)}{|\calN_m|} 
\nonumber \\ &
+ \calN(0,2s\sigma_m^2\calI_v)
\\ u_m^{\tau}(s,\calD') &:=   \model_m^{\tau}(\calD')+\frac{\sum_{j\in \calN_m}\text{Clip}\left({\Delta\mper_{j,m}^{\tau}}(\calD'),c_m\right)}{|\calN_m|}
\nonumber \\ & 
+\calN(0,2s\sigma_m^2\calI_v).\label{app:lemm1:umD'}
\end{align}

When $s=\frac{{c}_m^2}{2}$, $u^{\tau}(s,\calD)=\model_m^{\tau+1}(\calD)$ and $u^{\tau}(s,\calD')=\model_m^{\tau+1}(\calD')$. By Lem.~\ref{app:lem:convmap} in App.~\ref{app:preliminaries} and Lem.~\ref{app:lem:LSIalg} in App.~\ref{app:lem:LSI}, $\model_m^{\tau}(\calD)+\frac{\sum_{j\in \calN_m}\text{Clip}\left(\Delta\mper_{j,m}^{\tau}(\calD),c_m\right)}{|\calN_m|}$ and $\model_m^{\tau}(\calD')+\frac{\sum_{j\in \calN_m}\text{Clip}\left({\Delta\mper_{j,m}^{\tau}}(\calD'),c_m\right)}{|\calN_m|}$ satisfy LSI with constant 
\begin{align}\label{app:eq:lsi:sumClips}
\bar{h}_m^{\tau}=\left(\frac{1}{b_m^{\tau}}+\sum_{j\in\calW_m^{\tau}}\frac{|\calN_m|^2}{h_{j,m}^{\tau}}\right)^{-1}.
\end{align}

By applying Lem.~3.1. from~\cite{ye2022differentially}, we derive (\ref{app:cor1:eq}). 
This derivation is built on our assumption that the loss functions $\lossfunc_n$, $n\in \calN$, are convex and $\smooth$-smooth. Given any $\alpha>1$, 
\begin{align}\label{app:cor1:eq}
&\frac{\partial}{\partial s}R_{\alpha}(\distr_{u_m^{\tau}(s,\calD)}\|\distr_{u_m^{\tau}(s,\calD')})\leq -\frac{2\sigma_m^2(\alpha-1)}{\frac{1}{\bar{h}_m^{\tau}}+2s\sigma_m^2}\times \nonumber \\ &\left(\frac{R_{\alpha}(\distr_{u_m^{\tau}(s,\calD)}\|\distr_{u_m^{\tau}(s,\calD')})}{\alpha(\alpha-1)}+\frac{\partial R_{\alpha}\left(\distr_{u_m^{\tau}(s,\calD)}\| \distr_{u_m^{\tau}(s,\calD')}\right)}{\partial \alpha}\right).
\end{align}

Let's denote the Renyi divergence at $s = 0$ (i.e. before the model of group $m$ is perturbed by Gaussian noise) as $\check{\epsilon}_{n,m,{\tau}}:= R_{\alpha}(\distr_{u_m^{\tau}(0,\calD)}\|\distr_{u_m^{\tau}(0,\calD')})$. 
Since $n\notin \calN_m$, we compute $\check{\epsilon}_{n,m,{\tau}}$, considering two cases: $(\round\mod S)\neq 0$ and $(\round\mod S)= 0$. When $(\tau\mod S)\neq 0$, $\model_{m}^{\tau+1}$ receives information about prior epochs through $\model_{m}^{\tau}$. Thus,
\begin{align}
\check{\epsilon}_{n,m,{\tau}}=R_{\alpha}\left(\distr_{\model_{m}^{\tau}(\calD)} \| \distr_{\model_{m}^{\tau}(\calD')}\right).
\end{align}
When $(\tau\mod S)= 0$,  $\model_{m}^{\tau+1}$ receives previous epochs' information through $\model_{m'}^{\tau}$ for all $m'\in \calM$ st. $\groupdist_{m,m'}\leq 1$. Thus,
\begin{align*}
&\check{\epsilon}_{n,m,{\tau}}
\nonumber \\
&=R_{\alpha}\left(\distr_{\bigcup_{m'\in \calM, \groupdist_{m,m'}\leq 1}\model_{m'}^{\tau}(\calD)} \| \distr_{\bigcup_{m'\in \calM, \groupdist_{m,m'}\leq 1}\model_{m'}^{\tau}(\calD')}\right).
\end{align*}

Applying Lem.~3.2 from~\cite{ye2022differentially} and for any order $\alpha > 1$, we attain multiplicative recursion as
\begin{align*}
\frac{R_{\alpha}\left(\distr_{\model_m^{\tau+1}(\calD)} \|\distr_{ \model_m^{\tau+1}(\calD')}\right)}{\alpha}\leq \frac{\check{\epsilon}_{n,m,{\tau}}}{\alpha+\bar{h}_m^{\tau}{c}_m^2\sigma_m^2}.
\end{align*}

\textbf{Case 2 ($n\in \calN_m$):} Given any $\alpha > 1$, we reuse Lem.~3.2 from~\cite{ye2022differentially} to attain additive recursion as
\begin{align*}
R_{\alpha}\left(\distr_{\model_m^{\tau+1}(\calD)} \|\distr_{ \model_m^{\tau+1}(\calD')} \right) \leq \check{\epsilon}_{n,m,{\tau}} + \frac{\alpha}{2\sigma_m^2}. \; \blacksquare
\end{align*}

\subsubsection{Proof of Lem.~\ref{lem:scenario2} for DP-OGL+ under the Threat Model~\ref{threat2}}\label{app:corollaries_dpogl+}
The proof of the lemma for this scenario is similar to the previous one, except for the following changes. In Case 1, where $n\notin \calN_m$, instead of (\ref{app:lemm1:umD}) and (\ref{app:lemm1:umD'}), we use the following definitions for $u_m^{\tau}(s,\calD)$ and $u_m^{\tau}(s,\calD')$.
\begin{align}\label{app:lemm1:umD+}
&u_m^{\tau}(s,\calD) := \model_m^{S\tau +1}(\calD)+\calN(0,2s\sigma_m^2\calI_v) + 
\nonumber \\ &
\frac{\sum_{j\in \calN_m}\text{Clip}\left(\sum_{t=S\tau+1}^{S(\tau+1)}(\mper_{j,m}^{t,L}-\mper_{j,m}^{t,0})(\calD),\sqrt{S}c_m\right)}{|\calN_m|} 
\\ & u_m^{\tau}(s,\calD') :=   \model_m^{S\tau +1}(\calD')+\calN(0,2s\sigma_m^2\calI_v) + 
\nonumber \\ &
 \frac{\sum_{j\in \calN_m}\text{Clip}\left({\sum_{t=S\tau+1}^{S(\tau+1)}(\mper_{j,m}^{t,L}-\mper_{j,m}^{t,0})}(\calD'),\sqrt{S}c_m\right)}{|\calN_m|}.
\label{app:lemm1:umD'+}
\end{align}

When $s=\frac{S{c}_m^2}{2}$, $u^{\tau}(s,\calD)=\model_m^{S(\tau+1)+1}(\calD)$ and $u^{\tau}(s,\calD')=\model_m^{S(\tau+1)+1}(\calD')$. By Lem.~\ref{app:lem:convmap} in App.~\ref{app:preliminaries} and Lem.~\ref{app:lem:LSIalg} in App.~\ref{app:lem:LSI}, $\model_m^{S\tau +1}(\calD)+\frac{\sum_{j\in \calN_m}\text{Clip}\left(\sum_{t=S\tau+1}^{S(\tau+1)}(\mper_{j,m}^{t,L}(\calD)-\mper_{j,m}^{t,0}(\calD)),\sqrt{S}c_m\right)}{|\calN_m|}$ and $\model_m^{S\tau +1}(\calD')+\frac{\sum_{j\in \calN_m}\text{Clip}\left({\sum_{t=S\tau+1}^{S(\tau+1)}(\mper_{j,m}^{t,L}-\mper_{j,m}^{t,0})}(\calD'),\sqrt{S}c_m\right)}{|\calN_m|}$ satisfy LSI with constant 
\begin{align}\label{app:eq:lsi:sumClips+}
\bar{h}_m^{\tau}=\left(\frac{1}{b_m^{S\tau+1}}+\sum_{t=S\tau+1}^{S(\tau+1)}\sum_{j\in\calW_m^{t}}\frac{|\calN_m|^2}{h_{j,m}^{t}}\right)^{-1}.
\end{align}

Therefore, by applying Lem.~3.1. from~\cite{ye2022differentially}, we derive (\ref{app:cor1:eq}). By Lem.~3.2 from~\cite{ye2022differentially}, we attain multiplicative recursion as
\begin{align*}
\frac{R_{\alpha}\left(\distr_{\model_m^{S\tau+1}(\calD)} \|\distr_{ \model_m^{S\tau+1}(\calD')}\right)}{\alpha}\leq \frac{\check{\epsilon}_{n,m,{S(\tau-1)+1}}}{\alpha+S\bar{h}_m^{\tau}{c}_m^2\sigma_m^2}.
\end{align*}

We reuse Lem.~3.2 from~\cite{ye2022differentially} to attain additive recursion:
\begin{align*}
R_{\alpha}\left(\distr_{\model_m^{S\tau+1}(\calD)} \|\distr_{ \model_m^{S\tau+1}(\calD')} \right) \leq \check{\epsilon}_{n,m,{S(\tau-1)+1}} + \frac{\alpha}{2\sigma_m^2}. \; \blacksquare
\end{align*}

\subsection{Proof of Thm.~\ref{thm:scenario2}}\label{app:thm:scenario2}

\subsubsection{Proof of Thm.~\ref{thm:scenario2} for DP-OGL under Threat Model~\ref{threat1}}\label{app:thm:scenario2_dpogl}
Without loss of generality, we assume that any group index of worker $n$ is equal to or smaller than any group index of worker $i$. In other words, whether worker $i$ participating in a single group $m$ or two groups $m$ and $m+1$ and whether worker $n$ participating in a single group $m'$ or two groups $m'$ and $m'+1$, we assume $m'\leq m$. On the opposite hand if $m'\geq m$, we can reverse the indexing of the groups in the string structure and reuse the following proof. 

 For different groups $m,m'\in \calM$ and epoch $\tau$, we define $T_{m,m'}^{\tau}:=\left\lfloor \frac{\tau}{S}\right\rfloor-\groupdist_{m',m}$. If $T_{m,m'}^{\tau}>0$, this variable represents the latest inter-group epoch when the information about $\model_{m'}^{ST_{m,m'}^{\tau}}$ propagates to group $m$ in epoch $\tau$. If $T_{m,m'}^{\tau}\leq 0$, this indicates that group $m$ in epoch $\tau$ does not receive information about any prior group model $\model_{m'}^{\tau'}$, where $\tau' \leq \tau$. To solve Thm.~\ref{thm:scenario2}, we consider three cases for the assumed string grouping structure. In the first case, the targeted worker $n$ and the HbC worker $i$ have no groups in common. In the second case, workers $n$ and $i$ share two groups. In the third case, workers $n$ and $i$ have only a single group in common.

\textbf{Case 1:} We first consider the group $m\in \calM_i$ that is the closest to every group $m'\in\calM_n$ amongst all groups in $\calM_i$. I.e., $m=\text{arg min}_{j\in \calM_i}\min_{m'\in\calM_n}\groupdist_{j,m'}$. We then consider the sequences of distinct group models, that start with $\model_{m'}^{\tau'}$ and end with $\model_{m}^{\tau+1}$. In this sequence, group $m'$ is the furthest from group $m$ among the groups in $\calM_n$ that propagates information to group $m$ in epoch $\tau$. Assume this group produces $\model_{m'}^{\tau'}$ in epoch $\tau'$ that influences $\model_{m}^{\tau+1}$. 
The starting epoch is $\tau'=ST_{m,m'}^{\tau}-S+1$. In each of such sequences, consecutive pairs of group models $\model_{m_1}^{\tau_1}$ and $\model_{m_2}^{\tau_2}$ follow the pattern: (1) $\tau_2=\tau_1+1$ and (2) $\groupdist_{m_1,m_2}=1$ if $(\tau_2\mod S)=1$, and $\groupdist_{m_1,m_2}=0$ otherwise. Under the assumption of string structure, for each epoch $\tau$ where $\tilde{\groupdist}_{m,n}\leq \left\lfloor \frac{\tau}{S}\right\rfloor$, there exists only one such sequence with the mentioned pattern. In this case, we combine all the group models in the sequence and denote that as $\hat{\randfunc}_{i}^{\tau}$. Note that if $\tilde{\groupdist}_{m,n}> \left\lfloor \frac{\tau}{S}\right\rfloor$, no such sequence exists.

Assume $\tilde{\groupdist}_{m,n}\leq \left\lfloor \frac{\tau}{S}\right\rfloor$ and that the combined mechanism $\randfunc_i^{1:\tau}$ generates outputs $z_{i}^{1:\tau}\sim \distr_{\randfunc_i^{1:\tau}(\calD')}$. Given $\randfunc_i^{1:\tau}(\calD)=z_{i}^{1:\tau}$ and $\randfunc_i^{1:\tau}(\calD')=z_{i}^{1:\tau}$, and using Lem.~\ref{lem:jointconvexity}, we have
\begin{align}\label{app:thm2:jconv1}
&e^{(\alpha-1)R_{\alpha}\left( \distr_{\model_m^{\tau+1}(\calD)\mid \randfunc_i^{1:\tau}(\calD)= z_{i}^{1:\tau}}  \| \distr_{\model_m^{\tau+1}(\calD')\mid \randfunc_i^{1:\tau}(\calD')= z_{i}^{1:\tau}} \right)} \nonumber \\
&\leq \int_{\hat{\tilde{z}}_i^{1:\tau,t}}\distr_{\hat{\tilde{\randfunc}}_i^{1:\tau,t}(\calD')\mid \text{Condition}_1}(\hat{\tilde{z}}_i^{1:\tau,t})
\times \nonumber \\ & \times 
e^{(\alpha-1)R_{\alpha}\left( \distr_{\model_m^{\tau+1}(\calD)\mid \text{Condition}_2}  \| \distr_{\model_m^{\tau+1}(\calD')\mid \text{Condition}_3} \right)},
\end{align}
where, given $\tilde{\randfunc}_i^{1:\tau,t}$ defined as (\ref{eq:app:thm1:ftilde}), 
\begin{align}\label{app:thm2:con1}
\hat{\tilde{\randfunc}}_i^{1:\tau,t}&:=\tilde{\randfunc}_i^{1:\tau,t}\backslash\hat{\randfunc}_i^{\tau},
\\ \text{Condition}_1 &\myeq \begin{cases} \randfunc_i^{1:\tau}(\calD')= z_{i}^{1:\tau}, & 
\\ \randfunc_i^{1:\tau}(\calD)= z_{i}^{1:\tau}, & \\ \hat{\tilde{\randfunc}}_i^{1:\tau,t}(\calD)=\hat{\tilde{\randfunc}}_i^{1:\tau,t}(\calD'),\end{cases} \label{app:thm2:con2} 
\\ \text{Condition}_2 &\myeq \begin{cases}
\randfunc_i^{1:\tau}(\calD)= z_{i}^{1:\tau}, &  
\\ \hat{\tilde{\randfunc}}_i^{1:\tau,t}(\calD) = \hat{\tilde{z}}_{i}^{1:\tau,t},\end{cases} \label{app:thm2:con3} 
\\ \text{Condition}_3 &\myeq \begin{cases}
 \randfunc_i^{1:\tau}(\calD')= z_{i}^{1:\tau}, & \\ \hat{\tilde{\randfunc}}_i^{1:\tau,t}(\calD') = \hat{\tilde{z}}_{i}^{1:\tau,t}.
\end{cases} \label{app:thm2:con4}
\end{align}

Since $\model_m^{\tau+1}$ belongs to the end of the sequence with combined mechanism $\hat{\randfunc}_i^{\tau}$, using Lem.~\ref{app:lem:moreinfo}, 
\begin{align}\label{app:thm2:lem:moreinfo}
&R_{\alpha}\left( \distr_{\model_m^{\tau+1}(\calD)\mid \text{Condition}_2}  \| \distr_{\model_m^{\tau+1}(\calD')\mid \text{Condition}_3} \right)\leq \nonumber \\ 
&R_{\alpha}\left( \distr_{\hat{\randfunc}_i^{\tau}(\calD)\mid \text{Condition}_2}  \| \distr_{\hat{\randfunc}_i^{\tau}(\calD')\mid \text{Condition}_3} \right).
\end{align}

Recursively using Lem.~\ref{lem:scenario2} on group models in $\hat{\randfunc}_i^{\tau}$, we obtain 
\begin{align}\label{app:thm2:recursive0}
&R_{\alpha}\left( \distr_{\hat{\randfunc}_i^{\tau}(\calD)\mid \text{Condition}_2}  \| \distr_{\hat{\randfunc}_i^{\tau}(\calD')\mid \text{Condition}_3} \right)\leq \nonumber  
\\ & \prod_{\tau'' \in (S(\left\lfloor \frac{\tau}{S}\right\rfloor - \tilde{\groupdist}_{m,n}) , S\left\lfloor \frac{\tau}{S}\right\rfloor ]}\mu_{m-S\left(\left\lfloor\frac{\tau}{S}\right\rfloor -\left\lfloor\frac{\tau''}{S}\right\rfloor\right)}^{\tau''} \times \nonumber
\\ & \sum_{m'\in \calM_n}\left(\sum_{\tau'\in \left[\max(1,S(T_{m,m'}^{\tau}-1)), \max(0,ST_{m,m'}^{\tau})\right]}\epsilon_{m'}\right),
\end{align}
where  $\mu_{m''}^{\tau''}=\frac{\alpha}{\alpha+\bar{h}_{m''}^{\tau''}{c}_{m''}^2\sigma_{m''}^2}$ and $\epsilon_{m'}=\frac{\alpha}{2\sigma_{m'}^2}$. Since $\alpha>1$,
\begin{align}\label{app:thm2:recursive}
&e^{(\alpha-1)R_{\alpha}\left( \distr_{\hat{\randfunc}_i^{\tau}(\calD)\mid \text{Condition}_2}  \| \distr_{\hat{\randfunc}_i^{\tau}(\calD')\mid \text{Condition}_3} \right)}  \nonumber \\ 
& \leq \exp \left((\alpha-1)\prod_{\tau'' \in (S(\left\lfloor \frac{\tau}{S}\right\rfloor - \tilde{\groupdist}_{m,n}) , S\left\lfloor \frac{\tau}{S}\right\rfloor ]}\mu_{m-S\left(\left\lfloor\frac{\tau}{S}\right\rfloor -\left\lfloor\frac{\tau''}{S}\right\rfloor\right)}^{\tau''} \right. \nonumber 
\\ &\times \left.  \sum_{m'\in \calM_n}\left(\sum_{\tau'\in \left[\max(1,S(T_{m,m'}^{\tau}-1)), \max(0,ST_{m,m'}^{\tau})\right]}\epsilon_{m'}\right)\right).
\end{align}

Combining (\ref{app:thm2:jconv1}) and (\ref{app:thm2:lem:moreinfo})-(\ref{app:thm2:recursive}), we obtain
\begin{align}\label{app:thm2:1}
&e^{(\alpha-1)R_{\alpha}\left( \distr_{\model_m^{\tau+1}(\calD)\mid \randfunc_i^{1:\tau}(\calD)= z_{i}^{1:\tau}}  \| \distr_{\model_m^{\tau+1}(\calD')\mid \randfunc_n^{1:\tau}(\calD')= z_{i}^{1:\tau}} \right)} \nonumber \\
&\leq \exp \left((\alpha-1)\prod_{\tau'' \in (S(\left\lfloor \frac{\tau}{S}\right\rfloor - \tilde{\groupdist}_{m,n}) , S\left\lfloor \frac{\tau}{S}\right\rfloor ]}\mu_{m-S\left(\left\lfloor\frac{\tau}{S}\right\rfloor -\left\lfloor\frac{\tau''}{S}\right\rfloor\right)}^{\tau''} \right. \nonumber 
\\ &\times \left.  \sum_{m'\in \calM_n}\left(\sum_{\tau'\in \left[\max(1,S(T_{m,m'}^{\tau}-1)), \max(0,ST_{m,m'}^{\tau})\right]}\epsilon_{m'}\right)\right).
\end{align}

Since any other group $j\in \calM_i$ that $j\neq m$ and $j\notin \calM_n$ when conditioned on prior $\randfunc_i^{1:\tau-1}$ does not reveal any information about groups models of worker $n$, we have
\begin{align}\label{app:thm2:2}
e^{(\alpha-1)R_{\alpha}\left( \distr_{\model_j^{\tau+1}(\calD)\mid \randfunc_i^{1:\tau}(\calD)= z_{i}^{1:\tau}}  \| \distr_{\model_j^{\tau+1}(\calD')\mid \randfunc_n^{1:\tau}(\calD')= z_{i}^{1:\tau}} \right)} =1.
\end{align}

By applying layered composition lemma from App.~\ref{app:lem:gcomposition} into (\ref{app:thm2:1}) and (\ref{app:thm2:2}), we derive the recursive formula:
\begin{align}\label{app:thm2:layered}
&\exp\left((\alpha-1)R_{\alpha}(\distr_{{\randfunc}_{i}^{1:\tau+1}(\datapool)} \| \distr_{{\randfunc}_{i}^{1:\tau+1}(\datapool'))}\right)\nonumber \\ &\leq \exp\left((\alpha)R_{\alpha}(\distr_{{\randfunc}_{i}^{1:\tau}(\datapool)} \| \distr_{{\randfunc}_{i}^{1:\tau-1}(\datapool'))}\right)\nonumber \\ 
&\times \exp \left((\alpha-1)\prod_{\tau'' \in (S(\left\lfloor \frac{\tau}{S}\right\rfloor - \tilde{\groupdist}_{m,n}) , S\left\lfloor \frac{\tau}{S}\right\rfloor ]}\mu_{m-S\left(\left\lfloor\frac{\tau}{S}\right\rfloor -\left\lfloor\frac{\tau''}{S}\right\rfloor\right)}^{\tau''} \right. \nonumber 
\\ &\times \left.  \sum_{m'\in \calM_n}\left(\sum_{\tau'\in \left[\max(1,S(T_{m,m'}^{\tau}-1)), \max(0,ST_{m,m'}^{\tau})\right]}\epsilon_{m'}\right)\right).
\end{align}

Recursively using (\ref{app:thm2:layered}) over $t$ epochs, that includes $\left\lfloor \frac{t-1}{S}\right\rfloor$ inter-group epochs, we obtain
\begin{align}\label{app:thm2:final}
&\exp\left((\alpha-1)R_{\alpha}(\distr_{{\randfunc}_{i}^{1:t}(\datapool)} \| \distr_{{\randfunc}_{i}^{1:t}(\datapool'))}\right)\nonumber \\ 
&\leq  \prod_{\tau \in \left[\left\lfloor \frac{t-1}{S}\right\rfloor\right]}\exp \left((\alpha-1)\prod_{\tau'' \in (S(\tau - \tilde{\groupdist}_{m,n}) , S\tau ]}\mu_{m-S\left(\tau -\left\lfloor\frac{\tau''}{S}\right\rfloor\right)}^{\tau''} \right. \nonumber 
\\ &\times \left.  \sum_{m'\in \calM_n}\left(\sum_{\tau'\in \left[\max(1,S(T_{m,m'}^{S\tau}-1)), \max(0,ST_{m,m'}^{S\tau})\right]}\epsilon_{m'}\right)\right).
\end{align}

\textbf{Case 2:} The proof closely resembles the one in Case 1 but with minor adjustments. In (\ref{app:thm2:jconv1}), we consider the group $m$ to be the closest group within $\calM_i$ to every group $m'\in \calM_n$. In Case 2, we reuse (\ref{app:thm2:jconv1}) for each group $m$ in $\calM_i$, which includes worker $n$, as well. In this case, there is no single group within $\calM_i$ that serves as the closest to all groups $m'$ in $\calM_n$ because $\calM_i=\calM_n$. Subsequently, (\ref{app:thm2:recursive}) changes into
\begin{align}\label{app:thm2:recursive2}
e^{(\alpha-1)R_{\alpha}\left( \distr_{\hat{\randfunc}_i^{\tau}(\calD)\mid \text{Condition}_2}  \| \distr_{\hat{\randfunc}_i^{\tau}(\calD')\mid \text{Condition}_3} \right)} \leq e^{(\alpha-1)\epsilon_m}.
\end{align}

Applying this changes into (\ref{app:thm2:final}), we obtain
\begin{align}\label{app:thm2:final2}
&\exp\left((\alpha-1)R_{\alpha}(\distr_{{\randfunc}_{i}^{1:t}(\datapool)} \| \distr_{{\randfunc}_{i}^{1:t}(\datapool'))}\right)\nonumber \\ 
&\leq  \prod_{\tau \in [t]}\prod_{m' \in \calM_i}e^{(\alpha-1)\epsilon_{m'}} \\
&= e^{(\alpha-1)\sum_{m'\in \calM_i}t\epsilon_{m'}}.
\end{align}

\textbf{Case 3:} The proof closely resembles those of Cases 1 and 2, with minor adjustments. Similar to Case 1, there is a single group $m\in \calM_i$ that serves as the closest to all groups $m'\in \calM_n$. However. in contrast to Case 1, in Case 3, group $m$ also includes worker $n$. Similar to Case 2, there is no information degradation because groups of worker $i$ receive information from worker $n$ either through the shared group $m$ or through a group adjacent to group $m$. Therefore, (\ref{app:thm2:recursive}) changes into
\begin{align}\label{app:thm2:recursive3}
&e^{(\alpha-1)R_{\alpha}\left( \distr_{\hat{\randfunc}_i^{\tau}(\calD)\mid \text{Condition}_2}  \| \distr_{\hat{\randfunc}_i^{\tau}(\calD')\mid \text{Condition}_3} \right)}\nonumber \\ &\leq \begin{cases}e^{(\alpha-1)\epsilon_{m'}}(S+1) & \text{ if } (\tau \mod S)=0 \text{ and } \tau>0 \\ e^{(\alpha-1)\epsilon_{m'}} & \text{ ow }\end{cases}.
\end{align}
Applying this changes into (\ref{app:thm2:final}), we obtain 
\begin{align}\label{app:thm2:final3}
&\exp\left((\alpha-1)R_{\alpha}(\distr_{{\randfunc}_{i}^{1:t}(\datapool)} \| \distr_{{\randfunc}_{i}^{1:t}(\datapool'))}\right)
\leq  \prod_{\tau \in \left[\left\lfloor \frac{t-1}{S}\right\rfloor\right]}\exp \left((\alpha-1) 
\right. \nonumber \\ 
&\left.  \sum_{m'\in \calM_n}\left(\sum_{\tau'\in \left[\max(1,S(T_{m,m'}^{S\tau}-1)), \max(0,ST_{m,m'}^{S\tau})\right]}\epsilon_{m'}\right)\right).\; \blacksquare
\end{align}

\subsubsection{Proof of Thm.~\ref{thm:scenario2} for DP-OGL+ under Threat Model~\ref{threat2}}\label{app:thm:scenario2_dpogl+}
The proof of the theorem for this scenario closely resembles the previous one but with the following adjustments. In Case 1 (in App.~\ref{app:thm:scenario2_dpogl}), under Threat Model~\ref{threat2}, $\hat{\randfunc}_i^{\tau}$ revealed to HbC worker $i$ comprises group models $\model_{m_j}^{S\tau_j+1}$ for $j\in [J]$, where $\groupdist_{m_j,m_{j+1}}=1$, $\tau_{j+1}=\tau_{j}+1$, $m_1=m'$, $m_J=m$, $\tau_1=T_{m,m'}^{\tau}-1$, and $\tau_J=\left\lfloor \frac{\tau}{S} \right\rfloor$. Assuming a string structure, only one such sequence exists. Given $\tilde{\randfunc}_i^{1:\tau,t}$ (as of (\ref{eq:app:thm1:ftilde+})) and the conditions (\ref{app:thm2:con1})-(\ref{app:thm2:con4}), we obtain: 
\begin{align}\label{app:thm2:jconv1+}
&e^{(\alpha-1)R_{\alpha}\left( \distr_{\model_m^{S\tau+1}(\calD)\mid \randfunc_i^{1:\tau}(\calD)= z_{i}^{1:\tau}}  \| \distr_{\model_m^{S\tau+1}(\calD')\mid \randfunc_i^{1:\tau}(\calD')= z_{i}^{1:\tau}} \right)} 
\nonumber \\
&\leq \int_{\hat{\tilde{z}}_i^{1:\tau,t}}\distr_{\hat{\tilde{\randfunc}}_i^{1:\tau,t}(\calD')\mid \text{Condition}_1}(\hat{\tilde{z}}_i^{1:\tau,t})
\times \nonumber \\ & \times 
e^{(\alpha-1)R_{\alpha}\left( \distr_{\model_m^{S\tau+1}(\calD)\mid \text{Condition}_2}  \| \distr_{\model_m^{S\tau+1}(\calD')\mid \text{Condition}_3} \right)}.
\end{align}

Recursively applying Lem.~\ref{lem:scenario2} on the group models in $\hat{\randfunc}_i^{\tau}$,
\begin{align}\label{app:thm2:recursive0+}
&R_{\alpha}\left( \distr_{\hat{\randfunc}_i^{\tau}(\calD)\mid \text{Condition}_2}  \| \distr_{\hat{\randfunc}_i^{\tau}(\calD')\mid \text{Condition}_3} \right)\leq \nonumber  
\\ & \prod_{\tau'' \in (\tau - \tilde{\groupdist}_{m,n} , \tau ]}\bar{\mu}_{m-S\left(\tau -\tau''\right)}^{\tau''} \times \nonumber
\\ & \sum_{m'\in \calM_n}\left(\sum_{\tau'\in \left[\max(1,T_{m,m'}^{\tau}-1), \max(0,T_{m,m'}^{\tau})\right]}\epsilon_{m'}\right),
\end{align}
where  $\bar{\mu}_{m''}^{\tau''}=\frac{\alpha}{\alpha+\bar{h}_{m''}^{\tau''}S{c}_{m''}^2\sigma_{m''}^2}$ and $\epsilon_{m'}=\frac{\alpha}{2\sigma_{m'}^2}$ with $\bar{h}_{m''}^{\tau''}$ obtained as in (\ref{app:eq:lsi:sumClips+}). Following the steps in App.~\ref{app:thm:scenario2_dpogl}, we obtain:
\begin{align}\label{app:thm2:final+}
&\exp\left((\alpha-1)R_{\alpha}(\distr_{{\randfunc}_{i}^{1:t}(\datapool)} \| \distr_{{\randfunc}_{i}^{1:t}(\datapool'))}\right)\nonumber \\ 
&\leq  \prod_{\tau \in \left[\left\lfloor \frac{t-1}{S}\right\rfloor\right]}\exp \left((\alpha-1)\prod_{\tau'' \in (\tau - \tilde{\groupdist}_{m,n} , \tau ]}\bar{\mu}_{m-S\left(\tau -\tau''\right)}^{\tau''} \right. \nonumber 
\\ &\times \left.  \sum_{m'\in \calM_n}\left(\sum_{\tau'\in \left[\max(1,T_{m,m'}^{S\tau}-1), \max(0,T_{m,m'}^{S\tau})\right]}\epsilon_{m'}\right)\right).
\end{align}

\begin{figure*}[ht!]\vspace{-3ex}
\centering 
\subfloat[{\small Structures and $S$}]{\includegraphics[width=0.206\textwidth]{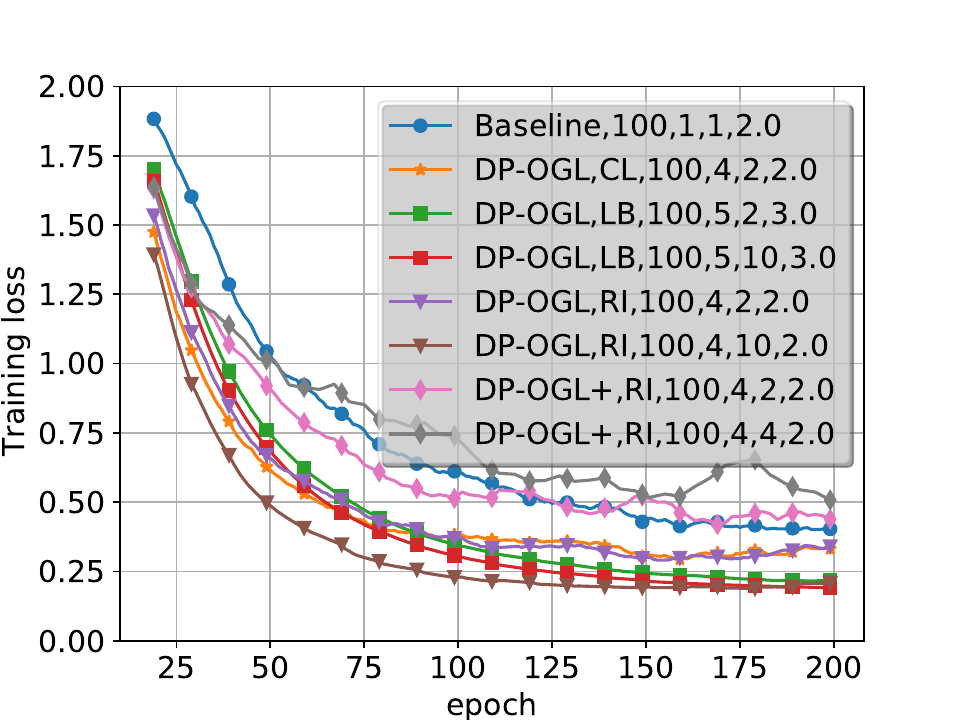
}}
\subfloat[{\small Structures and $S$}]{\includegraphics[width=0.206\textwidth]{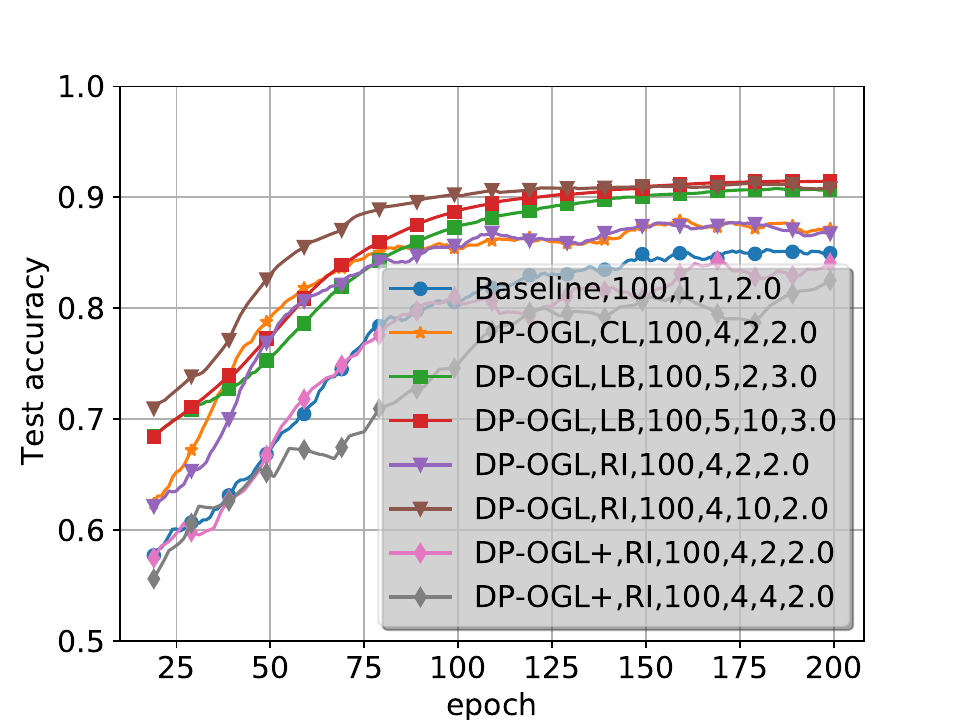
}}
\subfloat[{\small $N/M$ and noise}]{\includegraphics[width=0.206\textwidth]{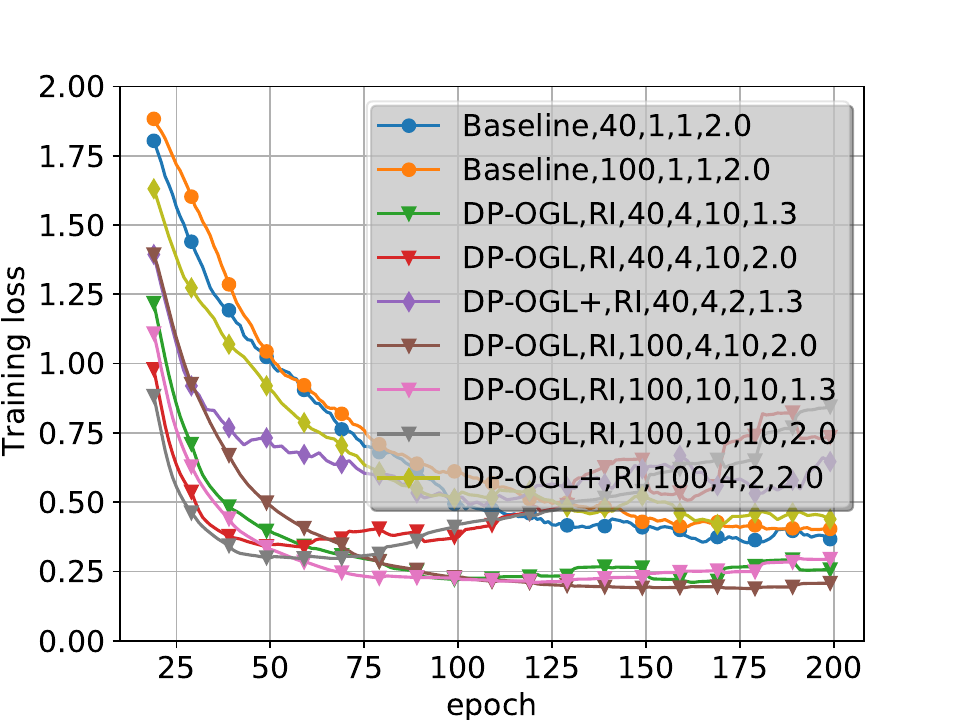
}} 
\subfloat[{\small $N/M$ and noise}]{\includegraphics[width=0.206\textwidth]{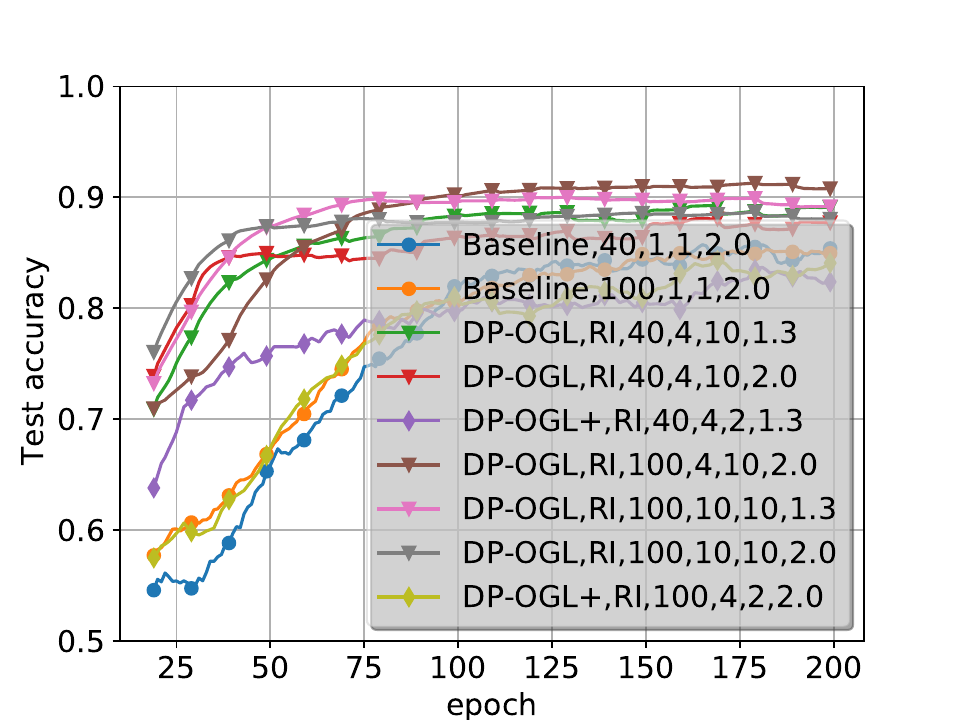
}}
\subfloat[{\small PwP $\epsilon$}]{\includegraphics[width=0.206\textwidth]{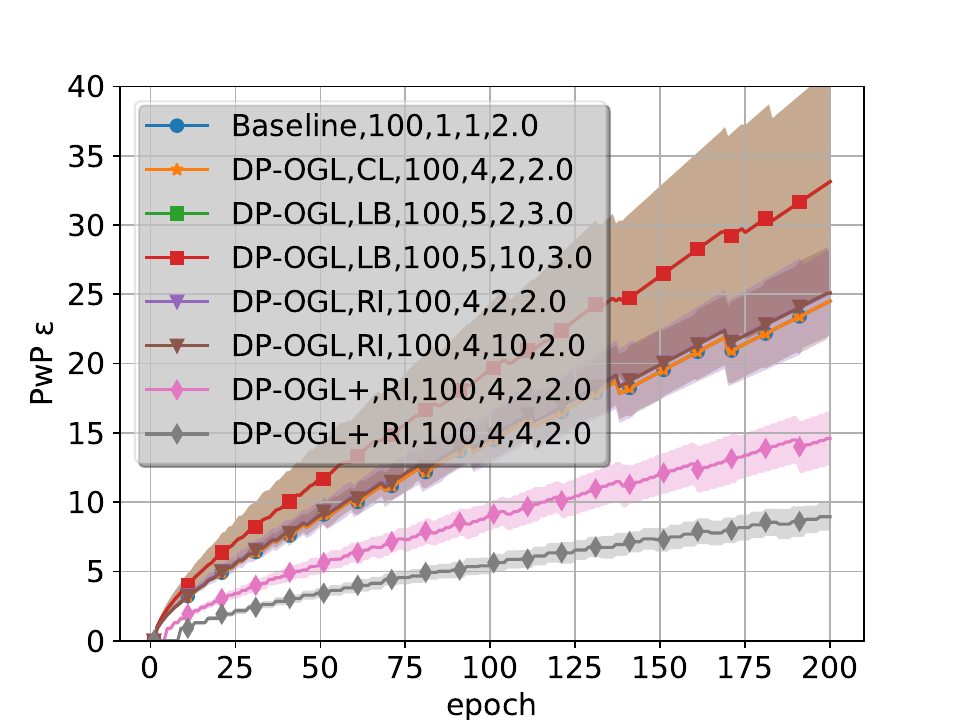
}}
 \caption{In legends, ``LB'', ``CL'', and ``RI'' stand for label-based, clustered, and ring group structures. The legends also include four parameters: the number of workers $N$, the number of groups $M$, the value of $S$, and the noise multiplier $\sigma_m$, listed left to right.}
\label{fig:sim:acc_loss_fmnist} \vspace{-2ex}
\end{figure*}

In Cases~2 and~3 (in App.~\ref{app:thm:scenario2_dpogl}), since the HbC worker $i$ and the targeted worker $n$ have shared group(s), under Threat Model~\ref{threat2}, they mutually trust each other. Therefore, no privacy constraint is necessary for computation between them.

\subsection{Experimental Results on FMNIST Dataset}\label{App:sim}
Our experimental results using the FMNIST dataset, presented in Figs.~\ref{fig:sim:acc_loss_fmnist}(a)-(d), mirror the results illustrated in Fig.~\ref{fig:sim:acc_loss} for the MNIST dataset. Comparing these two sets of figures reveals almost identical convergence behavior across different curves under the same settings. This suggests that our DP-OGL and DP-OGL+ algorithms maintain their superiority compared to the baseline DP-FedAvg algorithm when training CNN on a different dataset, FMNIST. However, there is a slight superiority in every curve of Fig.~\ref{fig:sim:acc_loss} compared to the corresponding curves in Figs.~\ref{fig:sim:acc_loss_fmnist}(a)-(d). This better convergence that is obtained in the MNIST experiments indicates the less complex nature of MNIST when compared to FMNIST. To conserve space, we condensed the curves in Figs.~\ref{fig:sim:acc_loss}(a) and (e) and plotted all corresponding curves for the FMNIST dataset in Fig.~\ref{fig:sim:acc_loss_fmnist}(a). Similarly, Figs.~\ref{fig:sim:acc_loss_fmnist}(b), (c), and (d) combine curves presented, respectively, in Figs.~\ref{fig:sim:acc_loss}(b) and (f), in Figs.~\ref{fig:sim:acc_loss}(c) and (g), and in Figs.~\ref{fig:sim:acc_loss}(d) and (h).

In terms of PwP bounds, Figs.~\ref{fig:sim:acc_loss_fmnist}(e) and \ref{fig:sim:heatmaps}(a) exhibit similar curves, except for the curve corresponding to the label-based structure. This particular curve displays a slight distinction between MNIST and FMNIST datasets. This is because the other group structures we tested are independent of the dataset itself, while the label-based structure depends on how the data points with distinct labels are distributed among the workers. 


\begin{IEEEbiography}[{\includegraphics[width=1in,height=1.25in,clip,keepaspectratio]{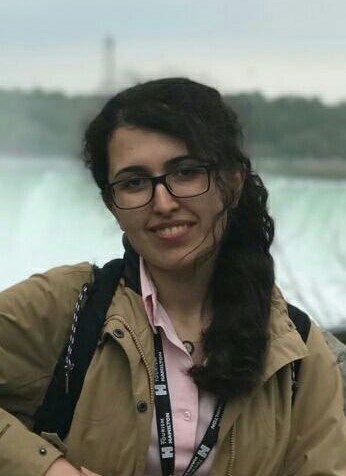}}]{Shahrzad Kianidehkordi} (Graduate Student Member, IEEE) received the
B.Sc. degree in electrical engineering and the minor degree in economics from Sharif University of Technology (SUT), Tehran, Iran, in 2017, and the M.A.Sc. degree in electrical and computer engineering from the University of Toronto (UofT), Toronto, ON, Canada, in 2019, where she is currently pursuing the Ph.D. degree in electrical and computer engineering. During her B.Sc. degree, she conducted research in the Image and Multimedia Processing Laboratory at SUT; and she interned at the Department of Information Engineering, Chinese University of Hong Kong. During her graduate studies, she was certified at the international high-performance computing summer school, Kobe, Japan, on a full scholarship; she interned as a Machine Learning researcher in the Accelerated Neural Technology Team at Huawei, Montreal; and she visited the CISPA Helmholtz Center for Information Security, Germany, granted with Mitacs Globalink Research Award Abroad. Her research interests include distributed learning, approximation, privacy, and coding theory. She won the Gold Medal in the Iranian National Mathematical Olympiad, Iran, in 2011. She has been the recipient of the Ontario Graduate Scholarship (OGS) for 2019–2021, the DiDi graduate award for 2020–2024, and the NSERC Alexander Graham Bell Canada Graduate Scholarship-Doctoral (CGS D3) for 2021-2024. 
\end{IEEEbiography}

\begin{IEEEbiography}[{\includegraphics[width=1in,height=1.25in,clip,keepaspectratio]{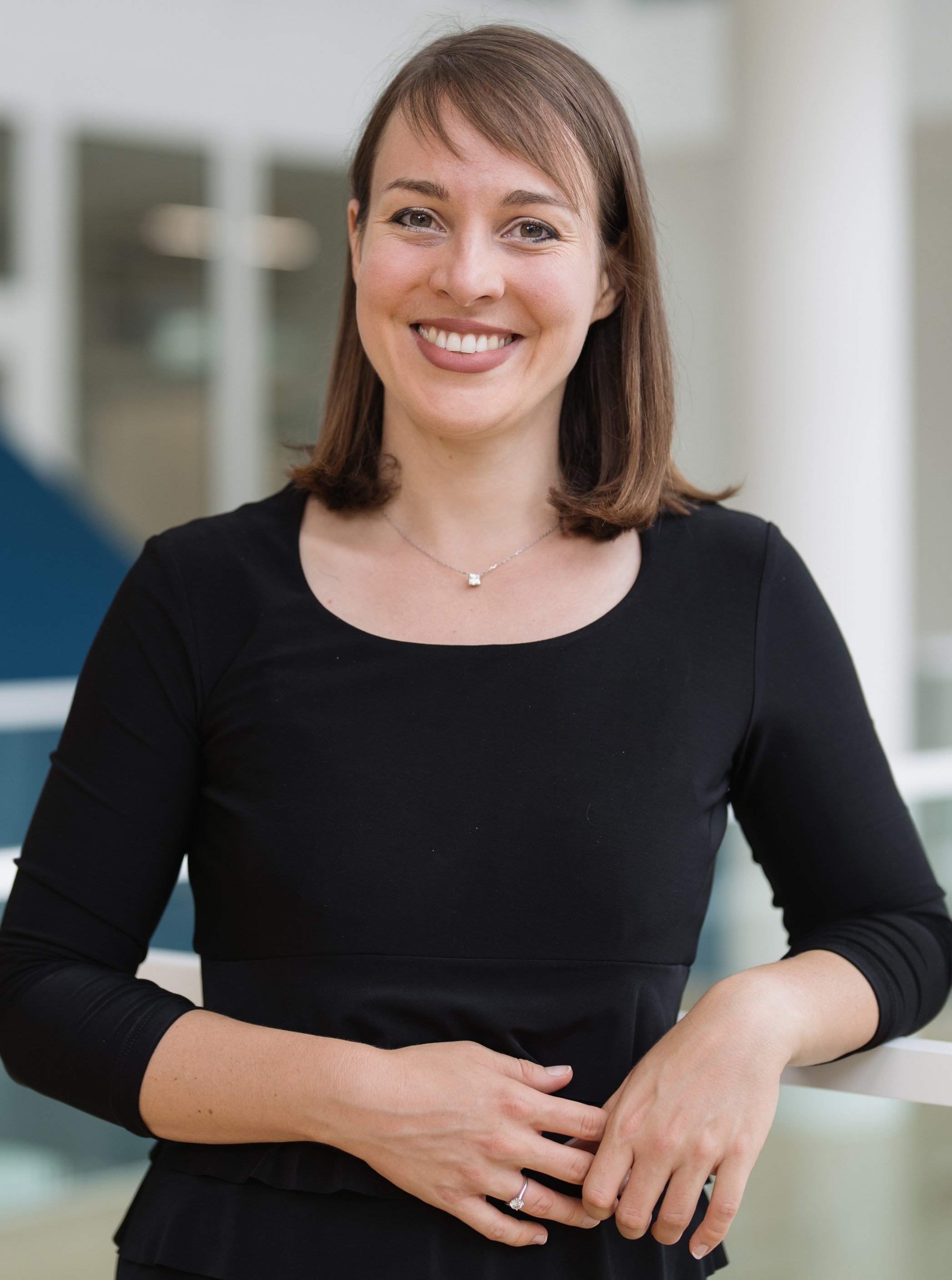}}] 
{Franziska Boenisch} is a tenure-track faculty at the CISPA Helmholtz Center for Information Security where she co-leads the SprintML lab. Before, she was a Postdoctoral Fellow at the University of Toronto and Vector Institute advised by Prof. Nicolas Papernot. Franziska obtained her Ph.D. at the Computer Science Department at Freie University Berlin. During her Ph.D., she was a research associate at the Fraunhofer Institute for Applied and Integrated Security (AISEC), Germany. Her current research centers around how to make foundation models more private and trustworthy. Franziska received a Fraunhofer TALENTA grant for outstanding female early career researchers, the German Industrial Research Foundation prize for her applied research on machine learning privacy, and the Fraunhofer ICT Dissertation Award 2023.
\end{IEEEbiography}

\begin{IEEEbiography}[{\includegraphics[width=1in,height=1.25in,clip,keepaspectratio]{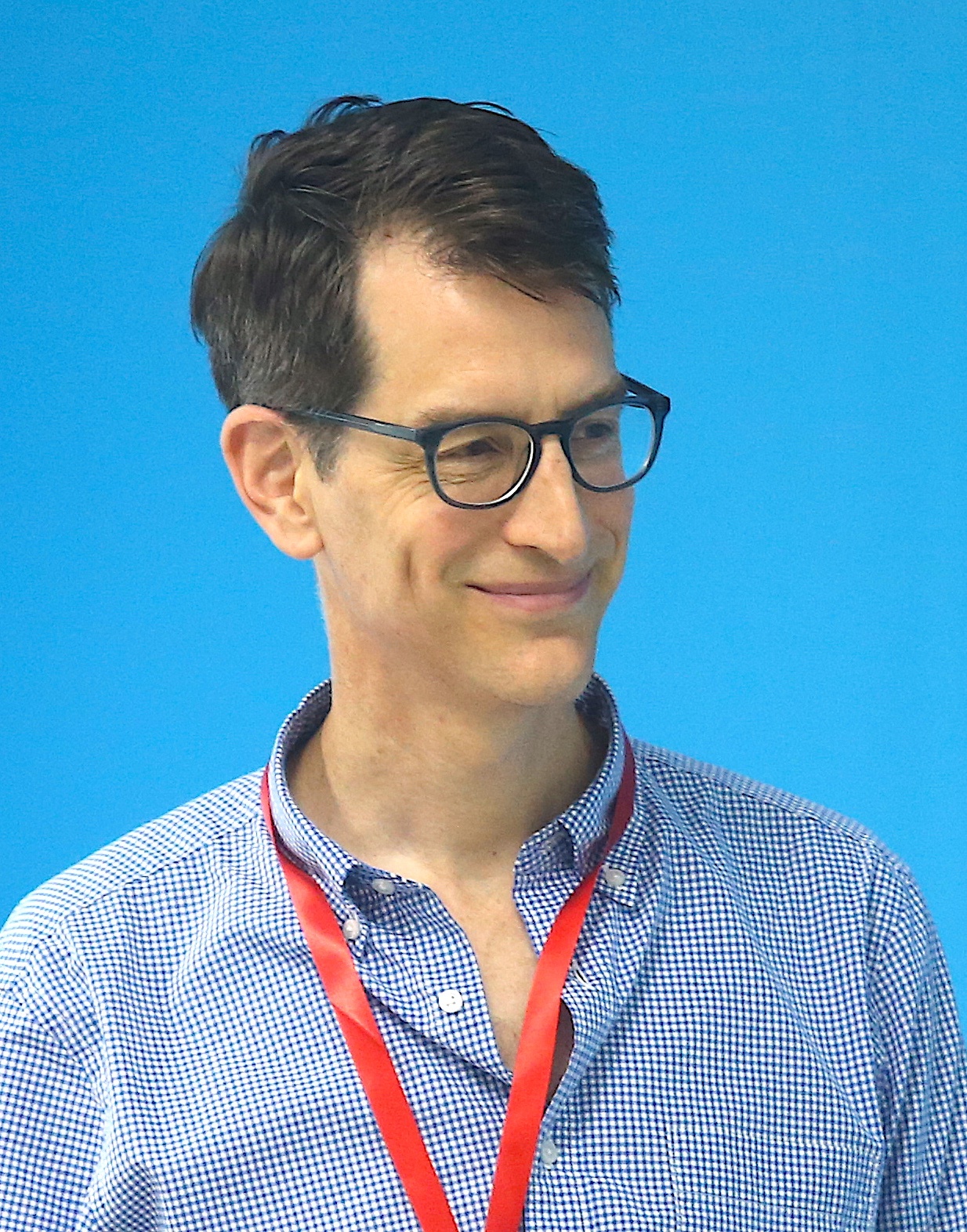}}] {Stark C. Draper} (Senior Member, IEEE) received the B.S. degree in
Electrical Engineering and the B.A. degree in History from Stanford
University, and the M.S. and Ph.D. degrees in Electrical Engineering
and Computer Science from the Massachusetts Institute of Technology
(MIT).  He completed postdocs at the University of Toronto (UofT) and
at the University of California, Berkeley. He is a Professor in the
Department of Electrical and Computer Engineering at the University of
Toronto and was an Associate Professor at the University of Wisconsin,
Madison. As a Research Scientist he has worked at the Mitsubishi
Electric Research Labs (MERL), Disney's Boston Research Lab, Arraycomm
Inc., the C. S. Draper Laboratory, and Ktaadn Inc. His research
interests include information theory, optimization, error-correction
coding, security, and the application of tools and perspectives from
these fields in communications, computing, learning, and astronomy. He
has been the recipient of the NSERC Discovery Award, the NSF CAREER
Award, the 2010 MERL President's Award, and teaching awards from UofT,
the University of Wisconsin, and MIT. He received an Intel Graduate
Fellowship, Stanford's Frederick E. Terman Engineering Scholastic
Award, and a U.S. State Department Fulbright Fellowship. He spent the
2019–2020 academic year on sabbatical visiting the Chinese University
of Hong Kong, Shenzhen, and the Canada-France-Hawaii Telescope (CFHT),
Hawaii, USA. Among his service roles, he was the founding chair of the
Machine Intelligence major at UofT, was the Faculty of Applied Science
and Engineering (FASE) representative on the UofT Governing Council,
is the FASE Vice-Dean of Research, and is the President of the IEEE
Information Theory Society for 2024.

\end{IEEEbiography}

\end{document}